\RequirePackage[svgnames]{xcolor}
\RequirePackage{amsmath,amsthm,amssymb}
\PassOptionsToPackage{section}{placeins}
\PassOptionsToPackage{normalem}{ulem}
\PassOptionsToPackage{expansion=false}{microtype}
\documentclass{bytedance_seed}
\DeclareFontShape{T1}{bytesans}{b}{n}{<-> s * [1] seed/bytesans}{}
\DeclareFontShape{T1}{bytesans}{bx}{n}{<-> s * [1] seed/bytesans}{}
\DeclareFontFamily{T1}{opdcmr}{}
\pdfmapline{+opd98ecrm0500 SFRM0500 "0.98 ExtendFont T1Encoding ReEncodeFont" <cm-super-t1.enc <sfrm0500.pfb}
\pdfmapline{+opd98ecrm0600 SFRM0600 "0.98 ExtendFont T1Encoding ReEncodeFont" <cm-super-t1.enc <sfrm0600.pfb}
\pdfmapline{+opd98ecrm0700 SFRM0700 "0.98 ExtendFont T1Encoding ReEncodeFont" <cm-super-t1.enc <sfrm0700.pfb}
\pdfmapline{+opd98ecrm0800 SFRM0800 "0.98 ExtendFont T1Encoding ReEncodeFont" <cm-super-t1.enc <sfrm0800.pfb}
\pdfmapline{+opd98ecrm0900 SFRM0900 "0.98 ExtendFont T1Encoding ReEncodeFont" <cm-super-t1.enc <sfrm0900.pfb}
\pdfmapline{+opd98ecrm1000 SFRM1000 "0.98 ExtendFont T1Encoding ReEncodeFont" <cm-super-t1.enc <sfrm1000.pfb}
\pdfmapline{+opd98ecrm1095 SFRM1095 "0.98 ExtendFont T1Encoding ReEncodeFont" <cm-super-t1.enc <sfrm1095.pfb}
\pdfmapline{+opd98ecrm1200 SFRM1200 "0.98 ExtendFont T1Encoding ReEncodeFont" <cm-super-t1.enc <sfrm1200.pfb}
\pdfmapline{+opd98ecrm1440 SFRM1440 "0.98 ExtendFont T1Encoding ReEncodeFont" <cm-super-t1.enc <sfrm1440.pfb}
\pdfmapline{+opd98ecrm1728 SFRM1728 "0.98 ExtendFont T1Encoding ReEncodeFont" <cm-super-t1.enc <sfrm1728.pfb}
\pdfmapline{+opd98ecrm2074 SFRM2074 "0.98 ExtendFont T1Encoding ReEncodeFont" <cm-super-t1.enc <sfrm2074.pfb}
\pdfmapline{+opd98ecrm2488 SFRM2488 "0.98 ExtendFont T1Encoding ReEncodeFont" <cm-super-t1.enc <sfrm2488.pfb}
\pdfmapline{+opd98ecrm2986 SFRM2986 "0.98 ExtendFont T1Encoding ReEncodeFont" <cm-super-t1.enc <sfrm2986.pfb}
\pdfmapline{+opd98ecrm3583 SFRM3583 "0.98 ExtendFont T1Encoding ReEncodeFont" <cm-super-t1.enc <sfrm3583.pfb}
\DeclareFontShape{T1}{opdcmr}{m}{n}{<-5.5> opd-fonts98/opd98ecrm0500
  <5.5-6.5> opd-fonts98/opd98ecrm0600
  <6.5-7.5> opd-fonts98/opd98ecrm0700
  <7.5-8.5> opd-fonts98/opd98ecrm0800
  <8.5-9.5> opd-fonts98/opd98ecrm0900
  <9.5-10.475> opd-fonts98/opd98ecrm1000
  <10.475-11.475> opd-fonts98/opd98ecrm1095
  <11.475-13.2> opd-fonts98/opd98ecrm1200
  <13.2-15.84> opd-fonts98/opd98ecrm1440
  <15.84-19.009999999999998> opd-fonts98/opd98ecrm1728
  <19.009999999999998-22.81> opd-fonts98/opd98ecrm2074
  <22.81-27.369999999999997> opd-fonts98/opd98ecrm2488
  <27.369999999999997-32.845> opd-fonts98/opd98ecrm2986
  <32.845-> opd-fonts98/opd98ecrm3583}{}
\pdfmapline{+opd98ecsl0500 SFSL0500 "0.98 ExtendFont T1Encoding ReEncodeFont" <cm-super-t1.enc <sfsl0500.pfb}
\pdfmapline{+opd98ecsl0600 SFSL0600 "0.98 ExtendFont T1Encoding ReEncodeFont" <cm-super-t1.enc <sfsl0600.pfb}
\pdfmapline{+opd98ecsl0700 SFSL0700 "0.98 ExtendFont T1Encoding ReEncodeFont" <cm-super-t1.enc <sfsl0700.pfb}
\pdfmapline{+opd98ecsl0800 SFSL0800 "0.98 ExtendFont T1Encoding ReEncodeFont" <cm-super-t1.enc <sfsl0800.pfb}
\pdfmapline{+opd98ecsl0900 SFSL0900 "0.98 ExtendFont T1Encoding ReEncodeFont" <cm-super-t1.enc <sfsl0900.pfb}
\pdfmapline{+opd98ecsl1000 SFSL1000 "0.98 ExtendFont T1Encoding ReEncodeFont" <cm-super-t1.enc <sfsl1000.pfb}
\pdfmapline{+opd98ecsl1095 SFSL1095 "0.98 ExtendFont T1Encoding ReEncodeFont" <cm-super-t1.enc <sfsl1095.pfb}
\pdfmapline{+opd98ecsl1200 SFSL1200 "0.98 ExtendFont T1Encoding ReEncodeFont" <cm-super-t1.enc <sfsl1200.pfb}
\pdfmapline{+opd98ecsl1440 SFSL1440 "0.98 ExtendFont T1Encoding ReEncodeFont" <cm-super-t1.enc <sfsl1440.pfb}
\pdfmapline{+opd98ecsl1728 SFSL1728 "0.98 ExtendFont T1Encoding ReEncodeFont" <cm-super-t1.enc <sfsl1728.pfb}
\pdfmapline{+opd98ecsl2074 SFSL2074 "0.98 ExtendFont T1Encoding ReEncodeFont" <cm-super-t1.enc <sfsl2074.pfb}
\pdfmapline{+opd98ecsl2488 SFSL2488 "0.98 ExtendFont T1Encoding ReEncodeFont" <cm-super-t1.enc <sfsl2488.pfb}
\pdfmapline{+opd98ecsl2986 SFSL2986 "0.98 ExtendFont T1Encoding ReEncodeFont" <cm-super-t1.enc <sfsl2986.pfb}
\pdfmapline{+opd98ecsl3583 SFSL3583 "0.98 ExtendFont T1Encoding ReEncodeFont" <cm-super-t1.enc <sfsl3583.pfb}
\DeclareFontShape{T1}{opdcmr}{m}{sl}{<-5.5> opd-fonts98/opd98ecsl0500
  <5.5-6.5> opd-fonts98/opd98ecsl0600
  <6.5-7.5> opd-fonts98/opd98ecsl0700
  <7.5-8.5> opd-fonts98/opd98ecsl0800
  <8.5-9.5> opd-fonts98/opd98ecsl0900
  <9.5-10.475> opd-fonts98/opd98ecsl1000
  <10.475-11.475> opd-fonts98/opd98ecsl1095
  <11.475-13.2> opd-fonts98/opd98ecsl1200
  <13.2-15.84> opd-fonts98/opd98ecsl1440
  <15.84-19.009999999999998> opd-fonts98/opd98ecsl1728
  <19.009999999999998-22.81> opd-fonts98/opd98ecsl2074
  <22.81-27.369999999999997> opd-fonts98/opd98ecsl2488
  <27.369999999999997-32.845> opd-fonts98/opd98ecsl2986
  <32.845-> opd-fonts98/opd98ecsl3583}{}
\pdfmapline{+opd98ecti0500 SFTI0500 "0.98 ExtendFont T1Encoding ReEncodeFont" <cm-super-t1.enc <sfti0500.pfb}
\pdfmapline{+opd98ecti0600 SFTI0600 "0.98 ExtendFont T1Encoding ReEncodeFont" <cm-super-t1.enc <sfti0600.pfb}
\pdfmapline{+opd98ecti0700 SFTI0700 "0.98 ExtendFont T1Encoding ReEncodeFont" <cm-super-t1.enc <sfti0700.pfb}
\pdfmapline{+opd98ecti0800 SFTI0800 "0.98 ExtendFont T1Encoding ReEncodeFont" <cm-super-t1.enc <sfti0800.pfb}
\pdfmapline{+opd98ecti0900 SFTI0900 "0.98 ExtendFont T1Encoding ReEncodeFont" <cm-super-t1.enc <sfti0900.pfb}
\pdfmapline{+opd98ecti1000 SFTI1000 "0.98 ExtendFont T1Encoding ReEncodeFont" <cm-super-t1.enc <sfti1000.pfb}
\pdfmapline{+opd98ecti1095 SFTI1095 "0.98 ExtendFont T1Encoding ReEncodeFont" <cm-super-t1.enc <sfti1095.pfb}
\pdfmapline{+opd98ecti1200 SFTI1200 "0.98 ExtendFont T1Encoding ReEncodeFont" <cm-super-t1.enc <sfti1200.pfb}
\pdfmapline{+opd98ecti1440 SFTI1440 "0.98 ExtendFont T1Encoding ReEncodeFont" <cm-super-t1.enc <sfti1440.pfb}
\pdfmapline{+opd98ecti1728 SFTI1728 "0.98 ExtendFont T1Encoding ReEncodeFont" <cm-super-t1.enc <sfti1728.pfb}
\pdfmapline{+opd98ecti2074 SFTI2074 "0.98 ExtendFont T1Encoding ReEncodeFont" <cm-super-t1.enc <sfti2074.pfb}
\pdfmapline{+opd98ecti2488 SFTI2488 "0.98 ExtendFont T1Encoding ReEncodeFont" <cm-super-t1.enc <sfti2488.pfb}
\pdfmapline{+opd98ecti2986 SFTI2986 "0.98 ExtendFont T1Encoding ReEncodeFont" <cm-super-t1.enc <sfti2986.pfb}
\pdfmapline{+opd98ecti3583 SFTI3583 "0.98 ExtendFont T1Encoding ReEncodeFont" <cm-super-t1.enc <sfti3583.pfb}
\DeclareFontShape{T1}{opdcmr}{m}{it}{<-5.5> opd-fonts98/opd98ecti0500
  <5.5-6.5> opd-fonts98/opd98ecti0600
  <6.5-7.5> opd-fonts98/opd98ecti0700
  <7.5-8.5> opd-fonts98/opd98ecti0800
  <8.5-9.5> opd-fonts98/opd98ecti0900
  <9.5-10.475> opd-fonts98/opd98ecti1000
  <10.475-11.475> opd-fonts98/opd98ecti1095
  <11.475-13.2> opd-fonts98/opd98ecti1200
  <13.2-15.84> opd-fonts98/opd98ecti1440
  <15.84-19.009999999999998> opd-fonts98/opd98ecti1728
  <19.009999999999998-22.81> opd-fonts98/opd98ecti2074
  <22.81-27.369999999999997> opd-fonts98/opd98ecti2488
  <27.369999999999997-32.845> opd-fonts98/opd98ecti2986
  <32.845-> opd-fonts98/opd98ecti3583}{}
\pdfmapline{+opd98eccc0500 SFCC0500 "0.98 ExtendFont T1Encoding ReEncodeFont" <cm-super-t1.enc <sfcc0500.pfb}
\pdfmapline{+opd98eccc0600 SFCC0600 "0.98 ExtendFont T1Encoding ReEncodeFont" <cm-super-t1.enc <sfcc0600.pfb}
\pdfmapline{+opd98eccc0700 SFCC0700 "0.98 ExtendFont T1Encoding ReEncodeFont" <cm-super-t1.enc <sfcc0700.pfb}
\pdfmapline{+opd98eccc0800 SFCC0800 "0.98 ExtendFont T1Encoding ReEncodeFont" <cm-super-t1.enc <sfcc0800.pfb}
\pdfmapline{+opd98eccc0900 SFCC0900 "0.98 ExtendFont T1Encoding ReEncodeFont" <cm-super-t1.enc <sfcc0900.pfb}
\pdfmapline{+opd98eccc1000 SFCC1000 "0.98 ExtendFont T1Encoding ReEncodeFont" <cm-super-t1.enc <sfcc1000.pfb}
\pdfmapline{+opd98eccc1095 SFCC1095 "0.98 ExtendFont T1Encoding ReEncodeFont" <cm-super-t1.enc <sfcc1095.pfb}
\pdfmapline{+opd98eccc1200 SFCC1200 "0.98 ExtendFont T1Encoding ReEncodeFont" <cm-super-t1.enc <sfcc1200.pfb}
\pdfmapline{+opd98eccc1440 SFCC1440 "0.98 ExtendFont T1Encoding ReEncodeFont" <cm-super-t1.enc <sfcc1440.pfb}
\pdfmapline{+opd98eccc1728 SFCC1728 "0.98 ExtendFont T1Encoding ReEncodeFont" <cm-super-t1.enc <sfcc1728.pfb}
\pdfmapline{+opd98eccc2074 SFCC2074 "0.98 ExtendFont T1Encoding ReEncodeFont" <cm-super-t1.enc <sfcc2074.pfb}
\pdfmapline{+opd98eccc2488 SFCC2488 "0.98 ExtendFont T1Encoding ReEncodeFont" <cm-super-t1.enc <sfcc2488.pfb}
\pdfmapline{+opd98eccc2986 SFCC2986 "0.98 ExtendFont T1Encoding ReEncodeFont" <cm-super-t1.enc <sfcc2986.pfb}
\pdfmapline{+opd98eccc3583 SFCC3583 "0.98 ExtendFont T1Encoding ReEncodeFont" <cm-super-t1.enc <sfcc3583.pfb}
\DeclareFontShape{T1}{opdcmr}{m}{sc}{<-5.5> opd-fonts98/opd98eccc0500
  <5.5-6.5> opd-fonts98/opd98eccc0600
  <6.5-7.5> opd-fonts98/opd98eccc0700
  <7.5-8.5> opd-fonts98/opd98eccc0800
  <8.5-9.5> opd-fonts98/opd98eccc0900
  <9.5-10.475> opd-fonts98/opd98eccc1000
  <10.475-11.475> opd-fonts98/opd98eccc1095
  <11.475-13.2> opd-fonts98/opd98eccc1200
  <13.2-15.84> opd-fonts98/opd98eccc1440
  <15.84-19.009999999999998> opd-fonts98/opd98eccc1728
  <19.009999999999998-22.81> opd-fonts98/opd98eccc2074
  <22.81-27.369999999999997> opd-fonts98/opd98eccc2488
  <27.369999999999997-32.845> opd-fonts98/opd98eccc2986
  <32.845-> opd-fonts98/opd98eccc3583}{}
\pdfmapline{+opd98ecbx0500 SFBX0500 "0.98 ExtendFont T1Encoding ReEncodeFont" <cm-super-t1.enc <sfbx0500.pfb}
\pdfmapline{+opd98ecbx0600 SFBX0600 "0.98 ExtendFont T1Encoding ReEncodeFont" <cm-super-t1.enc <sfbx0600.pfb}
\pdfmapline{+opd98ecbx0700 SFBX0700 "0.98 ExtendFont T1Encoding ReEncodeFont" <cm-super-t1.enc <sfbx0700.pfb}
\pdfmapline{+opd98ecbx0800 SFBX0800 "0.98 ExtendFont T1Encoding ReEncodeFont" <cm-super-t1.enc <sfbx0800.pfb}
\pdfmapline{+opd98ecbx0900 SFBX0900 "0.98 ExtendFont T1Encoding ReEncodeFont" <cm-super-t1.enc <sfbx0900.pfb}
\pdfmapline{+opd98ecbx1000 SFBX1000 "0.98 ExtendFont T1Encoding ReEncodeFont" <cm-super-t1.enc <sfbx1000.pfb}
\pdfmapline{+opd98ecbx1095 SFBX1095 "0.98 ExtendFont T1Encoding ReEncodeFont" <cm-super-t1.enc <sfbx1095.pfb}
\pdfmapline{+opd98ecbx1200 SFBX1200 "0.98 ExtendFont T1Encoding ReEncodeFont" <cm-super-t1.enc <sfbx1200.pfb}
\pdfmapline{+opd98ecbx1440 SFBX1440 "0.98 ExtendFont T1Encoding ReEncodeFont" <cm-super-t1.enc <sfbx1440.pfb}
\pdfmapline{+opd98ecbx1728 SFBX1728 "0.98 ExtendFont T1Encoding ReEncodeFont" <cm-super-t1.enc <sfbx1728.pfb}
\pdfmapline{+opd98ecbx2074 SFBX2074 "0.98 ExtendFont T1Encoding ReEncodeFont" <cm-super-t1.enc <sfbx2074.pfb}
\pdfmapline{+opd98ecbx2488 SFBX2488 "0.98 ExtendFont T1Encoding ReEncodeFont" <cm-super-t1.enc <sfbx2488.pfb}
\pdfmapline{+opd98ecbx2986 SFBX2986 "0.98 ExtendFont T1Encoding ReEncodeFont" <cm-super-t1.enc <sfbx2986.pfb}
\pdfmapline{+opd98ecbx3583 SFBX3583 "0.98 ExtendFont T1Encoding ReEncodeFont" <cm-super-t1.enc <sfbx3583.pfb}
\DeclareFontShape{T1}{opdcmr}{bx}{n}{<-5.5> opd-fonts98/opd98ecbx0500
  <5.5-6.5> opd-fonts98/opd98ecbx0600
  <6.5-7.5> opd-fonts98/opd98ecbx0700
  <7.5-8.5> opd-fonts98/opd98ecbx0800
  <8.5-9.5> opd-fonts98/opd98ecbx0900
  <9.5-10.475> opd-fonts98/opd98ecbx1000
  <10.475-11.475> opd-fonts98/opd98ecbx1095
  <11.475-13.2> opd-fonts98/opd98ecbx1200
  <13.2-15.84> opd-fonts98/opd98ecbx1440
  <15.84-19.009999999999998> opd-fonts98/opd98ecbx1728
  <19.009999999999998-22.81> opd-fonts98/opd98ecbx2074
  <22.81-27.369999999999997> opd-fonts98/opd98ecbx2488
  <27.369999999999997-32.845> opd-fonts98/opd98ecbx2986
  <32.845-> opd-fonts98/opd98ecbx3583}{}
\pdfmapline{+opd98ecrb0500 SFRB0500 "0.98 ExtendFont T1Encoding ReEncodeFont" <cm-super-t1.enc <sfrb0500.pfb}
\pdfmapline{+opd98ecrb0600 SFRB0600 "0.98 ExtendFont T1Encoding ReEncodeFont" <cm-super-t1.enc <sfrb0600.pfb}
\pdfmapline{+opd98ecrb0700 SFRB0700 "0.98 ExtendFont T1Encoding ReEncodeFont" <cm-super-t1.enc <sfrb0700.pfb}
\pdfmapline{+opd98ecrb0800 SFRB0800 "0.98 ExtendFont T1Encoding ReEncodeFont" <cm-super-t1.enc <sfrb0800.pfb}
\pdfmapline{+opd98ecrb0900 SFRB0900 "0.98 ExtendFont T1Encoding ReEncodeFont" <cm-super-t1.enc <sfrb0900.pfb}
\pdfmapline{+opd98ecrb1000 SFRB1000 "0.98 ExtendFont T1Encoding ReEncodeFont" <cm-super-t1.enc <sfrb1000.pfb}
\pdfmapline{+opd98ecrb1095 SFRB1095 "0.98 ExtendFont T1Encoding ReEncodeFont" <cm-super-t1.enc <sfrb1095.pfb}
\pdfmapline{+opd98ecrb1200 SFRB1200 "0.98 ExtendFont T1Encoding ReEncodeFont" <cm-super-t1.enc <sfrb1200.pfb}
\pdfmapline{+opd98ecrb1440 SFRB1440 "0.98 ExtendFont T1Encoding ReEncodeFont" <cm-super-t1.enc <sfrb1440.pfb}
\pdfmapline{+opd98ecrb1728 SFRB1728 "0.98 ExtendFont T1Encoding ReEncodeFont" <cm-super-t1.enc <sfrb1728.pfb}
\pdfmapline{+opd98ecrb2074 SFRB2074 "0.98 ExtendFont T1Encoding ReEncodeFont" <cm-super-t1.enc <sfrb2074.pfb}
\pdfmapline{+opd98ecrb2488 SFRB2488 "0.98 ExtendFont T1Encoding ReEncodeFont" <cm-super-t1.enc <sfrb2488.pfb}
\pdfmapline{+opd98ecrb2986 SFRB2986 "0.98 ExtendFont T1Encoding ReEncodeFont" <cm-super-t1.enc <sfrb2986.pfb}
\pdfmapline{+opd98ecrb3583 SFRB3583 "0.98 ExtendFont T1Encoding ReEncodeFont" <cm-super-t1.enc <sfrb3583.pfb}
\DeclareFontShape{T1}{opdcmr}{b}{n}{<-5.5> opd-fonts98/opd98ecrb0500
  <5.5-6.5> opd-fonts98/opd98ecrb0600
  <6.5-7.5> opd-fonts98/opd98ecrb0700
  <7.5-8.5> opd-fonts98/opd98ecrb0800
  <8.5-9.5> opd-fonts98/opd98ecrb0900
  <9.5-10.475> opd-fonts98/opd98ecrb1000
  <10.475-11.475> opd-fonts98/opd98ecrb1095
  <11.475-13.2> opd-fonts98/opd98ecrb1200
  <13.2-15.84> opd-fonts98/opd98ecrb1440
  <15.84-19.009999999999998> opd-fonts98/opd98ecrb1728
  <19.009999999999998-22.81> opd-fonts98/opd98ecrb2074
  <22.81-27.369999999999997> opd-fonts98/opd98ecrb2488
  <27.369999999999997-32.845> opd-fonts98/opd98ecrb2986
  <32.845-> opd-fonts98/opd98ecrb3583}{}
\pdfmapline{+opd98ecbi0500 SFBI0500 "0.98 ExtendFont T1Encoding ReEncodeFont" <cm-super-t1.enc <sfbi0500.pfb}
\pdfmapline{+opd98ecbi0600 SFBI0600 "0.98 ExtendFont T1Encoding ReEncodeFont" <cm-super-t1.enc <sfbi0600.pfb}
\pdfmapline{+opd98ecbi0700 SFBI0700 "0.98 ExtendFont T1Encoding ReEncodeFont" <cm-super-t1.enc <sfbi0700.pfb}
\pdfmapline{+opd98ecbi0800 SFBI0800 "0.98 ExtendFont T1Encoding ReEncodeFont" <cm-super-t1.enc <sfbi0800.pfb}
\pdfmapline{+opd98ecbi0900 SFBI0900 "0.98 ExtendFont T1Encoding ReEncodeFont" <cm-super-t1.enc <sfbi0900.pfb}
\pdfmapline{+opd98ecbi1000 SFBI1000 "0.98 ExtendFont T1Encoding ReEncodeFont" <cm-super-t1.enc <sfbi1000.pfb}
\pdfmapline{+opd98ecbi1095 SFBI1095 "0.98 ExtendFont T1Encoding ReEncodeFont" <cm-super-t1.enc <sfbi1095.pfb}
\pdfmapline{+opd98ecbi1200 SFBI1200 "0.98 ExtendFont T1Encoding ReEncodeFont" <cm-super-t1.enc <sfbi1200.pfb}
\pdfmapline{+opd98ecbi1440 SFBI1440 "0.98 ExtendFont T1Encoding ReEncodeFont" <cm-super-t1.enc <sfbi1440.pfb}
\pdfmapline{+opd98ecbi1728 SFBI1728 "0.98 ExtendFont T1Encoding ReEncodeFont" <cm-super-t1.enc <sfbi1728.pfb}
\pdfmapline{+opd98ecbi2074 SFBI2074 "0.98 ExtendFont T1Encoding ReEncodeFont" <cm-super-t1.enc <sfbi2074.pfb}
\pdfmapline{+opd98ecbi2488 SFBI2488 "0.98 ExtendFont T1Encoding ReEncodeFont" <cm-super-t1.enc <sfbi2488.pfb}
\pdfmapline{+opd98ecbi2986 SFBI2986 "0.98 ExtendFont T1Encoding ReEncodeFont" <cm-super-t1.enc <sfbi2986.pfb}
\pdfmapline{+opd98ecbi3583 SFBI3583 "0.98 ExtendFont T1Encoding ReEncodeFont" <cm-super-t1.enc <sfbi3583.pfb}
\DeclareFontShape{T1}{opdcmr}{bx}{it}{<-5.5> opd-fonts98/opd98ecbi0500
  <5.5-6.5> opd-fonts98/opd98ecbi0600
  <6.5-7.5> opd-fonts98/opd98ecbi0700
  <7.5-8.5> opd-fonts98/opd98ecbi0800
  <8.5-9.5> opd-fonts98/opd98ecbi0900
  <9.5-10.475> opd-fonts98/opd98ecbi1000
  <10.475-11.475> opd-fonts98/opd98ecbi1095
  <11.475-13.2> opd-fonts98/opd98ecbi1200
  <13.2-15.84> opd-fonts98/opd98ecbi1440
  <15.84-19.009999999999998> opd-fonts98/opd98ecbi1728
  <19.009999999999998-22.81> opd-fonts98/opd98ecbi2074
  <22.81-27.369999999999997> opd-fonts98/opd98ecbi2488
  <27.369999999999997-32.845> opd-fonts98/opd98ecbi2986
  <32.845-> opd-fonts98/opd98ecbi3583}{}
\pdfmapline{+opd98ecbl0500 SFBL0500 "0.98 ExtendFont T1Encoding ReEncodeFont" <cm-super-t1.enc <sfbl0500.pfb}
\pdfmapline{+opd98ecbl0600 SFBL0600 "0.98 ExtendFont T1Encoding ReEncodeFont" <cm-super-t1.enc <sfbl0600.pfb}
\pdfmapline{+opd98ecbl0700 SFBL0700 "0.98 ExtendFont T1Encoding ReEncodeFont" <cm-super-t1.enc <sfbl0700.pfb}
\pdfmapline{+opd98ecbl0800 SFBL0800 "0.98 ExtendFont T1Encoding ReEncodeFont" <cm-super-t1.enc <sfbl0800.pfb}
\pdfmapline{+opd98ecbl0900 SFBL0900 "0.98 ExtendFont T1Encoding ReEncodeFont" <cm-super-t1.enc <sfbl0900.pfb}
\pdfmapline{+opd98ecbl1000 SFBL1000 "0.98 ExtendFont T1Encoding ReEncodeFont" <cm-super-t1.enc <sfbl1000.pfb}
\pdfmapline{+opd98ecbl1095 SFBL1095 "0.98 ExtendFont T1Encoding ReEncodeFont" <cm-super-t1.enc <sfbl1095.pfb}
\pdfmapline{+opd98ecbl1200 SFBL1200 "0.98 ExtendFont T1Encoding ReEncodeFont" <cm-super-t1.enc <sfbl1200.pfb}
\pdfmapline{+opd98ecbl1440 SFBL1440 "0.98 ExtendFont T1Encoding ReEncodeFont" <cm-super-t1.enc <sfbl1440.pfb}
\pdfmapline{+opd98ecbl1728 SFBL1728 "0.98 ExtendFont T1Encoding ReEncodeFont" <cm-super-t1.enc <sfbl1728.pfb}
\pdfmapline{+opd98ecbl2074 SFBL2074 "0.98 ExtendFont T1Encoding ReEncodeFont" <cm-super-t1.enc <sfbl2074.pfb}
\pdfmapline{+opd98ecbl2488 SFBL2488 "0.98 ExtendFont T1Encoding ReEncodeFont" <cm-super-t1.enc <sfbl2488.pfb}
\pdfmapline{+opd98ecbl2986 SFBL2986 "0.98 ExtendFont T1Encoding ReEncodeFont" <cm-super-t1.enc <sfbl2986.pfb}
\pdfmapline{+opd98ecbl3583 SFBL3583 "0.98 ExtendFont T1Encoding ReEncodeFont" <cm-super-t1.enc <sfbl3583.pfb}
\DeclareFontShape{T1}{opdcmr}{bx}{sl}{<-5.5> opd-fonts98/opd98ecbl0500
  <5.5-6.5> opd-fonts98/opd98ecbl0600
  <6.5-7.5> opd-fonts98/opd98ecbl0700
  <7.5-8.5> opd-fonts98/opd98ecbl0800
  <8.5-9.5> opd-fonts98/opd98ecbl0900
  <9.5-10.475> opd-fonts98/opd98ecbl1000
  <10.475-11.475> opd-fonts98/opd98ecbl1095
  <11.475-13.2> opd-fonts98/opd98ecbl1200
  <13.2-15.84> opd-fonts98/opd98ecbl1440
  <15.84-19.009999999999998> opd-fonts98/opd98ecbl1728
  <19.009999999999998-22.81> opd-fonts98/opd98ecbl2074
  <22.81-27.369999999999997> opd-fonts98/opd98ecbl2488
  <27.369999999999997-32.845> opd-fonts98/opd98ecbl2986
  <32.845-> opd-fonts98/opd98ecbl3583}{}
\pdfmapline{+opd98ecxc0500 SFXC0500 "0.98 ExtendFont T1Encoding ReEncodeFont" <cm-super-t1.enc <sfxc0500.pfb}
\pdfmapline{+opd98ecxc0600 SFXC0600 "0.98 ExtendFont T1Encoding ReEncodeFont" <cm-super-t1.enc <sfxc0600.pfb}
\pdfmapline{+opd98ecxc0700 SFXC0700 "0.98 ExtendFont T1Encoding ReEncodeFont" <cm-super-t1.enc <sfxc0700.pfb}
\pdfmapline{+opd98ecxc0800 SFXC0800 "0.98 ExtendFont T1Encoding ReEncodeFont" <cm-super-t1.enc <sfxc0800.pfb}
\pdfmapline{+opd98ecxc0900 SFXC0900 "0.98 ExtendFont T1Encoding ReEncodeFont" <cm-super-t1.enc <sfxc0900.pfb}
\pdfmapline{+opd98ecxc1000 SFXC1000 "0.98 ExtendFont T1Encoding ReEncodeFont" <cm-super-t1.enc <sfxc1000.pfb}
\pdfmapline{+opd98ecxc1095 SFXC1095 "0.98 ExtendFont T1Encoding ReEncodeFont" <cm-super-t1.enc <sfxc1095.pfb}
\pdfmapline{+opd98ecxc1200 SFXC1200 "0.98 ExtendFont T1Encoding ReEncodeFont" <cm-super-t1.enc <sfxc1200.pfb}
\pdfmapline{+opd98ecxc1440 SFXC1440 "0.98 ExtendFont T1Encoding ReEncodeFont" <cm-super-t1.enc <sfxc1440.pfb}
\pdfmapline{+opd98ecxc1728 SFXC1728 "0.98 ExtendFont T1Encoding ReEncodeFont" <cm-super-t1.enc <sfxc1728.pfb}
\pdfmapline{+opd98ecxc2074 SFXC2074 "0.98 ExtendFont T1Encoding ReEncodeFont" <cm-super-t1.enc <sfxc2074.pfb}
\pdfmapline{+opd98ecxc2488 SFXC2488 "0.98 ExtendFont T1Encoding ReEncodeFont" <cm-super-t1.enc <sfxc2488.pfb}
\pdfmapline{+opd98ecxc2986 SFXC2986 "0.98 ExtendFont T1Encoding ReEncodeFont" <cm-super-t1.enc <sfxc2986.pfb}
\pdfmapline{+opd98ecxc3583 SFXC3583 "0.98 ExtendFont T1Encoding ReEncodeFont" <cm-super-t1.enc <sfxc3583.pfb}
\DeclareFontShape{T1}{opdcmr}{bx}{sc}{<-5.5> opd-fonts98/opd98ecxc0500
  <5.5-6.5> opd-fonts98/opd98ecxc0600
  <6.5-7.5> opd-fonts98/opd98ecxc0700
  <7.5-8.5> opd-fonts98/opd98ecxc0800
  <8.5-9.5> opd-fonts98/opd98ecxc0900
  <9.5-10.475> opd-fonts98/opd98ecxc1000
  <10.475-11.475> opd-fonts98/opd98ecxc1095
  <11.475-13.2> opd-fonts98/opd98ecxc1200
  <13.2-15.84> opd-fonts98/opd98ecxc1440
  <15.84-19.009999999999998> opd-fonts98/opd98ecxc1728
  <19.009999999999998-22.81> opd-fonts98/opd98ecxc2074
  <22.81-27.369999999999997> opd-fonts98/opd98ecxc2488
  <27.369999999999997-32.845> opd-fonts98/opd98ecxc2986
  <32.845-> opd-fonts98/opd98ecxc3583}{}
\pdfmapline{+opd98ecui0500 SFUI0500 "0.98 ExtendFont T1Encoding ReEncodeFont" <cm-super-t1.enc <sfui0500.pfb}
\pdfmapline{+opd98ecui0600 SFUI0600 "0.98 ExtendFont T1Encoding ReEncodeFont" <cm-super-t1.enc <sfui0600.pfb}
\pdfmapline{+opd98ecui0700 SFUI0700 "0.98 ExtendFont T1Encoding ReEncodeFont" <cm-super-t1.enc <sfui0700.pfb}
\pdfmapline{+opd98ecui0800 SFUI0800 "0.98 ExtendFont T1Encoding ReEncodeFont" <cm-super-t1.enc <sfui0800.pfb}
\pdfmapline{+opd98ecui0900 SFUI0900 "0.98 ExtendFont T1Encoding ReEncodeFont" <cm-super-t1.enc <sfui0900.pfb}
\pdfmapline{+opd98ecui1000 SFUI1000 "0.98 ExtendFont T1Encoding ReEncodeFont" <cm-super-t1.enc <sfui1000.pfb}
\pdfmapline{+opd98ecui1095 SFUI1095 "0.98 ExtendFont T1Encoding ReEncodeFont" <cm-super-t1.enc <sfui1095.pfb}
\pdfmapline{+opd98ecui1200 SFUI1200 "0.98 ExtendFont T1Encoding ReEncodeFont" <cm-super-t1.enc <sfui1200.pfb}
\pdfmapline{+opd98ecui1440 SFUI1440 "0.98 ExtendFont T1Encoding ReEncodeFont" <cm-super-t1.enc <sfui1440.pfb}
\pdfmapline{+opd98ecui1728 SFUI1728 "0.98 ExtendFont T1Encoding ReEncodeFont" <cm-super-t1.enc <sfui1728.pfb}
\pdfmapline{+opd98ecui2074 SFUI2074 "0.98 ExtendFont T1Encoding ReEncodeFont" <cm-super-t1.enc <sfui2074.pfb}
\pdfmapline{+opd98ecui2488 SFUI2488 "0.98 ExtendFont T1Encoding ReEncodeFont" <cm-super-t1.enc <sfui2488.pfb}
\pdfmapline{+opd98ecui2986 SFUI2986 "0.98 ExtendFont T1Encoding ReEncodeFont" <cm-super-t1.enc <sfui2986.pfb}
\pdfmapline{+opd98ecui3583 SFUI3583 "0.98 ExtendFont T1Encoding ReEncodeFont" <cm-super-t1.enc <sfui3583.pfb}
\DeclareFontShape{T1}{opdcmr}{m}{ui}{<-5.5> opd-fonts98/opd98ecui0500
  <5.5-6.5> opd-fonts98/opd98ecui0600
  <6.5-7.5> opd-fonts98/opd98ecui0700
  <7.5-8.5> opd-fonts98/opd98ecui0800
  <8.5-9.5> opd-fonts98/opd98ecui0900
  <9.5-10.475> opd-fonts98/opd98ecui1000
  <10.475-11.475> opd-fonts98/opd98ecui1095
  <11.475-13.2> opd-fonts98/opd98ecui1200
  <13.2-15.84> opd-fonts98/opd98ecui1440
  <15.84-19.009999999999998> opd-fonts98/opd98ecui1728
  <19.009999999999998-22.81> opd-fonts98/opd98ecui2074
  <22.81-27.369999999999997> opd-fonts98/opd98ecui2488
  <27.369999999999997-32.845> opd-fonts98/opd98ecui2986
  <32.845-> opd-fonts98/opd98ecui3583}{}
\pdfmapline{+opd98ecsc0500 SFSC0500 "0.98 ExtendFont T1Encoding ReEncodeFont" <cm-super-t1.enc <sfsc0500.pfb}
\pdfmapline{+opd98ecsc0600 SFSC0600 "0.98 ExtendFont T1Encoding ReEncodeFont" <cm-super-t1.enc <sfsc0600.pfb}
\pdfmapline{+opd98ecsc0700 SFSC0700 "0.98 ExtendFont T1Encoding ReEncodeFont" <cm-super-t1.enc <sfsc0700.pfb}
\pdfmapline{+opd98ecsc0800 SFSC0800 "0.98 ExtendFont T1Encoding ReEncodeFont" <cm-super-t1.enc <sfsc0800.pfb}
\pdfmapline{+opd98ecsc0900 SFSC0900 "0.98 ExtendFont T1Encoding ReEncodeFont" <cm-super-t1.enc <sfsc0900.pfb}
\pdfmapline{+opd98ecsc1000 SFSC1000 "0.98 ExtendFont T1Encoding ReEncodeFont" <cm-super-t1.enc <sfsc1000.pfb}
\pdfmapline{+opd98ecsc1095 SFSC1095 "0.98 ExtendFont T1Encoding ReEncodeFont" <cm-super-t1.enc <sfsc1095.pfb}
\pdfmapline{+opd98ecsc1200 SFSC1200 "0.98 ExtendFont T1Encoding ReEncodeFont" <cm-super-t1.enc <sfsc1200.pfb}
\pdfmapline{+opd98ecsc1440 SFSC1440 "0.98 ExtendFont T1Encoding ReEncodeFont" <cm-super-t1.enc <sfsc1440.pfb}
\pdfmapline{+opd98ecsc1728 SFSC1728 "0.98 ExtendFont T1Encoding ReEncodeFont" <cm-super-t1.enc <sfsc1728.pfb}
\pdfmapline{+opd98ecsc2074 SFSC2074 "0.98 ExtendFont T1Encoding ReEncodeFont" <cm-super-t1.enc <sfsc2074.pfb}
\pdfmapline{+opd98ecsc2488 SFSC2488 "0.98 ExtendFont T1Encoding ReEncodeFont" <cm-super-t1.enc <sfsc2488.pfb}
\pdfmapline{+opd98ecsc2986 SFSC2986 "0.98 ExtendFont T1Encoding ReEncodeFont" <cm-super-t1.enc <sfsc2986.pfb}
\pdfmapline{+opd98ecsc3583 SFSC3583 "0.98 ExtendFont T1Encoding ReEncodeFont" <cm-super-t1.enc <sfsc3583.pfb}
\DeclareFontShape{T1}{opdcmr}{m}{scsl}{<-5.5> opd-fonts98/opd98ecsc0500
  <5.5-6.5> opd-fonts98/opd98ecsc0600
  <6.5-7.5> opd-fonts98/opd98ecsc0700
  <7.5-8.5> opd-fonts98/opd98ecsc0800
  <8.5-9.5> opd-fonts98/opd98ecsc0900
  <9.5-10.475> opd-fonts98/opd98ecsc1000
  <10.475-11.475> opd-fonts98/opd98ecsc1095
  <11.475-13.2> opd-fonts98/opd98ecsc1200
  <13.2-15.84> opd-fonts98/opd98ecsc1440
  <15.84-19.009999999999998> opd-fonts98/opd98ecsc1728
  <19.009999999999998-22.81> opd-fonts98/opd98ecsc2074
  <22.81-27.369999999999997> opd-fonts98/opd98ecsc2488
  <27.369999999999997-32.845> opd-fonts98/opd98ecsc2986
  <32.845-> opd-fonts98/opd98ecsc3583}{}
\pdfmapline{+opd98ecoc0500 SFOC0500 "0.98 ExtendFont T1Encoding ReEncodeFont" <cm-super-t1.enc <sfoc0500.pfb}
\pdfmapline{+opd98ecoc0600 SFOC0600 "0.98 ExtendFont T1Encoding ReEncodeFont" <cm-super-t1.enc <sfoc0600.pfb}
\pdfmapline{+opd98ecoc0700 SFOC0700 "0.98 ExtendFont T1Encoding ReEncodeFont" <cm-super-t1.enc <sfoc0700.pfb}
\pdfmapline{+opd98ecoc0800 SFOC0800 "0.98 ExtendFont T1Encoding ReEncodeFont" <cm-super-t1.enc <sfoc0800.pfb}
\pdfmapline{+opd98ecoc0900 SFOC0900 "0.98 ExtendFont T1Encoding ReEncodeFont" <cm-super-t1.enc <sfoc0900.pfb}
\pdfmapline{+opd98ecoc1000 SFOC1000 "0.98 ExtendFont T1Encoding ReEncodeFont" <cm-super-t1.enc <sfoc1000.pfb}
\pdfmapline{+opd98ecoc1095 SFOC1095 "0.98 ExtendFont T1Encoding ReEncodeFont" <cm-super-t1.enc <sfoc1095.pfb}
\pdfmapline{+opd98ecoc1200 SFOC1200 "0.98 ExtendFont T1Encoding ReEncodeFont" <cm-super-t1.enc <sfoc1200.pfb}
\pdfmapline{+opd98ecoc1440 SFOC1440 "0.98 ExtendFont T1Encoding ReEncodeFont" <cm-super-t1.enc <sfoc1440.pfb}
\pdfmapline{+opd98ecoc1728 SFOC1728 "0.98 ExtendFont T1Encoding ReEncodeFont" <cm-super-t1.enc <sfoc1728.pfb}
\pdfmapline{+opd98ecoc2074 SFOC2074 "0.98 ExtendFont T1Encoding ReEncodeFont" <cm-super-t1.enc <sfoc2074.pfb}
\pdfmapline{+opd98ecoc2488 SFOC2488 "0.98 ExtendFont T1Encoding ReEncodeFont" <cm-super-t1.enc <sfoc2488.pfb}
\pdfmapline{+opd98ecoc2986 SFOC2986 "0.98 ExtendFont T1Encoding ReEncodeFont" <cm-super-t1.enc <sfoc2986.pfb}
\pdfmapline{+opd98ecoc3583 SFOC3583 "0.98 ExtendFont T1Encoding ReEncodeFont" <cm-super-t1.enc <sfoc3583.pfb}
\DeclareFontShape{T1}{opdcmr}{bx}{scsl}{<-5.5> opd-fonts98/opd98ecoc0500
  <5.5-6.5> opd-fonts98/opd98ecoc0600
  <6.5-7.5> opd-fonts98/opd98ecoc0700
  <7.5-8.5> opd-fonts98/opd98ecoc0800
  <8.5-9.5> opd-fonts98/opd98ecoc0900
  <9.5-10.475> opd-fonts98/opd98ecoc1000
  <10.475-11.475> opd-fonts98/opd98ecoc1095
  <11.475-13.2> opd-fonts98/opd98ecoc1200
  <13.2-15.84> opd-fonts98/opd98ecoc1440
  <15.84-19.009999999999998> opd-fonts98/opd98ecoc1728
  <19.009999999999998-22.81> opd-fonts98/opd98ecoc2074
  <22.81-27.369999999999997> opd-fonts98/opd98ecoc2488
  <27.369999999999997-32.845> opd-fonts98/opd98ecoc2986
  <32.845-> opd-fonts98/opd98ecoc3583}{}
\DeclareFontShape{T1}{opdcmr}{b}{scsl}{<->ssub*opdcmr/bx/scsl}{}
\DeclareMicrotypeAlias{opdcmr}{cmr}

\pdfmapline{+opd98bytesans ByteSans-Bold-Width98 <T1-WGL4.enc <opd-fonts98/ByteSans-Bold-Width98.ttf}
\DeclareFontFamily{T1}{opdbytesans}{}
\DeclareFontShape{T1}{opdbytesans}{m}{n}{<-> opd-fonts98/opd98bytesans}{}
\DeclareFontShape{T1}{opdbytesans}{b}{n}{<-> opd-fonts98/opd98bytesans}{}
\DeclareFontShape{T1}{opdbytesans}{bx}{n}{<-> opd-fonts98/opd98bytesans}{}
\pdfmapline{+opd98bytesansmedium ByteSans-Medium-Width98 <T1-WGL4.enc <opd-fonts98/ByteSans-Medium-Width98.ttf}
\DeclareFontShape{T1}{bytesansmedium}{m}{n}{<-> opd-fonts98/opd98bytesansmedium}{}
\DeclareFontShape{T1}{bytesansmedium}{b}{n}{<-> opd-fonts98/opd98bytesansmedium}{}
\DeclareFontShape{T1}{bytesansmedium}{bx}{n}{<-> opd-fonts98/opd98bytesansmedium}{}

\normalfont

\AtBeginDocument{\clearcaptionsetup{sub}}
\definecolor{YaleBlue}{HTML}{2A5487}
\usepackage{enumitem,tabularx,colortbl,pifont,needspace}
\setlist[itemize]{noitemsep}
\setlist[enumerate]{noitemsep}

\usepackage[all]{hypcap}
\usepackage{mathtools,mathrsfs,nicefrac,dsfont}
\usepackage{booktabs,multirow,makecell}
\usepackage{float,subcaption,adjustbox,rotating,stackengine}
\fancypagestyle{firststyle}{
  \fancyhead[R]{}
  \fancyhead[L]{\vskip 4mm
    \clipbox{0pt 0pt 17.8mm 0pt}{\includegraphics[width=53.3mm,height=6mm]{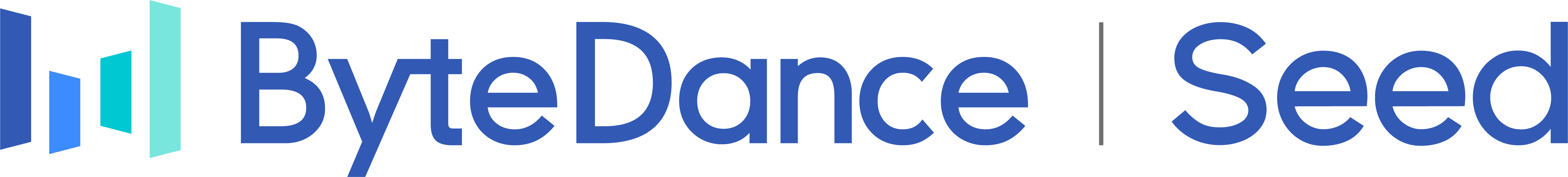}}}
}

\usepackage{wrapfig}
\usepackage[section]{placeins}
\usepackage{cleveref}
\crefname{section}{\S}{\S\S}
\Crefname{section}{\S}{\S\S}
\crefname{subsection}{\S}{\S\S}
\Crefname{subsection}{\S}{\S\S}
\crefname{subsubsection}{\S}{\S\S}
\Crefname{subsubsection}{\S}{\S\S}
\newcommand{\batch}{B}

\definecolor{blanchedalmond}{rgb}{1.0, 0.92, 0.8}
\definecolor{carmine}{rgb}{0.59, 0.0, 0.09}
\definecolor{lightblue}{rgb}{0.22,0.45,0.70}%

\renewcommand{\mathbf}{\boldsymbol}

\makeatletter
\def\Ddots{\mathinner{\mkern1mu\raise\p@
\vbox{\kern7\p@\hbox{.}}\mkern2mu
\raise4\p@\hbox{.}\mkern2mu\raise7\p@\hbox{.}\mkern1mu}}
\makeatother

\definecolor{amaranth}{rgb}{0.9, 0.17, 0.31}
\definecolor{antiquebrass}{rgb}{0.8, 0.58, 0.46}
\definecolor{antiquefuchsia}{rgb}{0.57, 0.36, 0.51}
\definecolor{chromeyellow}{rgb}{0.31, 0.47, 0.26}

\renewcommand{\paragraph}[1]{\textbf{#1}}

\title{On the Generalization Limitation of On-Policy Distillation for Post-RL Consolidation}
\author[1,*]{Yichen Wang}
\author[2]{Yufa Zhou}
\author[3]{Siqi Zhu}
\author[4]{Haozheng Luo}
\author[3]{Jingzhi Bao}
\author[5]{Zhichao Liu}
\author[5,\dagger]{Zijun Wei}
\affiliation[1]{University of Chicago}
\affiliation[2]{Duke University}
\affiliation[3]{UIUC}
\affiliation[4]{Northwestern University}
\affiliation[5]{ByteDance}
\contribution[*]{This work was conducted at ByteDance.}
\contribution[\dagger]{Corresponding author.}
\checkdata[Project page]{\url{https://yichenzw.com/opd-generalization.github.io/}}

\abstract{
Multi-teacher on-policy distillation (MOPD) extends on-policy distillation (OPD) to consolidate capabilities from multiple teachers. Although prior work reports near-perfect in-distribution (IND) distillation performance, we find that this can mask incomplete capability consolidation beyond the supervised distribution.
A preliminary study on synthetic reasoning shows that OPD does not fully transfer teachers’ generalization capabilities.
We then extend this investigation to real-world tasks, evaluating extrapolative and compositional generalization across mathematical reasoning, coding, agent, and long-context capabilities using Qwen3 models.
Across extrapolative settings, OPD and MOPD reach 96.04-102.73\% of their teachers' IND scores.
Yet their out-of-distribution (OOD) scores remain 2.45-16.25\% below the base model's.
Across three compositional settings, MOPD improves  IND performance over the base model, but its performance on composed tasks as OOD remains 18.19–24.34\% below that of the best teachers.
To understand these gaps, we examine teacher capability and data coverage as two potential bottlenecks.
Jointly alleviating these bottlenecks eliminates OOD forgetting in our experiments and suggests the potential for generalization beyond the teacher.
(M)OPD also reduces OOD forgetting on average relative to its reinforcement learning (RL) baselines.
Analysis of the associated update geometry reveals smaller and more spectrally concentrated updates on average than those of the RL baselines, along with greater shared update structure under MOPD.
Motivated by these findings, we propose geometry-based patching and base-preserving replay to improve OOD performance while largely preserving IND performance. 
Our findings identify a blind spot in capability consolidation via (M)OPD, highlighting the need for algorithms that explicitly support OOD retention and generalization.
}

\begin{document}
\maketitle
\section{Introduction}
\label{sec:intro}

On-policy distillation (OPD) trains a student model on its own generated trajectories using supervision from a teacher~\citep{agarwal2024onpolicydistillationlanguagemodels,lu2025onpolicydistillation}. Its extension, multi-teacher on-policy distillation (MOPD), offers a way to consolidate capabilities from separately trained teachers into one model~\citep{ma2026mopdmultiteacheronpolicydistillation}. This approach is used as a consolidation stage in large-scale post-training pipelines to combine domain-specialized teachers~\citep{coreteam2026mimov2flashtechnicalreport,deepseekai2026deepseekv4highlyefficientmilliontoken,nvidia2026nemotron3ultraopen,kimiteam2026kimik3openfrontier}, achieving strong distillation performance.

The intended outcome of post-RL consolidation is to learn the teachers' improvements while retaining the base model's capabilities and supporting their use beyond the supervised tasks. We find that achieving near-perfect in-distribution (IND) distillation does not reliably yield either out-of-distribution (OOD) retention or OOD generalization beyond the base. \textbf{Near-perfect IND distillation can therefore mask incomplete capability consolidation.}

We start from a preliminary study on a synthetic reasoning task (\Cref{sec:preliminary}, ~\citet{power2022grokking,zhang2025interplaypretrainingmidtrainingrl}) to illustrate the generalization limitation of OPD. 
The self-RL teacher and OPD student start from a common pretrained base. When the teacher generalizes beyond the supervised range, OPD does not fully transfer that generalization despite strong IND performance. This motivates testing whether the same limitation persists when consolidating capabilities on real-world tasks, where useful behavior can extend far beyond narrow distillation data and require distinct capabilities together.

Our main experiments use Qwen3-based models~\citep{yang2025qwen3technicalreport} to study two forms of generalization across mathematical reasoning, coding, agent, and long-context capabilities (\Cref{sec:setup}). \emph{Extrapolative generalization} tests performance beyond the supervised range within a capability~\citep{anil2022lengthgeneralization,zhang2025interplaypretrainingmidtrainingrl}, using mathematical reasoning benchmarks of differing difficulty. \emph{Compositional generalization} tests whether separately supervised capabilities can be used together~\citep{lake2018systematicity,yuan2026f}.
We also distinguish two teacher setups: self-RL specialist teachers are independently trained from a common base and then distilled into a student initialized from the same base, reflecting specialization followed by consolidation; external generalist teachers are obtained separately and used to distill stronger capabilities into a smaller student (\Cref{sec:teachers}).
First, across both teacher setups, near-perfect IND distillation coexists with OOD forgetting (\Cref{sec:extrapolative}). Across individual training datasets and mixtures, OPD and MOPD reach 96.04-102.73\% of their teachers' mean IND scores, yet their mean OOD scores remain 2.45-16.25\% below the base. 
The gap persists even with a stronger external generalist teacher that outperforms the base model on OOD evaluations.

Second, component improvements do not ensure compositional generalization (\Cref{sec:compositional}). Across the three settings, MOPD improves mean IND performance by 0.48-22.46\% over the base but remains 18.19-24.34\% below the best teachers on composed tasks. 
For example, in code plus long context, MOPD improves all three component benchmarks, yet on the composed task (long-context coding) it is still at about the base's performance level, below both teachers. 
Improvements on separately supervised component tasks therefore need not translate into gains on tasks requiring those capabilities to be combined.

To understand these gaps, we examine teacher capability and data coverage as two potential bottlenecks to OOD performance.
Jointly alleviating these bottlenecks eliminates OOD forgetting in our experiments and suggests the potential for generalization beyond the teacher (\Cref{sec:coverage}).

Despite the remaining gaps, our main experiments also show that OPD and MOPD reduce average OOD forgetting relative to the self-RL teachers and mixed RL, respectively. 
To examine the associated parameter changes, we draw on analyses of multi-task training and OPD geometry~\citep{zhu2026sft,shen2026geometryonpolicydistillation}. 
(M)OPD produces smaller and more spectrally concentrated updates on average than their corresponding RL baselines.
Moreover, across both generalization settings, MOPD updates are better explained by the span of single-teacher OPD updates than mixed-RL updates are by the span of single-RL updates (\Cref{sec:geometry}).
Appendix~\ref{app:opd-rl-theory} provides a complementary local theoretical analysis.

Motivated by the observed update structure, we investigate geometry-based patching after distillation.
Patching partially closes the OOD gap on AIME25 while retaining most of the teacher's IND performance.
We also investigate base-preserving replay to explicitly preserve the base model’s behavior during distillation (\Cref{sec:retention}). 
Relative to plain OPD, replay improves scores on five of eight OOD benchmarks with only a small decrease in IND performance, and surpasses both the student base and RL teacher on AIME25.

Overall, near-perfect IND distillation can mask incomplete capability consolidation, motivating explicit support for OOD retention and generalization.

\section{Related Work}
\label{sec:related-work}

\subsection{On-Policy Distillation and Multi-teacher On-Policy Distillation}
\label{sec:related-opd}

On-policy distillation (OPD) trains a student on trajectories sampled from its own policy, using the teacher's next-token predictions as supervision~\citep{agarwal2024onpolicydistillationlanguagemodels,lu2025onpolicydistillation}.
It combines the nature of on-policy state distribution, like reinforcement learning with verifiable rewards (RLVR), with the dense token-level supervision signal of knowledge distillation~\citep{shao2024deepseekmathpushinglimitsmathematical,hinton2015distilling}.
Prior work reports strong distillation of teacher performance and rapid convergence in mathematical reasoning, knowledge question answering~\citep{lu2025onpolicydistillation}, and multi-turn agentic tasks~\citep{liao2026prefixreplay,wu2026seed}.
%
Multi-teacher on-policy distillation (MOPD) extends this approach by assigning training domains to teachers with corresponding capabilities~\citep{ma2026mopdmultiteacheronpolicydistillation}.
It is increasingly used for post-RL consolidation in frontier model development: domain experts are trained separately and then distilled into a single student~\citep{coreteam2026mimov2flashtechnicalreport,deepseekai2026deepseekv4highlyefficientmilliontoken,nvidia2026nemotron3ultraopen,kimiteam2026kimik3openfrontier}, showing great distillation performance.
%
 Much of the methodological work \citep{wu2026seed,xing2026trustregion,jiang2026trajectoryrefined} aims to improve target-task performance in domain.
 However, we suggest that this near-perfect distillation performance can be limited in IND evaluation. We complement these studies by focusing on OOD performance, suggesting that it can conceal incomplete generalization.
Two recent studies are close to ours but different.
\citet{gao2026openmopd} diagnose capability conflict and imbalance in MOPD, but the capabilities are all within the training distribution (i.e., IND).
\citet{li2026everycoin} examine generalization across domains (e.g., from math to science) or across languages, which differs from our generalization forms; it is more about the closeness of different tasks but not the algorithm's potential. 


\subsection{Generalization in Post-Training}
\label{sec:related-generalization}

Generalization is a model's ability to perform well on examples out of the training distribution.
Grokking demonstrates that generalization on algorithmic datasets emerges long after a model fits the training data~\citep{power2022grokking}.
Studies on modular arithmetic also show that models can generalize to tasks requiring skill composition~\citep{he2024learningtogrok}.
%
In LLM post-training, RLVR can improve reasoning on the training domain while producing inconsistent gains on unseen domains~\citep{hu2026breakingbarriers, zhang2025interplaypretrainingmidtrainingrl, fu2026compositional, yuan2026f}.
These findings motivate treating rapid IND convergence and OOD generalization as separate empirical questions.
Although OPD can approach teacher performance within relatively few updates~\citep{lu2025onpolicydistillation}, such convergence alone does not establish whether the student inherits the teacher's generalization.
We hence study whether (M)OPD transfers performance beyond the supervised distribution.

%
%
%
%
%
%
%
%
%
%
%
\subsection{Capability Consolidation}
\label{sec:related-consolidation}

Capability consolidation aims to combine multiple capabilities in a single model, often at the final stage of post-training.
Established approaches include model merging and mixed RL.
Model merging combines separately trained models directly in parameter space~\citep{ainsworth2023gitrebasin,ilharco2023taskarithmetic,yadav2023tiesmerging,yu2024dare,yang2026orthogonal}.
Mixed RL jointly optimizes a single model on data from multiple tasks using their corresponding rewards~\citep{lambert2024tulu3,yang2025qwen3technicalreport}.
On another thread, related work studies relationships among capabilities, for example, by pairwise transfer and latent factors in transfer matrices~\citep{kim2023taskweb,krishna2025latenttraits} or by using sparse autoencoder features~\citep{deng2026qwenscope}.
Our evaluations ask whether consolidation preserves performance beyond the supervised distribution, generalizing to tasks requiring multiple capabilities together.

\section{Experimental Setting}
\label{sec:setup}

\subsection{Teachers and Consolidation Methods}
\label{sec:teachers}

In the main experiments, we initialize students from Qwen3-8B~\citep{yang2025qwen3technicalreport}, with base parameters $\theta_0$.

\emph{Self-RL specialist teachers} are trained independently from this base on separate distributions~\citep{ma2026mopdmultiteacheronpolicydistillation,coreteam2026mimov2flashtechnicalreport}. This follows a post-RL pipeline that specializes individual capabilities before consolidating them into one student, often when directly advancing a frontier model's Pareto frontier across capabilities is difficult. Specialist teachers primarily improve their target capabilities but may exhibit forgetting on others~\citep{glm5team2026glm5vibecodingagentic}.

\emph{External generalist teachers} are obtained from sources outside the student's training pipeline. These teachers are often generalists that outperform the student not only on the target capability (aligned with the training distribution) but also on other capabilities.
This mirrors strong-to-weak distillation scenarios, such as distilling from a larger model into a smaller one~\citep{yang2025qwen3technicalreport,lu2025onpolicydistillation}.


Let $\pi_\theta$ be the trainable model. For $i=0,\ldots,K-1$, let $\mathcal{D}_i$ denote a prompt distribution, $r_i(x,y)$ its task reward, and $\pi_i^*$ its fixed teacher. For response $y=(y_1,\ldots,y_T)$, write $h_t=(x,y_{<t})$. Mixture weights satisfy $\alpha_i\geq0$ and $\sum_{i=0}^{K-1}\alpha_i=1$.
We consider the following training pipelines:

\paragraph{Single-task RL.}
Each self-RL teacher is trained from $\theta_0$ with group relative policy optimization (GRPO)~\citep{shao2024deepseekmathpushinglimitsmathematical} to maximize expected reward on one task distribution, $\mathcal{J}_{\mathrm{RL}}^{(i)}(\theta)=\mathbb{E}_{x\sim\mathcal{D}_i,\,y\sim\pi_\theta(\cdot\mid x)}[r_i(x,y)]$.

\paragraph{OPD.}
OPD minimizes student-to-teacher reverse KL at each token position along student-generated trajectories~\citep{agarwal2024onpolicydistillationlanguagemodels,lu2025onpolicydistillation}. By default, we use a single task distribution and its corresponding teacher:
\begin{equation}
\mathcal{J}_{\mathrm{OPD}}(\theta;\mathcal{D},\pi^*)
=\mathbb{E}_{x\sim\mathcal{D},\;y\sim\pi_\theta(\cdot\mid x)}
\left[\sum_{t=1}^{T}
D_{\mathrm{KL}}\!\left(\pi_\theta(\cdot\mid h_t)\,\|\,\pi^*(\cdot\mid h_t)\right)\right].
\label{eq:opd-objective}
\end{equation}

\paragraph{MOPD.}
MOPD minimizes a mixture of OPD objectives on different tasks and teachers:
\begin{equation}
\mathcal{J}_{\mathrm{MOPD}}(\theta)
=\sum_{i=0}^{K-1}\alpha_i\,\mathcal{J}_{\mathrm{OPD}}(\theta;\mathcal{D}_i,\pi_i^*).
\label{eq:mopd-objective}
\end{equation}
Each data source uses its corresponding self-RL teacher or the strongest evaluated external teacher for that domain; teacher assignments remain fixed across steps and epochs in each training run.

\paragraph{Mixed RL.}
In our setting, mixed RL serves as a comparison method for MOPD. It maximizes $\mathcal{J}_{\mathrm{mixed\text{-}RL}}(\theta)=\sum_{i=0}^{K-1}\alpha_i\mathcal{J}_{\mathrm{RL}}^{(i)}(\theta)$ using the same single-task RL recipes, with shuffled examples from all selected sources. Teacher selection, implementation details, and parameter-update notation are collected in \Cref{app:training-protocols}.

\subsection{IND Performance and Two Forms of Generalization}
\label{sec:evaluation}

We define the \emph{in-distribution (IND)} evaluations relative to the distributions directly supervised during distillation. \emph{Out-of-distribution (OOD)} evaluations probe distributions or combinations of capabilities outside that supervision. In multi-teacher experiments, the supervised distribution includes the data supplied by all teachers.

For the cross-benchmark mathematical aggregates, we operationalize this grouping using the self-RL teachers' performance relative to the base; the criterion and its limitations are given in \Cref{app:operational-ind-ood}.

\paragraph{Extrapolative generalization.}
We evaluate performance beyond the range covered by the supervised distribution, such as more complex problems within the same domain.
We separately measure \emph{OOD retention}, meaning maintenance of the base model's OOD performance, and \emph{OOD generalization}, meaning improvement beyond the base on OOD evaluations. Incomplete transfer of a teacher's positive OOD improvement and a student's loss relative to the base are distinct possible outcomes.

\paragraph{Compositional generalization.}
We train on data for separate component capabilities and evaluate on tasks requiring their combination. We compare the student with the base and both teachers. Improvement over the base establishes a benefit on the composed task; surpassing both teachers would provide stronger evidence that consolidation yields an additional benefit from combining their capabilities. We report component and composed-task evaluations separately.


\subsection{Training Data and Evaluation Benchmarks}
\label{sec:data-benchmarks}

For mathematical reasoning, we train on GSM8K~\citep{cobbe2021trainingverifierssolvemath}, MATH~\citep{hendrycks2021measuringmathematicalproblemsolving}, and DAPO~\citep{yu2025dapoopensourcellmreinforcement}, and evaluate on GSM8K, MATH500~\citep{lightman2023lets}, AMC23~\citep{aimo2024amc}, and AIME24/25~\citep{huggingfaceh42024aime,opencompass2025aime}.
For coding, we train on Nemotron competitive-coding data~\citep{nvidia2025nemotronrlcoding} and evaluate on HumanEval~\citep{chen2021evaluatingcode}, MBPP~\citep{austin2021programsynthesis}, and their EvalPlus extensions~\citep{liu2023evalplus}.
For agent capabilities, we train on synthetic BFCL live/non-live data~\citep{patil2025bfcl}, xLAM60K~\citep{liu2024apigen}, and ToolACE~\citep{liu2024toolace}, and evaluate primarily on BFCL V4's non-live and live subsets.
For long context, we train on DocQA-RL-1.6K (DocQA)~\citep{wan2025qwenlongl1} and evaluate on LongBench-v2~\citep{bai2024longbenchv2}, primarily its Hard subset.
We mainly probe extrapolative generalization through mathematical reasoning, using shifts from narrow training distributions such as GSM8K to harder benchmarks such as AIME24/25. This follows prior work evaluating mathematical reasoning beyond the training difficulty~\citep{hu2026breakingbarriers,li2026everycoin,motwani2025h1,cui2026clipo,lee2026beyondcorrectness}.
We test compositional generalization through long-context math reasoning on GSM-Infinite~\citep{zhou2025gsminfinite} and DocMath~\citep{zhao2023docmath}, long-context coding on LongBench-v2 Code, and agentic coding on SWE-bench Verified~\citep{jimenez2024swebench}.
Per-run data assignments, evaluation subsets, and scoring protocols are detailed in \Cref{app:datasets,app:scoring}.

\section{Preliminary Study: Strong IND Performance Leaves an OOD Gap}
\label{sec:preliminary}

OPD can achieve near-perfect IND distillation within relatively few training steps~\citep{lu2025onpolicydistillation,ma2026mopdmultiteacheronpolicydistillation}. Yet grokking shows that generalization can emerge long after a model fits its training data~\citep{power2022grokking}. This motivates asking whether rapid IND convergence in OPD is accompanied by transfer of the teacher's generalization. We examine this question in a controlled synthetic mathematical reasoning study.

\Needspace{0.52\textheight}
\begin{wrapfigure}{r}{0.49\linewidth}
\centering
\includegraphics[width=\linewidth]{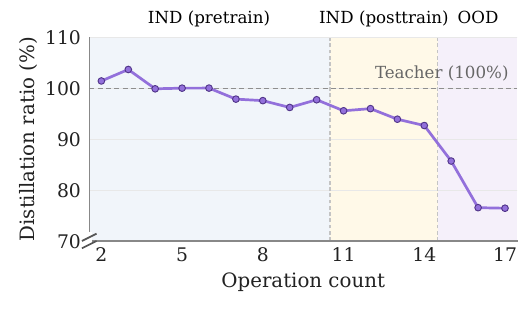}
\caption{\textbf{Near-perfect IND distillation leaves an OOD transfer gap.} The distillation ratio is $\frac{s_{\mathrm{student}}-s_{\mathrm{base}}}{s_{\mathrm{teacher}}-s_{\mathrm{base}}}$, where $s$ is acc@1 at each operation count. Pretraining IND covers 2-10 operations, post-training IND covers 11-14, and OOD covers 15-17. The dashed line marks complete transfer of the teacher's improvement over the base.}
\label{fig:preliminary}
\end{wrapfigure}

\textbf{Settings.} Following the framework of \citet{zhang2025interplaypretrainingmidtrainingrl}, we pretrain a randomly initialized Qwen2-100M model on GSM-Infinite-based problems~\citep{zhou2025gsminfinite} requiring 2-10 operations to obtain the base checkpoint $\pi_0$. 
We then train a self-RL teacher from $\pi_0$ on problems requiring 11-14 operations.
The OPD student starts from the same base $\pi_0$ and learns from this teacher on the same post-training data distribution. 
In this controlled study, we report the pretraining range (2-10 operations) and post-training range (11-14) separately as IND evaluations; 15-17 operations test OOD extrapolation beyond both ranges. 
Operation count provides an explicit axis of difficulty before we turn to post-training experiments on real-world tasks.

\textbf{Findings.} \Cref{fig:preliminary} shows near-perfect IND distillation alongside incomplete OOD transfer. While the distillation ratio of OPD is approximately 96-104\% on pretraining IND and 93-96\% on post-training IND, it falls to 76-86\% on OOD problems. 
Thus, \textbf{when the teacher generalizes beyond the supervised range, OPD does not fully transfer that generalization despite strong IND performance.} 
This motivates testing whether the same limitation persists when consolidating capabilities on real-world tasks, where useful behavior can extend far beyond narrow distillation data and require distinct capabilities together.


\section{Strong IND Performance Does Not Ensure Extrapolative Generalization}
\label{sec:extrapolative}

We examine whether near-perfect IND distillation yields extrapolative generalization in mathematical reasoning, using self-RL specialist teachers and external generalist teachers.

\paragraph{Overall findings.}
Across both teacher setups, \textbf{near-perfect IND distillation coexists with OOD forgetting}. OPD and MOPD reduce average forgetting relative to the self-RL teachers and mixed RL, respectively, but still fall below the base on OOD evaluations.

\subsection{Self-RL Specialist Teachers}
\label{sec:self-rl}

\begin{figure}[!htbp]
\centering
\includegraphics[width=\linewidth]{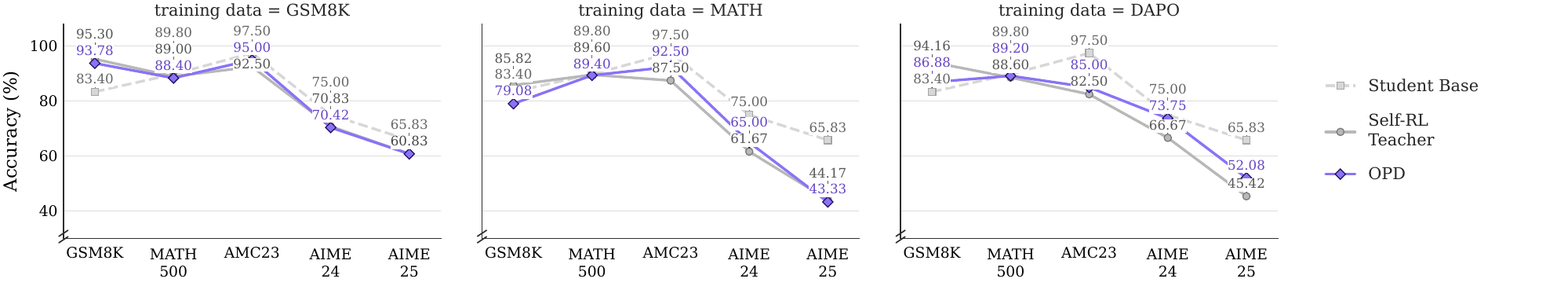}
\caption{\textbf{Near-perfect IND distillation does not ensure OOD retention.} Within each panel, benchmarks are arranged from IND on the left to OOD on the right. On most OOD evaluations, OPD performance (purple solid lines) lies between the base (gray dashed lines) and the corresponding teacher (gray solid lines), indicating reduced but persistent forgetting. In this experiment, teachers themselves also exhibit OOD forgetting, scoring below the base. \Cref{sec:external-generalist} further shows that OPD forgetting persists even with an external generalist teacher that outperforms the base on OOD. The self-RL specialists and their OPD students share a common base. AIME scores report avg@8.}
\label{fig:single-math}
\end{figure}

\paragraph{Single-teacher OPD.}
\textbf{OPD achieves near-perfect IND distillation}, reaching 97.10\% of its teachers' IND score on average across all three training datasets (\Cref{fig:single-math}).
Yet \textbf{its OOD performance decreases relative to the base} by 10.78\% (8.56 percentage points).
\footnote{For the cross-benchmark mathematical aggregates, we use an operational grouping: a benchmark is IND if each of the three self-RL teachers exceeds 97\% of the base score. This yields GSM8K and MATH500 as IND, and AMC23 and AIME24/25 as OOD. Thresholds of 95\%, 96\%, and 98\% give the same partition. These labels describe compatibility with the teachers' post-training regime rather than establish distributional membership or pretraining exposure; see \Cref{app:operational-ind-ood}.}
We find that the three self-RL teachers also exhibit OOD forgetting, similar results reported in \citet{li2026choicedivergence,phan2026beyondreasoninggains}, which are more severe than the corresponding OPD.
\textbf{OPD reduces average OOD forgetting by 25.09\% relative to its self-RL teachers}: the mean loss from the base falls from 11.43 to 8.56 percentage points.
Overall, near-perfect IND distillation does not ensure OOD improvement over the base.
\Cref{sec:geometry} examines the parameter-update differences accompanying OPD's milder forgetting.
This leaves open whether the student's OOD forgetting is inherited from teacher-side forgetting. \Cref{sec:external-generalist} tests whether the gaps persist with external generalist teachers that provide stronger OOD performance. 
Additional evaluations on coding benchmarks show that forgetting also extends beyond mathematical reasoning (\Cref{app:distant-coding,fig:distant-coding}).

\begin{figure}[t]
\centering
\includegraphics[width=\linewidth]{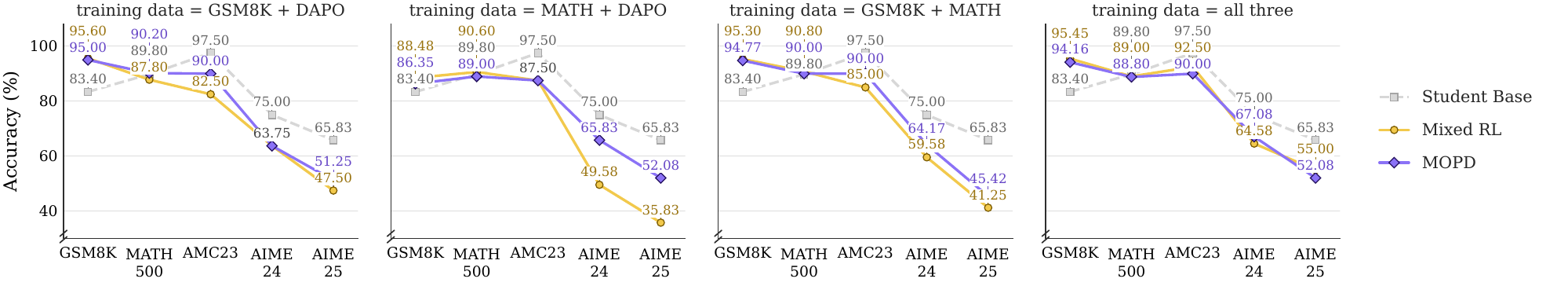}
\caption{\textbf{MOPD reduces average OOD forgetting relative to mixed RL but remains below the base.} Results cover all pairwise and three-teacher mixtures; AIME scores report avg@8.}
\label{fig:multi-math}
\end{figure}

\paragraph{MOPD and mixed RL.}
MOPD and mixed RL show the same pattern as single-teacher OPD.
Both improve average IND performance over the base but forget on OOD evaluations (\Cref{fig:multi-math}).
Across all three pairwise mixtures and the three-teacher mixture, MOPD's average OOD performance remains 14.07\% (11.18 percentage points) below the base. Nevertheless, \textbf{MOPD reduces average OOD forgetting by 28.92\% relative to mixed RL}, whose mean loss from the base is 15.73 points.


\subsection{External Generalist Teachers}
\label{sec:external-generalist}

\begin{figure}[!htbp]
\centering
\includegraphics[width=\linewidth]{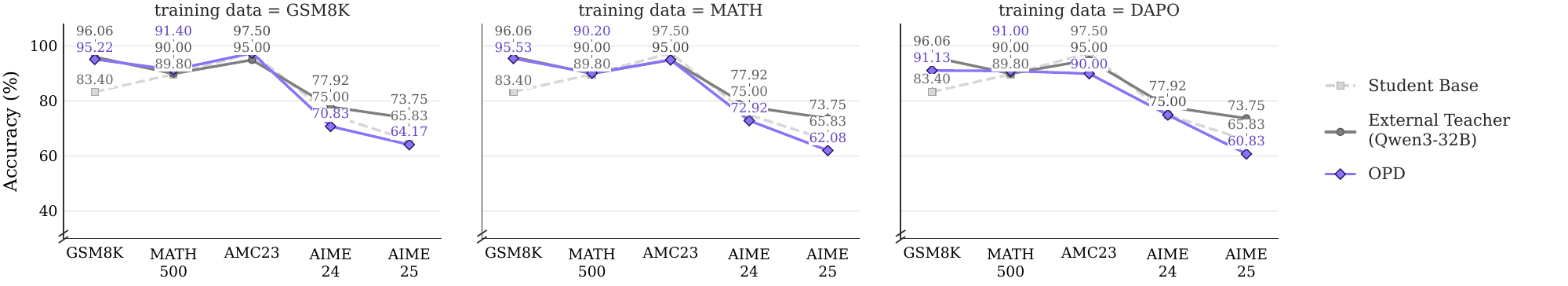}
\caption{\textbf{A stronger external teacher does not eliminate OOD gaps.} Qwen3-32B teaches Qwen3-8B; AIME scores report avg@8. All three figures use the same Qwen3-8B base scores.}
\label{fig:generalist-narrow}
\end{figure}

To test whether the gaps depend on teacher-side forgetting, we use an external generalist teacher that is not obtained by specializing the student base through our self-RL pipeline. 
Specifically, we distill Qwen3-32B into Qwen3-8B (\Cref{fig:generalist-narrow}).
The teacher model generally outperforms the student base across most evaluations.
This setup therefore makes stronger OOD performance available for transfer on OOD tasks like AIME.

Across the three distillation datasets, OPD achieves near-perfect IND distillation, reaching 99.34\% of the teacher's IND score (0.62 points below it), yet remains 3.73\% below the base on OOD evaluations (2.96 points). \textbf{Stronger teacher performance alone therefore does not ensure OOD transfer or retention under narrow supervision. }
\Cref{sec:coverage} further examines whether broader distillation data improves these outcomes.

\section{Component Improvements Do Not Ensure Compositional Generalization}
\label{sec:compositional}

\begin{figure}[!htbp]
\centering
\includegraphics[width=0.97\linewidth]{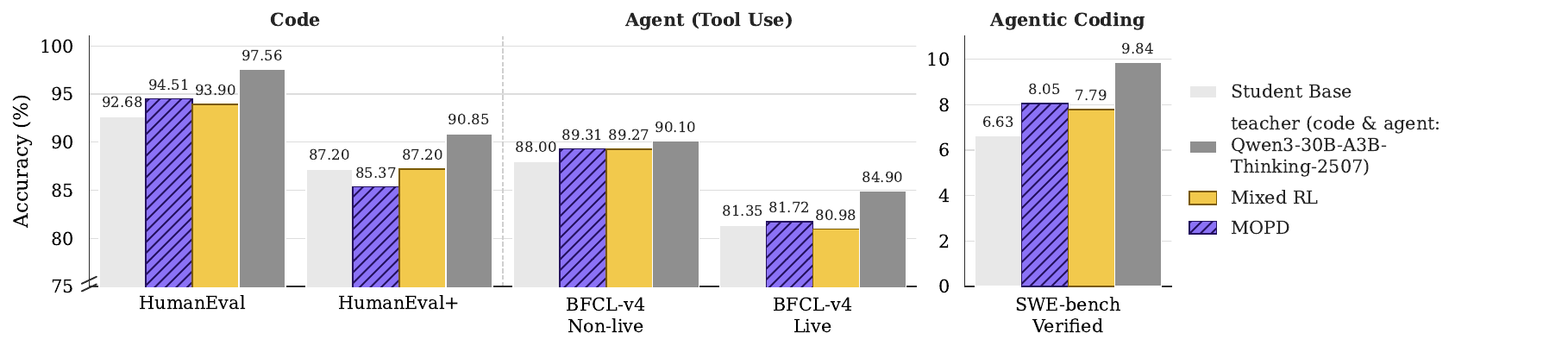}
\caption{\textbf{Compositional generalization to agentic coding}: MOPD reaches 96.6\% of the teacher's mean IND score but remains 18.2\% below the shared teacher OOD on SWE-bench Verified. IND scores average HumanEval, HumanEval+, and BFCL-v4 Non-live/Live. MOPD outperforms mixed RL on SWE-bench Verified and three of four IND evaluations.}
\label{fig:code-agent}
\end{figure}

\begin{figure}[!htbp]
\centering
\includegraphics[width=0.97\linewidth]{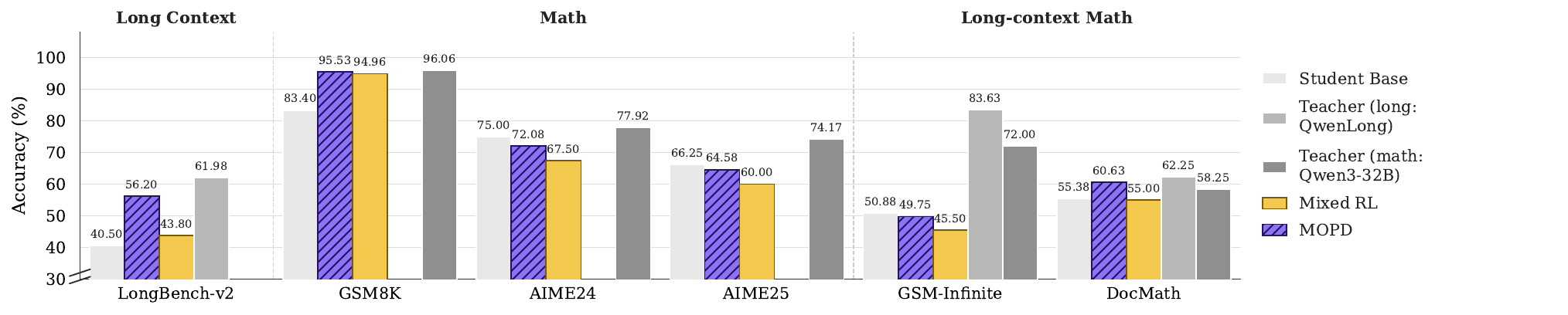}
\caption{\textbf{Compositional generalization to long-context math}: MOPD reaches 96.0\% of the teachers' mean IND score but remains 24.3\% below the best teacher on composed tasks. IND scores average LongBench-v2 and GSM8K; composed scores average GSM-Infinite and DocMath. MOPD outperforms mixed RL on all displayed benchmarks. GSM-Infinite uses the 8K mini800 subset; DocMath uses test-mini.}
\label{fig:math-long-context}
\end{figure}

\begin{figure}[!htbp]
\centering
\includegraphics[width=0.97\linewidth]{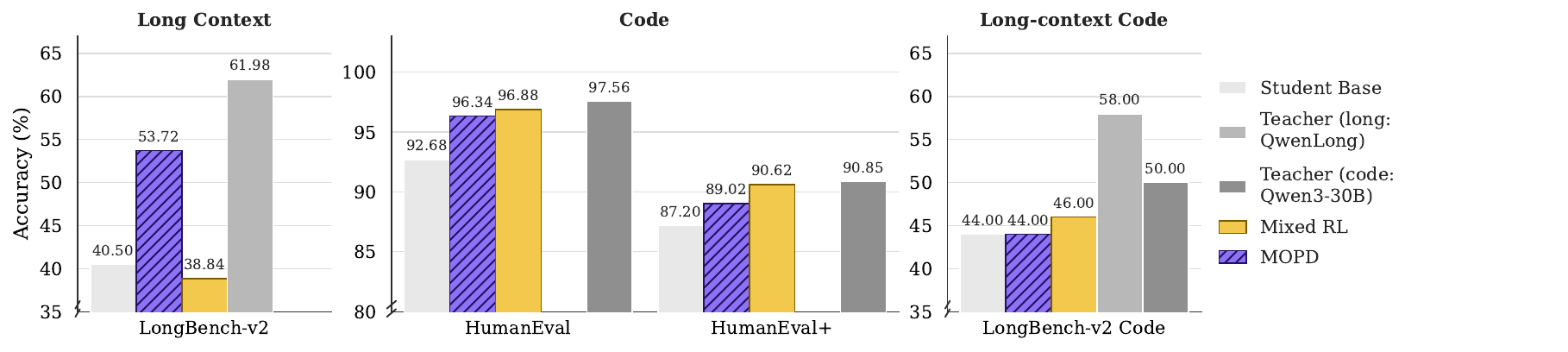}
\caption{\textbf{Compositional generalization to long-context code}: MOPD reaches 95.5\% of the teachers' mean IND score but remains 24.1\% below the best teacher on LongBench-v2 Code. IND scores average LongBench-v2, HumanEval, and HumanEval+. MOPD outperforms mixed RL on long context, but scores lower on coding and the composed task.}
\label{fig:code-long-context}
\end{figure}

We next ask whether MOPD can consolidate separate capabilities to benefit tasks requiring their combination, i.e., compositional generalization. Specifically, we study code plus agent, math plus long context, and code plus long context.
The desired outcome is a student that learns both components and surpasses its teacher or teachers on the composed task.
In the experiments, MOPD assigns an external teacher to each component's training data. 
For comparison, we construct mixed RL from single-task recipes, including training data, that achieve component performance comparable to MOPD. 
Both methods use the same fixed mixture proportions and random shuffling without further tuning to balance tasks for fair comparison.
\paragraph{Overall findings.}
Across the three settings, \textbf{MOPD improves IND performance by 9.24\% over the base (3.97\% below the corresponding teachers), but remains 23.82\% below the best teachers on composed tasks} (\Cref{fig:code-agent,fig:math-long-context,fig:code-long-context}). Mixed RL shows a similar compositional gap (26.10\%); MOPD's mean composed-task score is 3.08\% (1.07 percentage points) higher than mixed RL's.

\paragraph{Code and Agent.}
%
Coding and tool-use supervision improves SWE-bench Verified from the base's 6.63 to 8.05, above mixed RL (7.79) but below the shared teacher (9.84; \Cref{fig:code-agent}). 
MOPD also improves over the base on all component benchmarks except HumanEval+.

\paragraph{Math and Long Context.}
MOPD improves LongBench-v2 from 40.50 to 56.20 and GSM8K from 83.40 to 95.53, yet the composed-task outcomes diverge (\Cref{fig:math-long-context}). DocMath improves from 55.38 to 60.63, whereas GSM-Infinite falls from 50.88 to 49.75; both remain below the best teacher. MOPD nevertheless outperforms mixed RL on all displayed benchmarks.

\paragraph{Code and Long Context.}
%
MOPD improves all three component benchmarks, yet LongBench-v2 Code remains at the base's 44.00, below both teachers (50.00 and 58.00; \Cref{fig:code-long-context}). Mixed RL raises this composed-task score to 46.00 but lowers the LongBench-v2 component score from 40.50 to 38.84, illustrating a trade-off between component retention and composed-task performance.

Additional single-component controls show that code-only OPD and DocQA-only OPD improve their respective target benchmarks. MOPD retains these component improvements, but its LongBench-v2 Code score remains at the base's 44.00, despite exceeding the single-component students' 42.00 and 40.00 (\Cref{app:code-long-context-controls}).


The code plus agent, math plus long-context, and code plus long-context evaluations show a compositional generalization limitation: a student can acquire improvements on separately supervised capabilities without obtaining the expected performance when those capabilities must be combined.



\Needspace{6\baselineskip}
\section{Discussion}
\label{sec:analysis-discussion}
\subsection{Teacher Capability and Data Coverage Contribute to OOD Gaps}
\label{sec:coverage}

\begin{figure}[!htbp]
\centering
\begin{minipage}[t]{0.48\linewidth}
\vspace{0pt}\centering
\includegraphics[width=\linewidth]{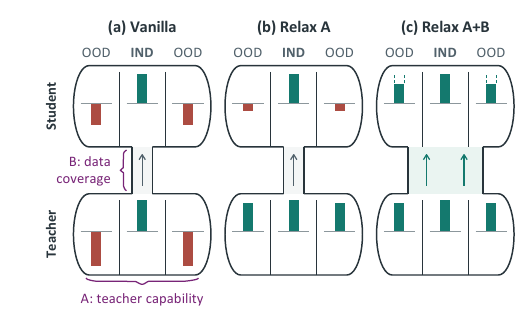}
\end{minipage}\hfill
\begin{minipage}[t]{0.46\linewidth}
\vspace{0pt}\centering
\includegraphics[width=\linewidth]{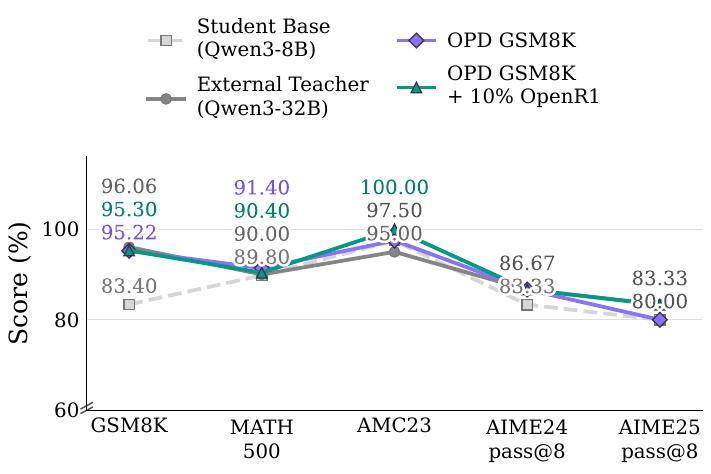}
\end{minipage}
\par\smallskip
\begin{minipage}[t]{0.52\linewidth}
\caption{\textbf{Teacher capability (A) and data coverage (B) as bottlenecks.} We use the flask~\citep{ujjain2021adsorption} as an analogy: lower/upper vessels denote teacher/student, and neck width denotes training data coverage. Teal/red bars illustrate qualitative gains/losses relative to the base. (a) Vanilla: The student exhibits less OOD forgetting than its teacher but remains below the base. (b) Relax A: A stronger teacher provides positive OOD gains, yet narrow distillation data still leaves the student below the base. (c) Relax A+B: Combining the stronger teacher with broader data coverage can turn student OOD losses into gains. 
}
\label{fig:teacher-data-bottlenecks}
\end{minipage}\hfill
\begin{minipage}[t]{0.43\linewidth}
\caption{\textbf{Broader data eliminates OOD forgetting.} With Qwen3-32B supervision on GSM8K with 10\% OpenR1 additional data to broaden the distribution, the Qwen3-8B student matches the teacher's AIME pass@8, surpasses it on AMC23, and exceeds the base on all five benchmarks. AIME reports pass@8; other benchmarks report accuracy.}
\label{fig:data-coverage}
\end{minipage}
\end{figure}

Under narrow supervision, students can remain below the base model on OOD evaluations with either self-RL specialists (\Cref{sec:self-rl}) or stronger external generalists (\Cref{sec:external-generalist}). Thus, stronger teacher capability alone is insufficient for OOD retention (panels a-b of \Cref{fig:teacher-data-bottlenecks}).
%
To relax both bottlenecks (panel c), we keep Qwen3-32B as teacher and train on 90\% GSM8K plus 10\% OpenR1~\citep{openr12025math}. 
Compared with GSM8K-only distillation, it exceeds the base on all five benchmarks.
The student matches the teacher's AIME24/25 pass@8 (86.67/83.33). On AMC23, it improves to 100.00, exceeding the teacher's 95.00 (\Cref{fig:data-coverage}).
Together, these results suggest considering teacher capability and data coverage jointly (\Cref{fig:teacher-data-bottlenecks}): a stronger teacher makes OOD capabilities available, while broader supervision helps the student retain or acquire them. Easing both bottlenecks can reduce OOD forgetting and potentially encourage further generalization.
\FloatBarrier

\subsection{Geometry of (M)OPD: Compact Updates and Shared Update Structure}
\label{sec:geometry}

MOPD exhibits a smaller OOD generalization gap than mixed RL in matched task mixtures (\Cref{fig:multi-math}). In this section, we examine the associated parameter updates to understand how the two consolidation procedures differ.
For each model, we measure its update relative to the common base, $\Delta W=W-W_0$. 

\Needspace{6\baselineskip}
\subsubsection{(M)OPD Produces More Compact Updates than (Mixed) RL}

\begin{wraptable}{r}{0.49\linewidth}
\centering\small
\caption{MOPD has smaller updates on average and shares more structure with corresponding single-teacher updates than mixed RL does in both generalization settings. Top: single-teacher update norms and top-64 energy. Bottom: means across the extrapolative and compositional settings; the norm ratio is MOPD/mixed RL. $R^2$ uses the single-OPD span for MOPD and the single-RL span for mixed RL.}
\label{tab:main_geometry_mopd}
\setlength{\tabcolsep}{4pt}
\begin{tabular}{@{}lrrr@{}}
\toprule
\multicolumn{2}{@{}l}{Single-teacher updates} & RL & OPD\\
\midrule
\multicolumn{2}{@{}l}{Absolute update norm} & $0.278$ & $0.154$\\
\multicolumn{2}{@{}l}{Top-64 energy} & $0.399$ & $0.690$\\
\midrule
Generalization & Norm ratio & \multicolumn{2}{c@{}}{$R^2$}\\
 & & Mixed RL & MOPD\\
\midrule
Extrapolative & 0.477 & 0.00173 & 0.157\\
Compositional & 0.617 & 0.000245 & 0.121\\
\bottomrule
\end{tabular}
\end{wraptable}

\textbf{Both OPD and MOPD produce smaller updates on average than their corresponding RL baselines} (\Cref{tab:main_geometry_mopd}). Across the extrapolative experiments, at approximately matched IND performance, OPD's update norm is 55.4\% of RL's (0.154 versus 0.278). For MOPD, the mean update norm is also smaller than that of mixed RL.
%

\textbf{OPD updates are also more spectrally concentrated.} OPD's top 64 singular directions capture 69.0\% of update energy, compared with 39.9\% for its RL teachers.
Stable rank, defined as $\|\Delta W\|_F^2/\|\Delta W\|_2^2$, measures concentration relative to the largest singular direction and decreases from 19.0 for RL to 6.90 for OPD. OPD also rotates the measured principal subspace of the base weights less.
%
OPD can therefore achieve near-perfect IND distillation while producing substantially more concentrated parameter updates than RL.

\WFclear
\subsubsection{MOPD Shares More Structure with Single-Teacher Updates}

For each mixture, we project the MOPD update onto the span of the corresponding single-teacher OPD updates and the mixed-RL update onto the span of the corresponding single-RL updates. We measure the explained fraction of each update's squared norm, $R^2=1-\|\Delta-\widehat{\Delta}\|_F^2/\|\Delta\|_F^2$, where $\widehat{\Delta}$ is the least-squares projection. \textbf{MOPD is better explained by the single-OPD span than mixed RL is by the single-RL span.} The respective spans explain 11.0-18.6\% of MOPD's squared update norm and less than 0.240\% of mixed RL's (\Cref{tab:main_geometry_mopd}).
Using the same single-RL teacher span for both methods also yields a higher $R^2$ for MOPD than for mixed RL.
These measurements suggest that MOPD is an algorithm with more shared update structure.
\textbf{Discussion.} Mergeability offers a perspective on capability consolidation. Understanding which models and capabilities can be merged successfully, and with which algorithms, remains an open question. 
The literature considers factors including update direction~\citep{ilharco2023taskarithmetic}, weight disentanglement~\citep{ortizjimenez2023tangent}, parameter-wise sign conflicts~\citep{yadav2023tiesmerging}, and subspace and gradient alignment~\citep{zhou2026mergeability}.
Our observations of smaller (M)OPD updates and greater shared update structure with single-teacher OPD motivate a hypothesis: compact updates may limit disruption to component capabilities, providing more favorable conditions for compositional generalization after capability consolidation. This perspective also motivates model-merging-style editing to improve OOD performance after distillation (\Cref{sec:patching}).
Appendix~\ref{app:opd-rl-theory} provides a complementary local theoretical analysis.
Please note that these connections remain a preliminary interpretation. A causal link between these geometric properties and generalization has yet to be established.


\section{Exploring Patching Methods for Mitigating OOD Gap}
\label{sec:retention}

Since near-perfect IND distillation can coexist with an OOD performance gap, improving generalization also requires preserving capabilities beyond the training distribution.
We investigate two explicit interventions: geometry-based patching after distillation and base-preserving replay during distillation. 

\subsection{Geometry-Based Patching}
\label{sec:patching}
The compact, spectrally concentrated OPD updates in \Cref{sec:geometry} motivate post-hoc editing: OPD students can forget less than self-RL teachers at similar IND performance. We hypothesize that suppressing update components that harm OOD retention but contribute little to IND performance can reduce forgetting; these components may arise from optimization noise.

\begin{wrapfigure}{r}{0.4\textwidth}
\centering
\vspace{-\intextsep}
\includegraphics[width=\linewidth]{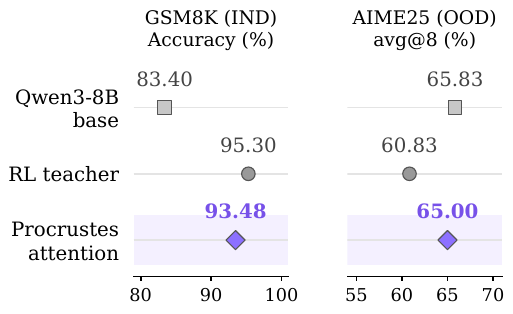}
\caption{Procrustes-attention patching recovers OOD performance. Patching ($\alpha=0.40$) raises AIME25 avg@8 from 60.83\% to 65.00\% while maintaining high GSM8K accuracy at 93.48\%.}
\label{fig:patching}
\end{wrapfigure}

We propose \emph{Procrustes-attention patching}, based on orthogonal Procrustes alignment~\citep{schonemann1966procrustes, yang2026orthogonal}. For each selected attention weight matrix, define $\Delta_S=W_{\mathrm{OPD}}-W_0$ and $\Delta_T=W_{\mathrm{RL\ teacher}}-W_0$, where $W_0$ denotes the base weights. We align the OPD update to the RL-teacher update and subtract the unmatched component:
\begin{equation}
\begin{aligned}
R^\star &= \arg\min_{R^\top R=I}\|\Delta_S R-\Delta_T\|_F^2,\\
E_{\mathrm{Proc}} &= \Delta_T-\Delta_S R^\star,\\
W_{\mathrm{patch}} &= W_{\mathrm{OPD}}-\alpha E_{\mathrm{Proc}}.
\end{aligned}
\label{eq:procrustes-patching}
\end{equation}
If $\Delta_S^\top\Delta_T=U\Sigma V^\top$ is a singular value decomposition, then $R^\star=UV^\top$; $\alpha$ controls the patch strength.
Conceptually, $E_{\mathrm{Proc}}$ mainly reflects the OOD behavior direction but has a minor impact on IND, since OPD and RL-teacher share near IND performance.
At $\alpha=0.40$, patching raises AIME25 avg@8 from 60.83\% to 65.00\% (+4.17 percentage points), leaving a gap of 0.83 percentage points to the base while retaining most of the teacher's IND performance (\Cref{fig:patching}).


\subsection{Base-Preserving Replay}
\label{sec:replay}
\begin{figure}[!htbp]
\centering
\includegraphics[width=0.85\linewidth]{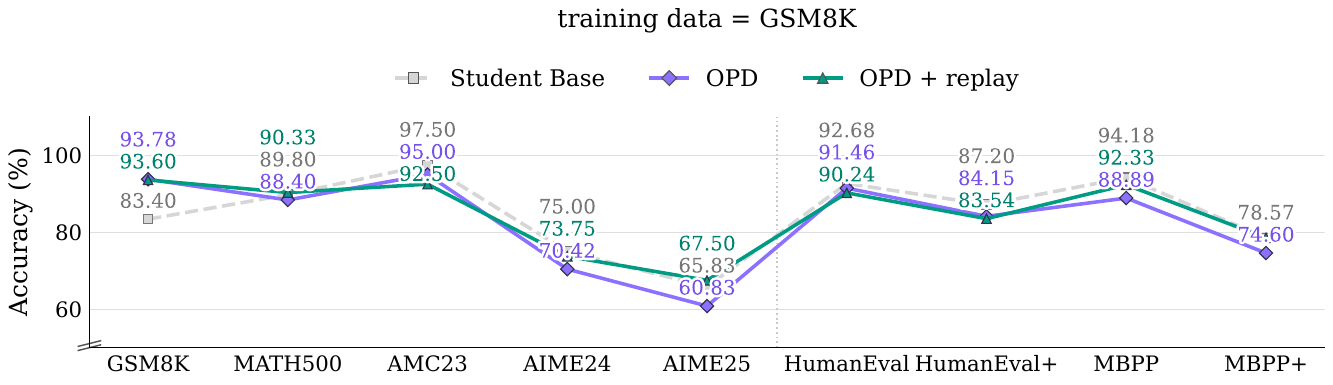}
\caption{Base-preserving replay improves five of eight OOD scores over plain OPD with a small IND change, and surpasses both the student base and RL teacher on AIME25. AIME scores report avg@8.}
\label{fig:replay}
\end{figure}

KL-regularized RLVR constrains policy drift from a reference model~\citep{shao2024deepseekmathpushinglimitsmathematical}. In contrast, the (M)OPD objectives in \Cref{sec:teachers} match task teachers without an explicit constraint to preserve the base model's behavior. We hypothesize that this missing constraint contributes to OOD forgetting. Motivated by replay methods for capability preservation during RL post-training~\citep{phan2026beyondreasoninggains}, we introduce base-preserving replay for (M)OPD.
%
Base-preserving replay mixes IND examples supervised by the task teachers with additional prompts supervised by the base model, using the same OPD loss. We test coding and instruction-following data as replay sources outside the specialist's mathematical training distribution; \Cref{fig:replay} reports coding replay for the GSM8K OPD student distilled from a self-RL teacher.

Relative to plain OPD, coding replay improves five of eight OOD scores, including a 6.67-point increase in AIME25 avg@8 to 67.50, surpassing both the student base and RL teacher with only a 0.18-point reduction in GSM8K accuracy. These results suggest that replay can support both retention and generalization beyond the replay domain.


\section{Further Discussion and Limitations}
\label{sec:discussion}

\subsection{Limitations}
\label{sec:limitations}

Our main post-training experiments are on Qwen3-8B students, and our largest external teacher is Qwen3-32B given computation constraints. 
Fixing the student supports within-model comparisons, but leaves dependence on student capacity, teacher strength, and model family under-explored. 
We have preliminarily tested self-RL specialist extrapolative generalization on Qwen3-235B-A22B on a smaller scale. In that run, we do not see the generalization limitation as minor as size scales. 

Public datasets make our comparisons inspectable, but do not reproduce the breadth or adaptive data selection of full post-training pipelines in frontier model development. 
Our OOD labels are relative to post-training data, not pretraining exposure. 
Broader training data from the real world and held-out evaluations with graded difficulty would test generalization under more demanding conditions.

Mixed RL and MOPD both combine component recipes without extensive tuning of reward balancing, mixture weights, or curricula (\Cref{app:training-protocols}). 
They may not establish each objective's best achievable performance.
However, we have tuned hyperparameters for single OPD and single RL recipes to ensure optimal performance.
Given the much larger tuning space for Mixed RL and MOPD, we treat the current settings as out-of-the-box usage and still a fair comparison. We leave these additional tunings as further work.

Our main findings on the relative performance of MOPD and mixed RL concern the tested LLM architecture and practical post-training tasks. At the algorithmic level, this ordering need not hold for simpler models and tasks with different generalization requirements, which we additionally discuss in \Cref{sec:arithmetic-discussion}.

Computational constraints prevent us from repeating every experiment with multiple sampling seeds.
For mathematical reasoning, we repeat evaluation with four sampling seeds on a 10\% subset of the evaluation questions and observe consistent qualitative conclusions across these checks.
These checks assess variability in sampled responses; they do not establish robustness across independent training runs.

\Needspace{6\baselineskip}
\subsection{Scope of MOPD's Advantage over Mixed RL}
\label{sec:arithmetic-discussion}


Our study focuses on empirical performance in practical post-training tasks, rather than establishing a universal ordering between MOPD and mixed RL. To be more careful, the advantage observed in \Cref{sec:compositional} is conditional on the evaluated models, training data, and tasks.
Task learnability for a given model may affect this comparison. Conceptually, MOPD learns from teacher distributions without an independent correctness signal, whereas RL uses verifier feedback on its training tasks. This distinction suggests that mixed RL may outperform MOPD when the model can learn effectively from verifier feedback, even if the teachers remain imperfect. We examine this possibility in a controlled synthetic arithmetic setting where the chosen model can achieve perfect accuracy on the target compositions under direct supervision.

\begin{wrapfigure}{r}{0.55\textwidth}
\centering
\vspace{-\intextsep}
\includegraphics[width=\linewidth]{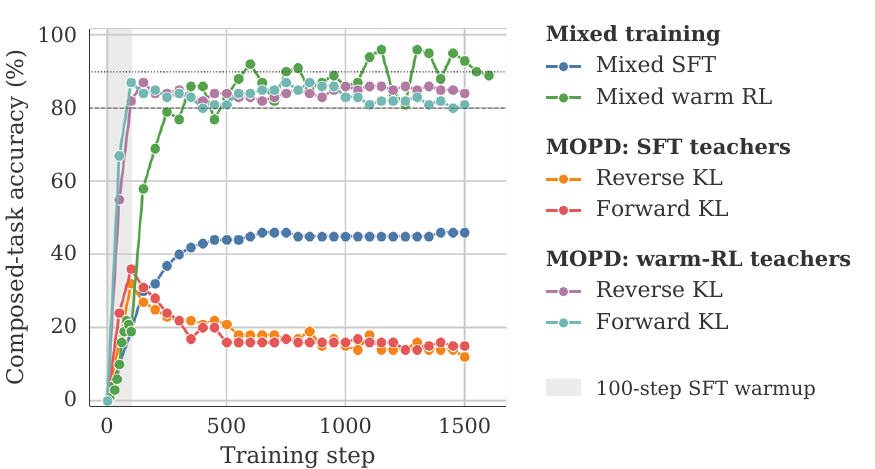}
\caption{Mixed RL (with warmup) can surpass MOPD on generalization to held-out arithmetic compositions. The training dynamics show $\min(\mathrm{AB\ accuracy},\mathrm{BA\ accuracy})$ (\%). MOPD uses SFT or warm-RL teachers with reverse or forward KL; all MOPD students start from the pretrained base. Shading marks mixed warm RL's 100-step SFT warmup (which contributes to better performance than direct RL).}
\label{fig:arithmetic-discussion}
\vspace{-\intextsep}
\end{wrapfigure}

The controlled synthetic arithmetic experiment run on a 3-layer Transformer model provides an example (\Cref{fig:arithmetic-discussion}; setup in \Cref{app:arithmetic}). 
We design an easier generalization task that the model of a given size can converge to high performance.
Models first train on operators $A$ and $B$ and their repetitions, with $AB$ and $BA$ held out from both MOPD and mixed RL. 
We design the task and model architecture that direct SFT on $AB/BA$ reaches 100\% accuracy, establishing learnability under direct supervision. 
In the displayed run, both MOPD variants from warm-RL teachers improve rapidly and then remain around $80$-$90$\% composed-task accuracy within the observed training budget, while mixed warm RL later surpasses them. The five-seed final results support this finding.

These results reveal that, considering different model architectures, task learnability, interference between tasks, etc., there can be different balances between component performance and compositional generalization.
Isolating their effects across architectures and training objectives remains future work. 
Our main conclusions concern the LLMs and practical post-training tasks.

\subsection{Future Work}
\label{sec:future-work}

The connection between update geometry, OOD retention, and compositional generalization remains a hypothesis. Causal tests should manipulate update directions while controlling update magnitude, training budget, and IND performance. Beyond post-hoc patching, step-level parameter-space control could project or regularize updates during training, with random-direction controls. Future work should also jointly study teacher assignment, data coverage, and base-preserving replay under a fixed budget, testing retention and further generalization beyond expanded training coverage.


\clearpage
\bibliography{main}

@misc{agarwal2024onpolicydistillationlanguagemodels,
      title={On-Policy Distillation of Language Models: Learning from Self-Generated Mistakes},
      author={Rishabh Agarwal and Nino Vieillard and Yongchao Zhou and Piotr Stanczyk and Sabela Ramos and Matthieu Geist and Olivier Bachem},
      year={2024},
      eprint={2306.13649},
      archivePrefix={arXiv},
      primaryClass={cs.LG},
      url={https://arxiv.org/abs/2306.13649},
}

@misc{ma2026mopdmultiteacheronpolicydistillation,
      title={MOPD: Multi-Teacher On-Policy Distillation for Capability Integration in LLM Post-Training},
      author={Wenhan Ma and Jianyu Wei and Liang Zhao and Hailin Zhang and Bangjun Xiao and Lei Li and Qibin Yang and Bofei Gao and Yudong Wang and Rang Li and Jinhao Dong and Zhifang Sui and Fuli Luo},
      year={2026},
      eprint={2606.30406},
      archivePrefix={arXiv},
      primaryClass={cs.CL},
      url={https://arxiv.org/abs/2606.30406},
}

@misc{coreteam2026mimov2flashtechnicalreport,
      title={MiMo-V2-Flash Technical Report},
      author={{Core Team} and Bangjun Xiao and Bingquan Xia and Bo Yang and Bofei Gao and Bowen Shen and Chen Zhang and Chenhong He and Chiheng Lou and Fuli Luo and Gang Wang and Gang Xie and Hailin Zhang and Hanglong Lv and Hanyu Li and Heyu Chen and Hongshen Xu and Houbin Zhang and Huaqiu Liu and Jiangshan Duo and Jianyu Wei and Jiebao Xiao and Jinhao Dong and Jun Shi and Junhao Hu and Kainan Bao and Kang Zhou and Lei Li and Liang Zhao and Linghao Zhang and Peidian Li and Qianli Chen and Shaohui Liu and Shihua Yu and Shijie Cao and Shimao Chen and Shouqiu Yu and Shuo Liu and Tianling Zhou and Weijiang Su and Weikun Wang and Wenhan Ma and Xiangwei Deng and Bohan Mao and Bowen Ye and Can Cai and Chenghua Wang and Chengxuan Zhu and Chong Ma and Chun Chen and Chunan Li and Dawei Zhu and Deshan Xiao and Dong Zhang and Duo Zhang and Fangyue Liu and Feiyu Yang and Fengyuan Shi and Guoan Wang and Hao Tian and Hao Wu and Heng Qu and Hongfei Yi and Hongxu An and Hongyi Guan and Xing Zhang and Yifan Song and Yihan Yan and Yihao Zhao and Yingchun Lai and Yizhao Gao and Yu Cheng and Yuanyuan Tian and Yudong Wang and Zhen Tang and Zhengju Tang and Zhengtao Wen and Zhichao Song and Zhixian Zheng and Zihan Jiang and Jian Wen and Jiarui Sun and Jiawei Li and Jinlong Xue and Jun Xia and Kai Fang and Menghang Zhu and Nuo Chen and Qian Tu and Qihao Zhang and Qiying Wang and Rang Li and Rui Ma and Shaolei Zhang and Shengfan Wang and Shicheng Li and Shuhao Gu and Shuhuai Ren and Sirui Deng and Tao Guo and Tianyang Lu and Weiji Zhuang and Weikang Zhang and Weimin Xiong and Wenshan Huang and Wenyu Yang and Xin Zhang and Xing Yong and Xu Wang and Xueyang Xie and Yilin Jiang and Yixin Yang and Yongzhe He and Yu Tu and Yuanliang Dong and Yuchen Liu and Yue Ma and Yue Yu and Yuxing Xiang and Zhaojun Huang and Zhenru Lin and Zhipeng Xu and Zhiyang Chen and Zhonghua Deng and Zihan Zhang and Zihao Yue},
      year={2026},
      eprint={2601.02780},
      archivePrefix={arXiv},
      primaryClass={cs.CL},
      url={https://arxiv.org/abs/2601.02780},
}

@misc{deepseekai2026deepseekv4highlyefficientmilliontoken,
      title={DeepSeek-V4: Towards Highly Efficient Million-Token Context Intelligence},
      author={{DeepSeek-AI} and Anyi Xu and Bangcai Lin and Bing Xue and Bingxuan Wang and Bingzheng Xu and Bochao Wu and Bowei Zhang and Chaofan Lin and Chen Dong and Chenchen Ling and Chengda Lu and Chenggang Zhao and Chengqi Deng and Chengyu Hou and Chenhao Xu and Chenze Shao and Chong Ruan and Conner Sun and Damai Dai and Daya Guo and Dejian Yang and Deli Chen and Donghao Li and Dongjie Ji and Erhang Li and Fang Wei and Fangyun Lin and Fangzhou Yuan and Feiyu Xia and Fucong Dai and Guangbo Hao and Guanting Chen and Guoai Cao and Guolai Meng and Guowei Li and Han Yu and Han Zhang and Hanwei Xu and Hao Li and Haofen Liang and Haoling Zhang and Haoming Luo and Haoran Wei and Haotian Yuan and Haowei Zhang and Haowen Luo and Haoyu Chen and Haozhe Ji and Hengqing Zhang and Honghui Ding and Hongxuan Tang and Huanqi Cao and Huazuo Gao and Hui Qu and Hui Zeng and J Yang and JQ Zhu and Jia Luo and Jia Song and Jia Yu and Jialiang Huang and Jialu Cai and Jian Liang and Jiangting Zhou and Jiasheng Ye and Jiashi Li and Jiaxin Xu and Jiewen Hu and Jieyu Yang and Jin Chen and Jin Yan and Jingchang Chen and Jingli Zhou and Jingting Xiang and Jingyang Yuan and Jingyuan Cheng and Jingzi Zhou and Jinhua Zhu and Jiping Yu and Joseph Sun and Jun Ran and Junguang Jiang and Junjie Qiu and Junlong Li and Junmin Zheng and Junxiao Song and Kai Dong and Kaige Gao and Kang Guan and Kexing Zhou and Kezhao Huang and Kuai Yu and Lean Wang and Lecong Zhang and Lei Wang and Leyi Xia and Li Zhang and Liang Zhao and Lihua Guo and Lingxiao Luo and Linwang Ma and Linyan Zhu and Litong Wang and Liyu Cai and Liyue Zhang and Longhao Chen and MS Di and MY Xu and Max Mei and Miaojun Wang and Mingchuan Zhang and Minghua Zhang and Minghui Tang and Mingming Li and Mingxu Zhou and Minmin Han and Ning Wang and Panpan Huang and Panpan Wang and Peixin Cong and Peiyi Wang and Peng Zhang and Qiancheng Wang and Qihao Zhu and Qingyang Li and Qinyu Chen and Qiushi Du and Qiwei Jiang and Rui Tian and Ruifan Xu and Ruijie Lu and Ruiling Xu and Ruiqi Ge and Ruisong Zhang and Ruizhe Pan and Runji Wang and Runqian Chen and Runqiu Yin and Runxin Xu and Ruomeng Shen and Ruoyu Zhang and Ruyi Chen and SH Liu and Shanghao Lu and Shangmian Sun and Shangyan Zhou and Shanhuang Chen and Shaofei Cai and Shaoheng Nie and Shaoqing Wu and Shaoyuan Chen and Shengding Hu and Shengyu Liu and Shiqiang Hu and Shirong Ma and Shiyu Wang and Shuiping Yu and Shunfeng Zhou and Shuting Pan and Shuying Yu and Songyang Zhou and Tao Ni and Tao Yun and Tian Jin and Tian Pei and Tian Ye and Tianle Lin and Tianran Ji and Tianyi Cui and Tianyuan Yue and Tingting Yu and Tun Wang and W Zhang and WL Xiao and Wangding Zeng and Wei An and Weilin Zhao and Wen Liu and Wenfeng Liang and Wenjie Pang and Wenjing Luo and Wenjing Yao and Wenjun Gao and Wenkai Yang and Wenlve Huang and Wenqing Hou and Wentao Zhang and Wenting Ma and Xi Gao and Xiang He and Xiangwen Wang and Xianzu Wang and Xiao Bi and Xiaodong Liu and Xiaohan Wang and Xiaokang Chen and Xiaokang Zhang and Xiaotao Nie and Xiaowen Sun and Xiaoxiang Wang and Xin Cheng and Xin Liu and Xin Xie and Xingchao Liu and Xingchen Liu and Xingkai Yu and Xingyou Li and Xinyu Yang and Xinyu Zhang and Xu Chen and Xuanyu Wang and Xuecheng Su and Xueyin Chen and Xuheng Lin and Xuwei Fu and YC Yan and YQ Wang and YW Ma and Yanfeng Luo and Yang Zhang and Yanhong Xu and Yanru Ma and Yanwen Huang and Yao Li and Yao Li and Yao Xu and Yao Zhao and Yaofeng Sun and Yaohui Wang and Yi Qian and Yi Shao and Yi Yu and Yichao Zhang and Yifan Ding and Yifan Shi and Yijia Wu and Yiliang Xiong and Yiling Ma and Ying He and Ying Tang and Ying Zhou and Yingjia Luo and Yinmin Zhong and Yishi Piao and Yisong Wang and Yixiang Zhang and Yixiao Chen and Yixuan Tan and Yixuan Wei and Yiyang Ma and Yiyuan Liu and Yonglun Yang and Yongqiang Guo and Yongtong Wu and Yu Wu and YuKun Li and Yuan Cheng and Yuan Ou and Yuanfan Xu and Yuanhao Li and Yuduan Wang and Yuehan Yang and Yuer Xu and Yuhan Wu and Yuhao Meng and Yuheng Zou and Yukun Zha and Yunfan Xiong and Yupeng Chen and Yuping Lin and Yuqian Cao and Yuqian Wang and Yushun Zhang and Yuting Yan and Yutong Lin and Yuxian Gu and Yuxiang Luo and Yuxiang You and Yuxuan Liu and Yuxuan Zhou and Yuyang Zhou and Yuzhen Huang and ZF Wu and Zehao Wang and Zehua Zhao and Zehui Ren and Zekai Zhang and Zhangli Sha and Zhe Fu and Zhe Ju and Zhean Xu and Zhenda Xie and Zhengyan Zhang and Zheren Gao and Zhewen Hao and Zhibin Gou and Zhicheng Ma and Zhigang Yan and Zhihong Shao and Zhixian Huang and Zhixuan Chen and Zhiyu Wu and Zhizhou Ren and Zhongyu Wu and Zhuoshu Li and Zhuping Zhang and Zian Xu and Zihao Wang and Zihua Qu and Zihui Gu and Zijia Zhu and Zilin Li and Zipeng Zhang and Ziwei Xie and Ziyi Gao and Ziyi Wan and Zizheng Pan and Zongqing Yao},
      year={2026},
      eprint={2606.19348},
      archivePrefix={arXiv},
      primaryClass={cs.CL},
      url={https://arxiv.org/abs/2606.19348},
}

@misc{glm5team2026glm5vibecodingagentic,
      title={GLM-5: from Vibe Coding to Agentic Engineering},
      author={{GLM-5-Team} and Aohan Zeng and Xin Lv and Zhenyu Hou and Zhengxiao Du and Qinkai Zheng and Bin Chen and Da Yin and Chendi Ge and Chenghua Huang and Chengxing Xie and Chenzheng Zhu and Congfeng Yin and Cunxiang Wang and Gengzheng Pan and Hao Zeng and Haoke Zhang and Haoran Wang and Huilong Chen and Jiajie Zhang and Jian Jiao and Jiaqi Guo and Jingsen Wang and Jingzhao Du and Jinzhu Wu and Kedong Wang and Lei Li and Lin Fan and Lucen Zhong and Mingdao Liu and Mingming Zhao and Pengfan Du and Qian Dong and Rui Lu and Shuang-Li and Shulin Cao and Song Liu and Ting Jiang and Xiaodong Chen and Xiaohan Zhang and Xuancheng Huang and Xuezhen Dong and Yabo Xu and Yao Wei and Yifan An and Yilin Niu and Yitong Zhu and Yuanhao Wen and Yukuo Cen and Yushi Bai and Zhongpei Qiao and Zihan Wang and Zikang Wang and Zilin Zhu and Ziqiang Liu and Zixuan Li and Bojie Wang and Bosi Wen and Can Huang and Changpeng Cai and Chao Yu and Chen Li and Chengwei Hu and Chenhui Zhang and Dan Zhang and Daoyan Lin and Dayong Yang and Di Wang and Ding Ai and Erle Zhu and Fangzhou Yi and Feiyu Chen and Guohong Wen and Hailong Sun and Haisha Zhao and Haiyi Hu and Hanchen Zhang and Hanrui Liu and Hanyu Zhang and Hao Peng and Hao Tai and Haobo Zhang and He Liu and Hongwei Wang and Hongxi Yan and Hongyu Ge and Huan Liu and Huanpeng Chu and Jia'ni Zhao and Jiachen Wang and Jiajing Zhao and Jiamin Ren and Jiapeng Wang and Jiaxin Zhang and Jiayi Gui and Jiayue Zhao and Jijie Li and Jing An and Jing Li and Jingwei Yuan and Jinhua Du and Jinxin Liu and Junkai Zhi and Junwen Duan and Kaiyue Zhou and Kangjian Wei and Ke Wang and Keyun Luo and Laiqiang Zhang and Leigang Sha and Liang Xu and Lindong Wu and Lintao Ding and Lu Chen and Minghao Li and Nianyi Lin and Pan Ta and Qiang Zou and Rongjun Song and Ruiqi Yang and Shangqing Tu and Shangtong Yang and Shaoxiang Wu and Shengyan Zhang and Shijie Li and Shuang Li and Shuyi Fan and Wei Qin and Wei Tian and Weining Zhang and Wenbo Yu and Wenjie Liang and Xiang Kuang and Xiangmeng Cheng and Xiangyang Li and Xiaoquan Yan and Xiaowei Hu and Xiaoying Ling and Xing Fan and Xingye Xia and Xinyuan Zhang and Xinze Zhang and Xirui Pan and Xu Zou and Xunkai Zhang and Yadi Liu and Yandong Wu and Yanfu Li and Yidong Wang and Yifan Zhu and Yijun Tan and Yilin Zhou and Yiming Pan and Ying Zhang and Yinpei Su and Yipeng Geng and Yong Yan and Yonglin Tan and Yuean Bi and Yuhan Shen and Yuhao Yang and Yujiang Li and Yunan Liu and Yunqing Wang and Yuntao Li and Yurong Wu and Yutao Zhang and Yuxi Duan and Yuxuan Zhang and Zezhen Liu and Zhengtao Jiang and Zhenhe Yan and Zheyu Zhang and Zhixiang Wei and Zhuo Chen and Zhuoer Feng and Zijun Yao and Ziwei Chai and Ziyuan Wang and Zuzhou Zhang and Bin Xu and Minlie Huang and Hongning Wang and Juanzi Li and Yuxiao Dong and Jie Tang},
      year={2026},
      eprint={2602.15763},
      archivePrefix={arXiv},
      primaryClass={cs.LG},
      url={https://arxiv.org/abs/2602.15763},
}

@misc{nvidia2026nemotron3ultraopen,
      title={Nemotron 3 Ultra: Open, Efficient Mixture-of-Experts Hybrid Mamba-Transformer Model for Agentic Reasoning},
      author={{NVIDIA} and Aaron Blakeman and Aaron Thomas and Aastha Jhunjhunwala and Abhibha Gupta and Abhinav Khattar and Adam Rajfer and Adi Renduchintala and Adil Asif and Aditya Vavre and Adriana Flores Miranda and Ahmad Bilal and Aileen Zaman and Ajay Hotchandani and Akanksha Shukla and Akhiad Bercovich and Aleksander Ficek and Alex Gronskiy and Alex Kondratenko and Alex Steiner and Alex Ye and Alexander Bukharin and Alexandre Milesi and Ali Taghibakhshi and Alice Gatti and Alisa Liu and Alok Kumar and Amar Phanishayee and Ameya Sunil Mahabaleshwarkar and Amir Klein and Amit Zuker and Amnon Geifman and Anahita Bhiwandiwalla and Ananth Subramaniam and Andrea Santilli and Andrew Fulks and Andrew McHarg and Andrew Tao and Andrii Skliar and Anjulie Agrusa and Ankur Srivastava and Ankur Verma and Anna Shors and Anna Warno and Antoni-Joan Solergibert I Llaquet and Arham Mehta and Arkadiusz Nowaczynski and Arti Jain and Ashwath Aithal and Ashwin Poojary and Asif Ahamed and Asit Mishra and Asma Kuriparambil Thekkumpate and Atefeh Sohrabizadeh and Avinash Kaur and Avinash Vem and Ayush Dattagupta and Barath Subramaniam Anandan and Bardiya Sadeghi and Ben Lanir and Benedikt Schifferer and Besmira Nushi and Bilal Kartal and Bill Thiede and Bita Darvish Rouhani and Bo Deng and Bob Schatz and Boris Ginsburg and Boxin Wang and Brad Nemire and Brandon Norick and Brian Dang and Brian Westphal and Brian Yu and Brucek Khailany and Bryan Catanzaro and Carlo del Mundo and Caryln Aarish and Chankyu Lee and Chantal Hwang and Charbel Sakr and Charles Wang and Charlie Truong and Chen Cui and Cheng Cheng and Cheng-Ping Hsieh and Chenghao Zhang and Chenhui Deng and Chintan Patel and Chris Alexiuk and Christian Cosgrove and Christian Munley and Christine Harvey and Christopher Parisien and Chunyang Shen and Coco Li and Collin Neale and Cynthia Gao and Cyril Meurillon and Dan Gil and Dan Su and Dan Zhao and Dane Corneil and Daniel Afrimi and Daniel Egert and Daniel Korzekwa and Daniel Lo and Daniel Machlab and Daniel Serebrenik and Daniil Sorokin and Daria Gitman and Daria Levy and Darko Stosic and David Mosallanezhad and David Yu and Davit Karamyan and Deena Donia and Deep Debroy and Deepak Narayanan and Devin O'Kelly and Dheeraj Peri and Dhruv Nathawani and Di and Wu and Dima Rekesh and Divyanshu Kakwani and Donald Plummer and Dong Anh and Dongfeng Yu and Dongfu Jiang and Donnie Kim and Dorrin Poorkay and Duncan Riach and Dusan Stosic and Dustin VanStee and Eavan Meng and Edgar Minasyan and Edward Lin and Eileen Margaret Peters Long and Elad Sarafin and Elad Segal and Elena Lantz and Ellie Evans and Elliott Ning and Eric Chung and Eric Harper and Eric Pham-Hung and Eric Tramel and Eric Yang and Erick Galinkin and Erik Pounds and Erika Goncalves Goncalves and Evan Briones and Evan Wu and Evelina Bakhturina and Evgeny Tsykunov and Ewa Dobrowolska and Faisal Ladhak and Farzan Memarian and Fay Wang and Fei Jia and Felipe Soares and Felipe Vieira Frujeri and Feng Chen and Fengguang Lin and Ferenc Galko and Frank Sun and Frankie Siino and Frida Hou and Gal Hubara Agam and Gal Kaplun and Gantavya Bhatt and Gargi Prasad and Garvit Kulshreshtha and George Armstrong and Gerald Shen and Giulio Borghesi and Gordana Neskovic and Gorkem Batmaz and Grace Lam and Greg Mason and Greg Pauloski and Grigor Nalbandyan and Grzegorz Chlebus and Grzegorz Karch and Guan-Ting Liu and Guoming Zhang and Guyue Huang and Haggai Maron and Haifeng Qian and Haim Elisha and Haoxing Ren and Haran Kumar Shiv Kumar and Haribhau Hud and Harris Nover and Harrison Saturley Hall and Hayate Iso and Helen Ngo and Herbert Hum and Herman Sahota and Hexin Wang and Himanshu Soni and Hovhannes Tamoyan and Hua Li and Huanhuan Chen and Hui Li and Hui Wang and Huy Nguyen and Ian Chiles and Ido Galil and Ido Shahaf and Igor Gitman and Igor Shovkun and Ilya Loshchilov and Ingo Guehring and Itamar Schen and Itay Levy and Itay Neeman and Ivan Moshkov and Izik Golan and Izzy Putterman and Jaemin Choi and Jakub Slowikowski and Jan Kautz and Jane Polak Scowcroft and Jared Casper and Jatin Mitra and Jeffrey Glick and Jenny Chen and Jesse Oliver and Jiacheng Xu and Jiafan Zhu and Jialin Song and Jian Zhang and Jiantao Jiao and Jiaqi Zeng and Jie Lou and Jim King and Jimmy Zhang and Jingquan Wang and Jinhang Choi and Jinju Chu and Joey Conway and Joey Guman and Johan Jatko and Johannes Rausch and John Kamalu and John Roberts and Johnny Greco and Johnny Mensel and Jonah Alben and Jonas Yang and Jonathan Cohen and Jonathan Raiman and Joseph Jennings and Joshua Mabry and Joshua Pierce and Joyjit Daw and Julien Veron Vialard and Junkeun Yi and Jupinder Parmar and Kajal Jain and Kan Zhu and Kari Briski and Katherine Cheung and Katherine Luna and Keith Willowhawk and Keith Wyss and Keshav Santhanam and Kevin Shih and Kezhi Kong and Khanh Nguyen and Khushi Bhardwaj and Kirthi Shankar Sivamani and Konstantinos Krommydas and Krishna C. Puvvada and Krzysztof Pawelec and Kumar Anik and Kyle Keprios and Kylie Day and Lawrence McAfee and Leo Du and Leon Derczynski and Li Ding and Linda Liu and Lingjie Wu and Lior Kadoch and Lizzie Wei and Luis Vega and Luke Robison and Lun Su and Maarten Van Segbroeck and Maciej Jakub Mikulski and Maer Rodrigues de Melo and Magda Sypula and Mahan Fathi and Makesh Narsimhan Sreedhar and Makesh Tarun Chandran and Manoj Kilaru and Maor Ashkenazi and Marc Cuevas and Marc Romeijn and Marcin Chochowski and Mark Cai and Mark Mozolewski and Markus Kliegl and Marta Stepniewska-Dziubinska and Martyna Patelka and Mattei Machczynski and Matvei Novikov and Mauricio Ferrato and Maximilian Golub and Mehrzad Samadi and Melissa Corpuz and Mengru Wang and Mengxi Wu and Meredith Price and Meriem Boubdir and Micah Schaffer and Michael Andersch and Michael Boone and Michael Gschwind and Michael Lightstone and Michael Loh and Michal Bien and Michal Zawalski and Michelle Gill and Miguel Martinez and Mikail Khona and Mike Chrzanowski and Mike Houston and Mingyuan Ma and Minseok Lee and Mohamed Fawzy and Mohammad Dabbah and Mohammad Shoeybi and Mostofa Patwary and Nabin Mulepati and Najeeb Nabwani and Namit Dhameja and Narimane Hennouni and Natalie Hereth and Nathaniel Pinckney and Nave Algarici and Nave Assaf and Netanel Haber and Nicholas Knight and Nick Reamaroon and Nickson Quak and Nidhi Bhatia and Nikhil Desai and Nikolai Ludwig and Nima Tajbakhsh and Ning Xu and Nir Ailon and Nirmal Juluru and Nitin Nitin and Ofri Masad and Oleg Rybakov and Oleksii Hrinchuk and Oleksii Kuchaiev and Olivia Viessmann and Olivier Delalleau and Oluwatobi Olabiyi and Omer Ullman Argov and Omri Puny and Oren Tropp and Pablo Ribalta and Pallab Bhattacharya and Panos Lampropoulos and Parth Mannan and Pasha Shamis and Patrick Legresley and Paul Gibbons and Pavlo Molchanov and Pawel Morkisz and Peter Dykas and Peter Jin and Pierre-Yves Aquilanti and Pinky Xu and Piotr Januszewski and Piotr Laskiewicz and Pooya Jannaty and Prakash Gurumurthy and Pranav Prashant Thombre and Prasoon Varshney and Pritam Gundecha and Przemek Tredak and Puhui Meng and Qiyu Wan and Rabeeh Karimi Mahabadi and Rachel Oberman and Rachit Garg and Radha Sri-Tharan and Rahul Kandu and Rakshit Sanadhya and Ran El-Yaniv and Ran Zilberstein and Rasoul Shafipour and Ray Macalisang and Rayen Tian and Reka Kovacs and Renjie Pi and Rick Izzo and Rima Shahbazyan and Rishabh Garg and Rishi Puri and Rita Fernandes Neves and Ritchie Zhao and Ritika Borkar and Ritu Gala and Riyad Islam and Robert Clark and Robert Hesse and Robert Kirby and Roger Waleffe and Rohit Watve and Roi Koren and Ron Banner and Ruoxi Zhang and Russell J. Hewett and Ryan Prenger and Ryan Stewart and Ryota Egashira and Sadegh Mahdavi and Saee Paliwal and Sagar Singh and Sahil Modi and Salika Dave and Samantha Shinagawa and Samuel Kriman and Sandip Bhaskar and Sangkug Lym and Sanjay Kariyappa and Sanjeev Satheesh and Saran Vikas Murari and Satish Pasumarthi and Saurabh Mishra and Saurav Muralidharan and Scott Hara and Sean Narentharen and Selvaraj Anandaraj and Seonjin Na and Seonmeyong Bak and Seonmyeong Bak and Sepehr Sameni and Seph Mard and Serge Panev and Seth Henneman and Seth Poulos and Shahar Mor and Shantanu Acharya and Shaona Ghosh and Sharath Turuvekere Sreenivas and Sharon Mendelson and Shaun Kotek and Shawn Wang and Shay Aharon and Shaya Gharghabi and Sheng-Chieh Lin and Shi Chen and Shiqing Fan and Shirish Baskaran and Shreya Gopa and Shrimai Prabhumoye and Shubham Pachori and Shubham Toshniwal and Shuoyang Ding and Shwetha Krishnamurthy and Siddharth Singh and Simeng Sun and Sirshak Das and Sivakumar Arayandi Thottakara and Smita Ithape and Somshubra Majumdar and Soumye Singhal and Sri Harsha Singudasu and Sridhar Bhuvanapalli and Srimukh Veccham and Stas Sergienko and Stefania Alborghetti and Stephen Ge and Su Rong and Sugam Dipak Devare and Sukrit Rao and Sumeet Kumar Barua and Sungsoo Ha and Sunny Gai and Suriya Gunasekar and Suseella Panguluri and Suyog Gupta and Sviataslau Hinzburh and Sweta Priyadarshi and Syeda Nahida Akter and Talor Abramovich and Tan Bui and Tanay Varshney and Tatevik Ter-Hovhannisyan and Teodor-Dumitru Ene and Terry Kong and Thanh Do and Tianhe Zhang and Tiffany Moore and Tijmen Blankevoort and Tim Moon and Tiyasa Mitra and Tom Balough and Tomasz Grzegorzek and Tomasz Hliwiak and Tomer Asida and Tomer Bar Natan and Tomer Keren and Tomer Ronen and Tony Salim and Tony Wang and Traian Rebedea and Tugrul Konuk and Twinkle Vashishth and Udi Karpas and Ushnish De and Vahid Noorozi and Venkat Srinivasan and Venmugil Elango and Vibhor Agrawal and Victor Cui and Vijay Korthikanti and Vikas Mehta and Vinay Rao and Virginia Wu and Vitaly Kurin and Vitaly Lavrukhin and Vladimir Anisimov and Vu Pham and Wanli Jiang and Wasi Uddin Ahmad and Wataru Ishihara and Wei Du and Wei Ping and Weiheng Chai and Wenliang Dai and Wesley Helmholz and Will Jennings and Will Zhu and Wojciech Prazuch and Xiaowei Ren and Xiwen Yu and Yan Breek and Yang Chen and Yang Yu and Yangyi Chen and Yaniv Galron and Yashaswi Karnati and Yejin Choi and Yev Meyer and Yi-Fu Wu and Yian Zhang and Ying Lin and Yonatan Geifman and Yonggan Fu and Youngeun Kwon and Yu Yao and Yugi Guvvla and Yuki Huang and Yunsheng Liu and Zach Moshe and Zachary Newell and Zhilin Wang and Zhiyu Li and Zhongbo Zhu and Zhuolin Yang and Zihan Liu and Zijie Yan and Zsolt-Alon Wertheimer},
      year={2026},
      eprint={2606.15007},
      archivePrefix={arXiv},
      primaryClass={cs.CL},
      url={https://arxiv.org/abs/2606.15007},
}

@misc{yang2025qwen3technicalreport,
      title={Qwen3 Technical Report},
      author={An Yang and Anfeng Li and Baosong Yang and Beichen Zhang and Binyuan Hui and Bo Zheng and Bowen Yu and Chang Gao and Chengen Huang and Chenxu Lv and Chujie Zheng and Dayiheng Liu and Fan Zhou and Fei Huang and Feng Hu and Hao Ge and Haoran Wei and Huan Lin and Jialong Tang and Jian Yang and Jianhong Tu and Jianwei Zhang and Jianxin Yang and Jiaxi Yang and Jing Zhou and Jingren Zhou and Junyang Lin and Kai Dang and Keqin Bao and Kexin Yang and Le Yu and Lianghao Deng and Mei Li and Mingfeng Xue and Mingze Li and Pei Zhang and Peng Wang and Qin Zhu and Rui Men and Ruize Gao and Shixuan Liu and Shuang Luo and Tianhao Li and Tianyi Tang and Wenbiao Yin and Xingzhang Ren and Xinyu Wang and Xinyu Zhang and Xuancheng Ren and Yang Fan and Yang Su and Yichang Zhang and Yinger Zhang and Yu Wan and Yuqiong Liu and Zekun Wang and Zeyu Cui and Zhenru Zhang and Zhipeng Zhou and Zihan Qiu},
      year={2025},
      eprint={2505.09388},
      archivePrefix={arXiv},
      primaryClass={cs.CL},
      url={https://arxiv.org/abs/2505.09388},
}

@misc{shao2024deepseekmathpushinglimitsmathematical,
      title={DeepSeekMath: Pushing the Limits of Mathematical Reasoning in Open Language Models},
      author={Zhihong Shao and Peiyi Wang and Qihao Zhu and Runxin Xu and Junxiao Song and Xiao Bi and Haowei Zhang and Mingchuan Zhang and Y. K. Li and Y. Wu and Daya Guo},
      year={2024},
      eprint={2402.03300},
      archivePrefix={arXiv},
      primaryClass={cs.CL},
      url={https://arxiv.org/abs/2402.03300},
}

@misc{cobbe2021trainingverifierssolvemath,
      title={Training Verifiers to Solve Math Word Problems},
      author={Karl Cobbe and Vineet Kosaraju and Mohammad Bavarian and Mark Chen and Heewoo Jun and Lukasz Kaiser and Matthias Plappert and Jerry Tworek and Jacob Hilton and Reiichiro Nakano and Christopher Hesse and John Schulman},
      year={2021},
      eprint={2110.14168},
      archivePrefix={arXiv},
      primaryClass={cs.LG},
      url={https://arxiv.org/abs/2110.14168},
}

@misc{hendrycks2021measuringmathematicalproblemsolving,
      title={Measuring Mathematical Problem Solving With the MATH Dataset},
      author={Dan Hendrycks and Collin Burns and Saurav Kadavath and Akul Arora and Steven Basart and Eric Tang and Dawn Song and Jacob Steinhardt},
      year={2021},
      eprint={2103.03874},
      archivePrefix={arXiv},
      primaryClass={cs.LG},
      url={https://arxiv.org/abs/2103.03874},
}

@misc{yu2025dapoopensourcellmreinforcement,
      title={DAPO: An Open-Source LLM Reinforcement Learning System at Scale},
      author={Qiying Yu and Zheng Zhang and Ruofei Zhu and Yufeng Yuan and Xiaochen Zuo and Yu Yue and Weinan Dai and Tiantian Fan and Gaohong Liu and Lingjun Liu and Xin Liu and Haibin Lin and Zhiqi Lin and Bole Ma and Guangming Sheng and Yuxuan Tong and Chi Zhang and Mofan Zhang and Wang Zhang and Hang Zhu and Jinhua Zhu and Jiaze Chen and Jiangjie Chen and Chengyi Wang and Hongli Yu and Yuxuan Song and Xiangpeng Wei and Hao Zhou and Jingjing Liu and Wei-Ying Ma and Ya-Qin Zhang and Lin Yan and Mu Qiao and Yonghui Wu and Mingxuan Wang},
      year={2025},
      eprint={2503.14476},
      archivePrefix={arXiv},
      primaryClass={cs.LG},
      url={https://arxiv.org/abs/2503.14476},
}

@misc{kimiteam2026kimik3openfrontier,
      title={Kimi K3: Open Frontier Intelligence},
      author={{Kimi Team} and Tongtong Bai and Yifan Bai and Yiping Bao and M. C. and Jianfeng Cai and Xinyuan Cai and Peizhou Cao and Yuxuan Cao and Ziwei Chai and Y. Charles and H. S. Che and Guanduo Chen and Guangyu Chen and Guanzheng Chen and Huarong Chen and Jia Chen and Jianlong Chen and Jun Chen and Kexin Chen and Peng Chen and Ruijue Chen and Wentao Chen and Xin Chen and Yang Chen and Yanru Chen and Yifei Chen and Yingjiang Chen and Yuankun Chen and Yujie Chen and Yutian Chen and Zhirong Chen and Dazhi Cheng and Yean Cheng and Jialei Cui and Jingbing Cui and Anqi Dai and Jiaqi Deng and Hao Ding and Rui Ding and Shaofeng Ding and Mengfan Dong and Mengnan Dong and Yuhao Dong and Yuxin Dong and Angang Du and Chenzhuang Du and Dikang Du and Jusen Du and Yulun Du and Yu Fan and Jing Feng and Qiulin Feng and Yichen Feng and Kelin Fu and Qiang Fu and Fuxuan Gao and Hongcheng Gao and Jingyue Gao and Tong Gao and Weijia Gao and Shangyi Geng and Jie Gong and Linhu Gong and Shengao Gong and Xiaochen Gong and Qizheng Gu and Yicheng Gu and Shuhao Guan and Haiqing Guo and Shiqi Guo and Xiang Guo and Zhengyan Guo and Beixi Hao and Wenxin Hao and Xiaoru Hao and Dailan He and Haotian He and Lehan He and Qi He and Weiran He and Xinran He and Xinyi He and Yibo He and Yunjia He and Chao Hong and Tiange Hong and Hao Hu and Jiaxi Hu and Ruikun Hu and Weiming Hu and Yangyang Hu and Zhenxing Hu and Liang Hua and Jinbin Huang and Ke Huang and Ruiyuan Huang and Siying Huang and Weixiao Huang and Yan Huang and Zhengjie Huang and Zhiqi Huang and Yulong Hui and Chaobo Jia and Yutong Jiang and Zhejun Jiang and Zuoyou Jiang and Wenyi Jin and Xinyi Jin and Yu Jing and Huanjun Kong and Guokun Lai and Aidi Li and Cheng Li and Chengyuan Li and Cong Li and Fang Li and Guanyu Li and Haoyang Li and Jia Li and Junxiong Li and Lei Li and Letian Li and Lincan Li and Weihong Li and Wentao Li and Xintong Li and Yang Li and Yishen Li and Yiwei Li and Yuxiao Li and Zhaowei Li and Zhaoxi Li and Zheming Li and Zhengxiao Li and Zhiyuan Li and Jiawei Lin and Xiaohan Lin and Yibo Lin and Zichao Lin and Ziyan Lin and Bill Liu and Boxiao Liu and Chuan Liu and Liang Liu and Shaowei Liu and Shudong Liu and Shuran Liu and Tianwei Liu and Weizhou Liu and Yangyang Liu and Yanming Liu and Yibo Liu and Yipeng Liu and Zhengying Liu and Zhiheng Liu and Enzhe Lu and Haoyu Lu and Linqiang Lu and Tingzhan Lu and Zhiyuan Lu and Aotian Luo and G. Luo and Junyu Luo and Yifan Luo and B. Lyu and Wenzhou Lyu and Shaoguang Mao and Yuan Mei and Xin Men and Minqing Ni and Yixuan Niu and Siyuan Pan and Shujun Peng and Zhangyang Qi and Ruoyu Qin and ZeChao Qin and Zeyu Qin and Haiquan Qiu and Jianxin Qiu and Jiezhong Qiu and Bowen Qu and Yuhao Qu and Zeyu Shang and Youbo Shao and Han Shen and Jincheng Shi and Juanfeng Shi and Lidong Shi and Shengyuan Shi and Wingchun Siu and Pengwei Song and Xiaoxi Song and Jianlin Su and Yunfeng Su and Zhaochen Su and Lin Sui and Jingsong Sun and Junyao Sun and Shaoning Sun and Shuzhe Sun and Tongyu Sun and Yujun Sun and Yunpeng Tai and Chuning Tang and Heyi Tang and Sirui Tang and Zecheng Tang and Chaoran Tian and Rongpeng Tian and Yu Tian and Wei Tu and Chensi Wang and Chuang Wang and Chunjie Wang and Dinglu Wang and Feng Wang and Hailong Wang and Haiming Wang and Hao Wang and Hao Wang and Huaqing Wang and Hui Wang and Jiayi Wang and Jinglong Wang and Jinhong Wang and Jiuzheng Wang and Linian Wang and Shaobo Wang and Shenzhi Wang and Shuyi Wang and Si Wang and Siyuan Wang and Tianfu Wang and Wenjue Wang and Xingran Wang and Xinmei Wang and Xinyuan Wang and Xusheng Wang and Yalin Wang and Yangkun Wang and Yao Wang and Yaoyu Wang and Yejie Wang and Yiqin Wang and Yucheng Wang and Yuzhi Wang and Zhaoji Wang and Zhaowei Wang and Zhengtao Wang and Zhenhao Wang and Zhongsheng Wang and Zifan Wang and Chu Wei and Ming Wei and Shouxin Wei and Zichen Wen and Fan Wu and Haoning Wu and Rucong Wu and Wenhao Wu and Xiaoxue Wu and Yingcong Wu and Yongqi Wu and Yuxin Wu and Zijian Wu and Xinglang Xian and Chenxuan Xiang and Yuye Xiang and Bocheng Xiao and Chenjun Xiao and Xin Xiao and Jin Xie and Xiaotong Xie and Yifeng Xie and Zhe Xie and Bowei Xing and Yiming Xiong and Baosheng Xu and Boyu Xu and Jiale Xu and Jianfan Xu and Jing Xu and Jinjing Xu and L. H. Xu and Qingtao Xu and Shuyao Xu and Suting Xu and Tiantian Xu and Tianxiang Xu and Weixin Xu and Xinran Xu and Yangchuan Xu and Ye Xu and Yueni Xu and Ziyao Xu and Haonan Xue and Junjie Yan and Yaoyao Yan and Fan Yang and Guangyao Yang and Hao Yang and Junwei Yang and Ruoyu Yang and Wenjie Yang and Xiaofei Yang and Xinyu Yang and Yi Yang and Yiling Yang and Ying Yang and Yuchen Yang and Zhen Yang and Zhilin Yang and Zian Yang and Zuhao Yang and Haotian Yao and Dan Ye and Haoran Ye and Wenjie Ye and Zhanbo Ye and Bohong Yin and Haoxiang Yin and Xietong Yin and Chengzhen Yu and Haozhen Yu and Longhui Yu and Shengnan Yu and Shuying Yu and Tianxiang Yu and Enming Yuan and Mengjie Yuan and Tongtian Yue and Wei Yue and Yang Yue and Dunyuan Zha and Haobing Zhan and B. H. Zhang and Dehao Zhang and Fei Zhang and Hao Zhang and Haoyuan Zhang and Huanyu Zhang and Jiapei Zhang and Jiaxuan Zhang and Jin Zhang and Kaiyi Zhang and Miaozhen Zhang and Puqi Zhang and Qinglei Zhang and Rong Zhang and Rui Zhang and Shaoshuai Zhang and Shiyi Zhang and Xiaobin Zhang and Xiaoyun Zhang and Y. Zhang and Yangkun Zhang and Ye Zhang and Yichi Zhang and Yikun Zhang and Yizhi Zhang and Yongting Zhang and Yu Zhang and Yutao Zhang and Yutong Zhang and Zheng Zhang and Zijing Zhang and Bin Zhao and Chenguang Zhao and Feifan Zhao and Jinglun Zhao and Jinxiang Zhao and Shuai Zhao and Wenshuo Zhao and Xiangyu Zhao and Xuanle Zhao and Yikai Zhao and Zijia Zhao and Haozhi Zheng and Huabin Zheng and Ruihan Zheng and Shaojie Zheng and Tengyang Zheng and Haofeng Zhong and Lei Zhong and Longguang Zhong and M. Zhou and Qiankang Zhou and Runjie Zhou and Ruozhang Zhou and Xinyu Zhou and Yiqiao Zhou and Zaida Zhou and Jinguo Zhu and Liya Zhu and Xinhao Zhu and Yangjunfeng Zhu and Yuxuan Zhu and Zhen Zhu and Chen Zhuang and Weiyu Zhuang and Xinxing Zu},
      year={2026},
      eprint={2607.24653},
      archivePrefix={arXiv},
      primaryClass={cs.CL},
      url={https://arxiv.org/abs/2607.24653},
}

@article{lu2025onpolicydistillation,
  author = {Kevin Lu and {Thinking Machines Lab}},
  title = {On-Policy Distillation},
  journal = {Thinking Machines Lab: Connectionism},
  year = {2025},
  note = {https://thinkingmachines.ai/blog/on-policy-distillation},
  doi = {10.64434/tml.20251026},
}

@misc{shen2026geometryonpolicydistillation,
      title={On the Geometry of On-Policy Distillation},
      author={Zhennan Shen and Yanshu Li and Qingyu Yin and Chak Tou Leong and Zhilin Wang and Yanxu Chen and Rongduo Han and Sunbowen Lee and Yi R. Fung},
      year={2026},
      eprint={2606.07082},
      archivePrefix={arXiv},
      primaryClass={cs.LG},
      url={https://arxiv.org/abs/2606.07082},
}

@misc{zhang2025interplaypretrainingmidtrainingrl,
      title={On the Interplay of Pre-Training, Mid-Training, and RL on Reasoning Language Models},
      author={Charlie Zhang and Graham Neubig and Xiang Yue},
      year={2025},
      eprint={2512.07783},
      archivePrefix={arXiv},
      primaryClass={cs.CL},
      url={https://arxiv.org/abs/2512.07783},
}

@article{zhu2026sft,
  title={SFT Conflicts, RL Coexists: A Theoretical and Empirical Analysis of Multi-Task Learning for LLMs},
  author={Zhu, Kejian and Jin, Zhuoran and Tu, Shangqing and Yuan, Hongbang and Bai, Yushi and Liu, Kang and Li, Juanzi and Zhao, Jun},
  journal={arXiv preprint arXiv:2608.03573},
  year={2026}
}

@article{yang2026orthogonal,
  title={Orthogonal model merging},
  author={Yang, Sihan and Shi, Kexuan and Liu, Weiyang},
  journal={arXiv preprint arXiv:2602.05943},
  year={2026}
}

@inproceedings{yuan2026f,
  title={From f (x) and g (x) to f (g (x)): Llms learn new skills in rl by composing old ones},
  author={Yuan, Lifan and Chen, Weize and Zhang, Yuchen and Cui, Ganqu and Wang, Hanbin and You, Ziming and Ding, Ning and Liu, Zhiyuan and Sun, Maosong and Peng, Hao},
  booktitle={International Conference on Learning Representations},
  volume={2026},
  pages={147547-147574},
  year={2026}
}

@article{power2022grokking,
  title={Grokking: Generalization beyond overfitting on small algorithmic datasets},
  author={Power, Alethea and Burda, Yuri and Edwards, Harri and Babuschkin, Igor and Misra, Vedant},
  journal={arXiv preprint arXiv:2201.02177},
  year={2022}
}

@article{chen2021evaluatingcode,
  title={Evaluating Large Language Models Trained on Code},
  author={Chen, Mark and Tworek, Jerry and Jun, Heewoo and Yuan, Qiming and de Oliveira Pinto, Henrique Ponde and Kaplan, Jared and Edwards, Harri and Burda, Yuri and Joseph, Nicholas and Brockman, Greg and others},
  journal={arXiv preprint arXiv:2107.03374},
  year={2021},
  url={https://arxiv.org/abs/2107.03374}
}

@article{austin2021programsynthesis,
  title={Program Synthesis with Large Language Models},
  author={Austin, Jacob and Odena, Augustus and Nye, Maxwell and Bosma, Maarten and Michalewski, Henryk and Dohan, David and Jiang, Ellen and Cai, Carrie and Terry, Michael and Le, Quoc and Sutton, Charles},
  journal={arXiv preprint arXiv:2108.07732},
  year={2021},
  url={https://arxiv.org/abs/2108.07732}
}

@article{liu2023evalplus,
  title={Is Your Code Generated by {ChatGPT} Really Correct? Rigorous Evaluation of Large Language Models for Code Generation},
  author={Liu, Jiawei and Xia, Chunqiu Steven and Wang, Yuyao and Zhang, Lingming},
  journal={arXiv preprint arXiv:2305.01210},
  year={2023},
  url={https://arxiv.org/abs/2305.01210}
}

@article{bai2024longbenchv2,
  title={{LongBench v2}: Towards Deeper Understanding and Reasoning on Realistic Long-context Multitasks},
  author={Bai, Yushi and Tu, Shangqing and Zhang, Jiajie and Peng, Hao and Wang, Xiaozhi and Lv, Xin and Cao, Shulin and Xu, Jiazheng and Hou, Lei and Dong, Yuxiao and Tang, Jie and Li, Juanzi},
  journal={arXiv preprint arXiv:2412.15204},
  year={2024},
  url={https://arxiv.org/abs/2412.15204}
}

@inproceedings{patil2025bfcl,
  title={The Berkeley Function Calling Leaderboard ({BFCL}): From Tool Use to Agentic Evaluation of Large Language Models},
  author={Patil, Shishir G. and Mao, Huanzhi and Ji, Charlie Cheng-Jie and Yan, Fanjia and Suresh, Vishnu and Stoica, Ion and Gonzalez, Joseph E.},
  booktitle={Forty-second International Conference on Machine Learning},
  year={2025},
  url={https://openreview.net/forum?id=2GmDdhBdDk}
}

@article{zhou2025gsminfinite,
  title={{GSM-Infinite}: How Do Your {LLMs} Behave over Infinitely Increasing Context Length and Reasoning Complexity?},
  author={Zhou, Yang and Liu, Hongyi and Chen, Zhuoming and Tian, Yuandong and Chen, Beidi},
  journal={arXiv preprint arXiv:2502.05252},
  year={2025},
  url={https://arxiv.org/abs/2502.05252}
}

@article{zhao2023docmath,
  title={{DocMath-Eval}: Evaluating Math Reasoning Capabilities of {LLMs} in Understanding Long and Specialized Documents},
  author={Zhao, Yilun and Long, Yitao and Liu, Hongjun and Kamoi, Ryo and Nan, Linyong and Chen, Lyuhao and Liu, Yixin and Tang, Xiangru and Zhang, Rui and Cohan, Arman},
  journal={arXiv preprint arXiv:2311.09805},
  year={2023},
  url={https://arxiv.org/abs/2311.09805}
}

@inproceedings{jimenez2024swebench,
  title={{SWE-bench}: Can Language Models Resolve Real-World {GitHub} Issues?},
  author={Jimenez, Carlos E. and Yang, John and Wettig, Alexander and Yao, Shunyu and Pei, Kexin and Press, Ofir and Narasimhan, Karthik R.},
  booktitle={The Twelfth International Conference on Learning Representations},
  year={2024},
  url={https://openreview.net/forum?id=VTF8yNQM66}
}

@article{anil2022lengthgeneralization,
  title={Exploring Length Generalization in Large Language Models},
  author={Anil, Cem and Wu, Yuhuai and Andreassen, Anders and Lewkowycz, Aitor and Misra, Vedant and Ramasesh, Vinay and Slone, Ambrose and Gur-Ari, Guy and Dyer, Ethan and Neyshabur, Behnam},
  journal={arXiv preprint arXiv:2207.04901},
  year={2022},
  url={https://arxiv.org/abs/2207.04901}
}

@inproceedings{lake2018systematicity,
  title={Generalization without Systematicity: On the Compositional Skills of Sequence-to-Sequence Recurrent Networks},
  author={Lake, Brenden and Baroni, Marco},
  booktitle={Proceedings of the 35th International Conference on Machine Learning},
  volume={80},
  pages={2873-2882},
  year={2018},
  publisher={PMLR},
  url={https://proceedings.mlr.press/v80/lake18a.html}
}

@article{hinton2015distilling,
  title={Distilling the Knowledge in a Neural Network},
  author={Hinton, Geoffrey and Vinyals, Oriol and Dean, Jeff},
  journal={arXiv preprint arXiv:1503.02531},
  year={2015},
  url={https://arxiv.org/abs/1503.02531}
}

@misc{xing2026trustregion,
  title = {Trust Region On-Policy Distillation},
  author = {Xing, Xingrun and Wang, Haoqing and Gao, Boyan and Li, Ziheng and Tang, Yehui},
  year = {2026},
  eprint = {2606.01249},
  archivePrefix = {arXiv},
  primaryClass = {cs.LG},
  url = {https://arxiv.org/abs/2606.01249v1}
}

@misc{liao2026prefixreplay,
  title = {Multi-Turn On-Policy Distillation with Prefix Replay},
  author = {Baohao Liao and Hanze Dong and Christof Monz and Xinxing Xu and Li Dong and Furu Wei},
  year = {2026},
  eprint = {2607.04763},
  archivePrefix = {arXiv},
  primaryClass = {cs.LG},
  url = {https://arxiv.org/abs/2607.04763}
}

@misc{wu2026seed,
  title = {{SEED}: Self-Evolving On-Policy Distillation for Agentic Reinforcement Learning},
  author = {Jinyang Wu and Shuo Yang and Zhengxi Lu and Fan Zhang and Yuhao Shen and Lang Feng and Haoran Luo and Zheng Lian and Shuai Zhang and Zhengqi Wen and Jianhua Tao},
  year = {2026},
  eprint = {2607.14777},
  archivePrefix = {arXiv},
  primaryClass = {cs.CL},
  url = {https://arxiv.org/abs/2607.14777}
}

@misc{gao2026openmopd,
  title = {{Open-MOPD}: Diagnosing and Fixing Capability Imbalance in Multi-Teacher On-Policy Distillation},
  author = {Gao, Huan-ang and Chi, Haohan and Yan, Yong and Feng, Shiyuan and Wu, Hanlin and Jiang, Zheng and He, Bingxiang and Ma, Wei-Ying and Zhang, Ya-Qin and Zhou, Hao},
  year = {2026},
  eprint = {2608.19098},
  archivePrefix = {arXiv},
  primaryClass = {cs.LG},
  url = {https://arxiv.org/abs/2608.19098}
}

@misc{li2026everycoin,
  title = {Every Coin Has Two Sides: On the Dual Nature of Generalization in On-Policy Distillation of Large Language Models},
  author = {Zhaoyi Li and Deyang Kong and Yuan Wei and Evan Yang and Ranran Shen and Mahardika Krisna Ihsani and Ming Yang and Wei Zhang and Chuan Hao and Jian Yang and Ran Tao and Bryan Dai and Shikun Zhang and Wei Ye and Ying Wei and Defu Lian},
  year = {2026},
  eprint = {2608.16647},
  archivePrefix = {arXiv},
  primaryClass = {cs.CL},
  url = {https://arxiv.org/abs/2608.16647}
}

@misc{jiang2026trajectoryrefined,
  title = {Trajectory-Refined Distillation},
  author = {Jiang, Li and Xu, Haoran and Ding, Yichuan and Zhang, Amy},
  year = {2026},
  eprint = {2606.08432},
  archivePrefix = {arXiv},
  primaryClass = {cs.AI},
  url = {https://arxiv.org/abs/2606.08432}
}

@inproceedings{he2024learningtogrok,
  title = {Learning to grok: Emergence of in-context learning and skill composition in modular arithmetic tasks},
  author = {He, Tianyu and Doshi, Darshil and Das, Aritra and Gromov, Andrey},
  booktitle = {Advances in Neural Information Processing Systems},
  year = {2024},
  url = {https://arxiv.org/abs/2406.02550}
}

@inproceedings{hu2026breakingbarriers,
  title = {Breaking Barriers: Do Reinforcement Post Training Gains Transfer To Unseen Domains?},
  author = {Hu, Chuxuan and Zhu, Yuxuan and Kellermann, Antony and Biddulph, Caleb and Waiwitlikhit, Suppakit and Benn, Jason and Kang, Daniel},
  booktitle = {The Fourteenth International Conference on Learning Representations},
  year = {2026},
  url = {https://arxiv.org/abs/2506.19733}
}

@misc{fu2026compositional,
  title = {Reinforcement Learning for Compositional Generalization with Outcome-Level Optimization},
  author = {Fu, Xiyan and Liu, Wei},
  year = {2026},
  eprint = {2605.04920},
  archivePrefix = {arXiv},
  primaryClass = {cs.LG},
  url = {https://arxiv.org/abs/2605.04920}
}

@inproceedings{ainsworth2023gitrebasin,
  title = {{Git Re-Basin}: Merging Models modulo Permutation Symmetries},
  author = {Ainsworth, Samuel K. and Hayase, Jonathan and Srinivasa, Siddhartha},
  booktitle = {The Eleventh International Conference on Learning Representations},
  year = {2023},
  url = {https://arxiv.org/abs/2209.04836}
}

@inproceedings{yadav2023tiesmerging,
  title = {{TIES-Merging}: Resolving Interference When Merging Models},
  author = {Yadav, Prateek and Tam, Derek and Choshen, Leshem and Raffel, Colin and Bansal, Mohit},
  booktitle = {Advances in Neural Information Processing Systems},
  year = {2023},
  url = {https://proceedings.neurips.cc/paper_files/paper/2023/hash/1644c9af28ab7916874f6fd6228a9bcf-Abstract-Conference.html}
}

@misc{zhou2026mergeability,
  title = {Demystifying Mergeability: Interpretable Properties to Predict Model Merging Success},
  author = {Zhou, Luca and Zhao, Bo and Yu, Rose and Rodol{\`a}, Emanuele},
  year = {2026},
  eprint = {2601.22285},
  archivePrefix = {arXiv},
  primaryClass = {cs.LG},
  url = {https://arxiv.org/abs/2601.22285}
}

@inproceedings{kim2023taskweb,
  title = {{TaskWeb}: Selecting Better Source Tasks for Multi-task {NLP}},
  author = {Kim, Joongwon and Asai, Akari and Ilharco, Gabriel and Hajishirzi, Hannaneh},
  booktitle = {Proceedings of the 2023 Conference on Empirical Methods in Natural Language Processing},
  year = {2023},
  pages = {11032-11052},
  publisher = {Association for Computational Linguistics},
  doi = {10.18653/v1/2023.emnlp-main.680},
  url = {https://aclanthology.org/2023.emnlp-main.680/}
}

@inproceedings{krishna2025latenttraits,
  title = {Latent Traits and Cross-Task Transfer: Deconstructing Dataset Interactions in {LLM} Fine-tuning},
  author = {Krishna, Shambhavi and Naik, Atharva and Agarwal, Chaitali and Govindan, Sudharshan and Lee, Taesung and Chang, Haw-Shiuan},
  booktitle = {Proceedings of the 14th Joint Conference on Lexical and Computational Semantics (*SEM 2025)},
  year = {2025},
  url = {https://arxiv.org/abs/2509.13624}
}

@misc{deng2026qwenscope,
  title = {{Qwen-Scope}: Turning Sparse Features into Development Tools for Large Language Models},
  author = {Deng, Boyi and Wang, Xu and Wang, Yaoning and Wan, Yu and Ma, Yubo and Yang, Baosong and Wei, Haoran and Tang, Jialong and Lin, Huan and Gao, Ruize and Li, Tianhao and Cao, Qian and Ren, Xuancheng and Deng, Xiaodong and Yang, An and Huang, Fei and Liu, Dayiheng and Zhou, Jingren},
  year = {2026},
  eprint = {2605.11887},
  archivePrefix = {arXiv},
  primaryClass = {cs.CL},
  url = {https://arxiv.org/abs/2605.11887}
}

@inproceedings{ilharco2023taskarithmetic,
  title = {Editing Models with Task Arithmetic},
  author = {Ilharco, Gabriel and Ribeiro, Marco Tulio and Wortsman, Mitchell and Gururangan, Suchin and Schmidt, Ludwig and Hajishirzi, Hannaneh and Farhadi, Ali},
  booktitle = {The Eleventh International Conference on Learning Representations},
  year = {2023},
  url = {https://arxiv.org/abs/2212.04089}
}

@inproceedings{yu2024dare,
  title = {Language Models are {Super Mario}: Absorbing Abilities from Homologous Models as a Free Lunch},
  author = {Yu, Le and Yu, Bowen and Yu, Haiyang and Huang, Fei and Li, Yongbin},
  booktitle = {Proceedings of the 41st International Conference on Machine Learning},
  year = {2024},
  url = {https://arxiv.org/abs/2311.03099}
}

@misc{lambert2024tulu3,
  title = {{Tulu 3}: Pushing Frontiers in Open Language Model Post-Training},
  author = {Lambert, Nathan and Morrison, Jacob and Pyatkin, Valentina and Huang, Shengyi and Ivison, Hamish and Brahman, Faeze and Miranda, Lester James V. and Liu, Alisa and Dziri, Nouha and Lyu, Shane and Gu, Yuling and Malik, Saumya and Graf, Victoria and Hwang, Jena D. and Yang, Jiangjiang and Le Bras, Ronan and Tafjord, Oyvind and Wilhelm, Chris and Soldaini, Luca and Smith, Noah A. and Wang, Yizhong and Dasigi, Pradeep and Hajishirzi, Hannaneh},
  year = {2024},
  eprint = {2411.15124},
  archivePrefix = {arXiv},
  primaryClass = {cs.CL},
  url = {https://arxiv.org/abs/2411.15124}
}

@article{lightman2023lets,
  title={Let's Verify Step by Step},
  author={Lightman, Hunter and Kosaraju, Vineet and Burda, Yura and Edwards, Harri and Baker, Bowen and Lee, Teddy and Leike, Jan and Schulman, John and Sutskever, Ilya and Cobbe, Karl},
  journal={arXiv preprint arXiv:2305.20050},
  year={2023},
  url={https://arxiv.org/abs/2305.20050}
}

@misc{aimo2024amc,
  title={{AIMO Validation AMC}},
  author={{AI-MO}},
  year={2024},
  howpublished={Hugging Face dataset},
  url={https://huggingface.co/datasets/AI-MO/aimo-validation-amc},
  note={Adapted AMC 12 problems from 2022 and 2023}
}

@misc{huggingfaceh42024aime,
  title={{AIME 2024}},
  author={{Hugging Face H4}},
  year={2024},
  howpublished={Hugging Face dataset},
  url={https://huggingface.co/datasets/HuggingFaceH4/aime_2024}
}

@misc{opencompass2025aime,
  title={{AIME 2025}},
  author={{OpenCompass}},
  year={2025},
  howpublished={Hugging Face dataset},
  url={https://huggingface.co/datasets/opencompass/AIME2025}
}

@misc{nvidia2025nemotronrlcoding,
  title={{Nemotron-RL-coding-competitive\_coding}},
  author={{NVIDIA}},
  year={2025},
  howpublished={Hugging Face dataset},
  url={https://huggingface.co/datasets/nvidia/Nemotron-RL-coding-competitive_coding}
}

@article{liu2024apigen,
  title={{APIGen}: Automated Pipeline for Generating Verifiable and Diverse Function-Calling Datasets},
  author={Liu, Zuxin and Hoang, Thai and Zhang, Jianguo and Zhu, Ming and Lan, Tian and Kokane, Shirley and Tan, Juntao and Yao, Weiran and Liu, Zhiwei and Feng, Yihao and others},
  journal={arXiv preprint arXiv:2406.18518},
  year={2024},
  url={https://arxiv.org/abs/2406.18518}
}

@article{liu2024toolace,
  title={{ToolACE}: Winning the Points of {LLM} Function Calling},
  author={Liu, Weiwen and Huang, Xu and Zeng, Xingshan and Hao, Xinlong and Yu, Shuai and Li, Dexun and Wang, Shuai and Gan, Weinan and Liu, Zhengying and Yu, Yuanqing and Wang, Zezhong and Wang, Yuxian and Ning, Wu and Hou, Yutai and Wang, Bin and Wu, Chuhan and Wang, Xinzhi and Liu, Yong and Wang, Yasheng and Tang, Duyu and Tu, Dandan and Shang, Lifeng and Jiang, Xin and Tang, Ruiming and Lian, Defu and Liu, Qun and Chen, Enhong},
  journal={arXiv preprint arXiv:2409.00920},
  year={2024},
  url={https://arxiv.org/abs/2409.00920}
}

@article{wan2025qwenlongl1,
  title={{QwenLong-L1}: Towards Long-Context Large Reasoning Models with Reinforcement Learning},
  author={Wan, Fanqi and Shen, Weizhou and Liao, Shengyi and Shi, Yingcheng and Li, Chenliang and Yang, Ziyi and Zhang, Ji and Huang, Fei and Zhou, Jingren and Yan, Ming},
  journal={arXiv preprint arXiv:2505.17667},
  year={2025},
  url={https://arxiv.org/abs/2505.17667}
}

@misc{openr12025math,
  title={{OpenR1-Math-220k}},
  author={{Open R1 Team}},
  year={2025},
  howpublished={Hugging Face dataset},
  url={https://huggingface.co/datasets/open-r1/OpenR1-Math-220k}
}

@inproceedings{phan2026beyondreasoninggains,
  title = {Beyond Reasoning Gains: Mitigating General-Capability Forgetting in Large Reasoning Models},
  author = {Phan, Hoang and Yang, Xianjun and Yao, Yuanshun and Zhang, Jingyu and Bi, Shengjie and Tang, Xiaocheng and Khabsa, Madian and Liu, Lijuan and Lei, Deren},
  booktitle = {Findings of the Association for Computational Linguistics: ACL 2026},
  year = {2026},
  month = jul,
  pages = {34368-34398},
  publisher = {Association for Computational Linguistics},
  doi = {10.18653/v1/2026.findings-acl.1717},
  url = {https://aclanthology.org/2026.findings-acl.1717/}
}

@inproceedings{li2026choicedivergence,
  title = {The Choice of Divergence: A Neglected Key to Mitigating Diversity Collapse in Reinforcement Learning with Verifiable Reward},
  author = {Li, Long and Zhou, Zhijian and Hao, Jiaran and Liu, Jason Klein and Miao, Yanting and Pang, Wei and Tan, Xiaoyu and Chu, Wei and Wang, Zhe and Pan, Shirui and Qu, Chao and Qi, Yuan},
  booktitle = {International Conference on Learning Representations},
  year = {2026},
  eprint = {2509.07430},
  archivePrefix = {arXiv},
  url = {https://arxiv.org/abs/2509.07430}
}

@misc{motwani2025h1,
  title = {{h1}: Bootstrapping {LLMs} to Reason over Longer Horizons via Reinforcement Learning},
  author = {Motwani, Sumeet Ramesh and Ivanova, Alesia and Cai, Ziyang and Torr, Philip and Islam, Riashat and Shah, Shital and Schroeder de Witt, Christian and London, Charles},
  year = {2025},
  eprint = {2510.07312},
  archivePrefix = {arXiv},
  primaryClass = {cs.LG},
  url = {https://arxiv.org/abs/2510.07312}
}

@misc{cui2026clipo,
  title = {{CLIPO}: Contrastive Learning in Policy Optimization Generalizes {RLVR}},
  author = {Cui, Sijia and Cheng, Pengyu and Song, Jiajun and Gai, Yongbo and Zhang, Guojun and Yu, Zhechao and Lin, Jianhe and Jiang, Xiaoxi and Jiang, Guanjun},
  year = {2026},
  eprint = {2603.10101},
  archivePrefix = {arXiv},
  primaryClass = {cs.LG},
  url = {https://arxiv.org/abs/2603.10101}
}

@misc{lee2026beyondcorrectness,
  title = {Beyond Correctness: Learning Robust Reasoning via Transfer},
  author = {Lee, Hyunseok and Abbasloo, Soheil and Tack, Jihoon and Shin, Jinwoo},
  year = {2026},
  eprint = {2602.08489},
  archivePrefix = {arXiv},
  primaryClass = {cs.LG},
  url = {https://arxiv.org/abs/2602.08489}
}

@article{ujjain2021adsorption,
  title   = {Adsorption separation of heavier isotope gases
             in subnanometer carbon pores},
  author  = {Ujjain, Sanjeev Kumar and Bagusetty, Abhishek
             and Matsuda, Yuki and Tanaka, Hideki
             and Ahuja, Preety and de Tomas, Carla
             and Sakai, Motomu and Vallejos-Burgos, Fernando
             and Futamura, Ryusuke and Suarez-Martinez, Irene
             and Matsukata, Masahiko and Kodama, Akio
             and Garberoglio, Giovanni and Gogotsi, Yury
             and Johnson, J. Karl and Kaneko, Katsumi},
  journal = {Nature Communications},
  volume  = {12},
  pages   = {546},
  year    = {2021},
  doi     = {10.1038/s41467-020-20744-6}
}

@inproceedings{ortizjimenez2023tangent,
  title = {Task Arithmetic in the Tangent Space: Improved Editing of Pre-Trained Models},
  author = {Ortiz-Jim{\'e}nez, Guillermo and Favero, Alessandro and Frossard, Pascal},
  booktitle = {Advances in Neural Information Processing Systems},
  volume = {36},
  year = {2023},
  url = {https://proceedings.neurips.cc/paper_files/paper/2023/hash/d28077e5ff52034cd35b4aa15320caea-Abstract.html}
}

@article{schonemann1966procrustes,
  author = {Sch{\"o}nemann, Peter H.},
  title = {A Generalized Solution of the Orthogonal {Procrustes} Problem},
  journal = {Psychometrika},
  volume = {31},
  number = {1},
  pages = {1-10},
  year = {1966},
  doi = {10.1007/BF02289451},
  url = {https://doi.org/10.1007/BF02289451}
}

@misc{kim2025rldistillation,
  title = {Reinforcement Learning vs. Distillation: Understanding Accuracy and Capability in {LLM} Reasoning},
  author = {Kim, Minwu and Shrestha, Anubhav and Shrestha, Safal and Nepal, Aadim and Ross, Keith},
  year = {2025},
  eprint = {2505.14216},
  archivePrefix = {arXiv},
  primaryClass = {cs.AI},
  url = {https://arxiv.org/abs/2505.14216}
}

@inproceedings{guo2023predictfuture,
  title = {Predict the Future from the Past? On the Temporal Data Distribution Shift in Financial Sentiment Classifications},
  author = {Guo, Yue and Hu, Chenxi and Yang, Yi},
  booktitle = {Proceedings of the 2023 Conference on Empirical Methods in Natural Language Processing},
  year = {2023},
  publisher = {Association for Computational Linguistics},
  doi = {10.18653/v1/2023.emnlp-main.65},
  url = {https://aclanthology.org/2023.emnlp-main.65/}
}

@inproceedings{jesson2025generativeicl,
  title = {Can Generative {AI} Solve Your In-Context Learning Problem? A Martingale Perspective},
  author = {Jesson, Andrew and Beltran-Velez, Nicolas and Blei, David},
  booktitle = {The Thirteenth International Conference on Learning Representations},
  year = {2025},
  url = {https://openreview.net/forum?id=bcynT7s2du}
}
\clearpage
\beginappendix
\setcounter{figure}{0}
\setcounter{table}{0}
\renewcommand{\thefigure}{A\arabic{figure}}
\renewcommand{\thetable}{A\arabic{table}}
%
%

\section{Experimental Details}
\label{app:protocols}

\subsection{Training and Teacher Assignment}
\label{app:training-protocols}
Self-RL specialists and their OPD students start independently from the same Qwen3-8B base. Each specialist supervises its own training distribution. External teachers are Qwen3-32B for mathematical reasoning, QwenLong-L1.5-\allowbreak30B-A3B for long context, and Qwen3-30B-A3B-Thinking-2507 for coding and tool use. Code plus agent uses one shared teacher; the other composed settings use separate teachers for the two components.
We select external teachers from a pool of open-source Qwen3-family models (including domain-specific models) and fine-tuned models released on Hugging Face that are based on them, with compatible vocabularies for token-level distillation. Among candidates whose sizes fit our compute budget, we choose the best-performing model on the relevant benchmarks under our controlled inference setup. Scores under this setup may differ slightly from published results.

In \Cref{eq:opd-objective}, both policies are evaluated at student-generated histories, and the teacher is fixed. Interactive histories include tool observations, but the loss sums only over model-generated tokens. Mixed RL uses the combination of single-task GRPO recipes. 
The highest mean IND score among evaluated configurations selects each method's recipe and checkpoint.
In Mixed RL experiments, the component recipes are selected for IND performance comparable to MOPD, then combined with equal data proportions and random shuffling, without additional tuning of balance. 
External teachers may have additional prior training.


The main extrapolative and compositional generalization experiments use a learning rate of $10^{-6}$. 
(M)OPD mainly uses the \texttt{k1} distillation loss with policy gradients enabled and task rewards disabled.
We smoke test that different distillation loss have non-significant difference in performance after sufficient training steps in our experiments. 
MOPD selects the teacher by data source. RL uses GRPO with one PPO epoch per update, a KL loss coefficient of $0.001$, and an entropy coefficient of $0$. (M)OPD rollouts in these experiments use temperature $1.0$ and top-$p$ $1.0$.

\begin{table}[!htbp]
\centering\small
\setlength{\tabcolsep}{4pt}
\caption{Training configurations for the main extrapolative and compositional generalization experiments. Paired entries are (M)OPD / RL. Batch size counts prompts; PPO mini-batch size is the configured value. Lengths are in tokens.}
\label{tab:training-configurations}
\begin{tabular}{@{}lcccrr@{}}
\toprule
Setting & \makecell{Train\\batch} & \makecell{PPO\\mini-batch} & \makecell{Responses\\per prompt} & \makecell{Max.\\prompt} & \makecell{Max.\\response}\\
\midrule
GSM8K & 16 / 64 & 16 / 8 & 1 / 8 & 1,024 & 8,192\\
DAPO & 16 / 64 & 16 / 8 & 1 / 8 & 2,048 & 16,384\\
Other math (single/mixed) & 16 / 64 & 16 / 8 & 1 / 8 & 2,048 & 8,192\\
Code \& agent (single/mixed) & 32 / 32 & 16 / 16 & 1 / 8 & 8,192 & 12,288\\
Long context (single/mixed) & 16 / 16 & 16 / 16 & 1 / 8 & 59,561 & 16,384\\
\bottomrule
\end{tabular}
\par\smallskip
\begin{minipage}{\linewidth}\footnotesize
$^{a}$ Long context tasks use YaRN with scaling factor $4$.
\end{minipage}
\end{table}

\subsection{Data and Evaluation Protocols}
\label{app:datasets}
Training datasets and benchmark citations appear in \Cref{sec:data-benchmarks}. Nemotron competitive-coding denotes \texttt{nvidia/}\allowbreak\texttt{Nemotron-RL-coding-competitive\_coding}; DocQA denotes \texttt{Tongyi-Zhiwen/}\allowbreak\texttt{DocQA-RL-1.6K}. Mathematical reasoning comparisons use GSM8K, MATH, and DAPO (held-out set from training), AMC23 and AIME24/25.
All evaluation benchmarks are checked to be held out from all training datasets.

%
%
\Cref{tab:evaluation-configurations} summarizes generation configs in evaluations, i.e., budgets, sampling, and prompting. EvalPlus v0.3.1 reports pass@1 under the protocol. Completed thinking blocks are removed before scoring; unclosed blocks are invalid.
\textbf{All generation (evaluation and rollouts in training) are thinking enabled in the experiments.}

\begin{table}[!htbp]
\centering\small
\setlength{\tabcolsep}{4pt}
\renewcommand{\arraystretch}{1.13}
\caption{Evaluation configurations. Generation and input limits are in tokens; context limits include input and generated tokens. }
\label{tab:evaluation-configurations}
\begin{tabular}{@{}p{0.24\linewidth}r p{0.54\linewidth}@{}}
\toprule
Benchmark & \makecell{Max. generated\\tokens} & Sampling and prompting\\
\midrule
GSM8K & 8,192 & Mainly 1 sample per problem; sample counts follow the corresponding run configuration.\\
MATH500 / AMC23 & 32,768 & Mainly 1 sample per problem; sample counts follow the corresponding run configuration.\\
AIME24 / AIME25 & 32,768 & 8 samples per problem; main metric: avg@8.\\
HumanEval(+) / MBPP(+) & 32,768 & Main metric: pass@1; context limit 40,960.\\
BFCL-v4 & 8,192 & \texttt{qwen\_fc} handler.\\
LongBench-v2 & 50,000 & Input limit 81,000; context limit 131,072; Qwen3 uses YaRN $\times4$.\\
DocMath & 50,000 & Temperature $0.7$, top-$p$ $0.95$ following the original benchmark settings; input limit 131,072; Qwen3 uses YaRN $\times6$ and context limit 196,608.\\
\bottomrule
\end{tabular}
\end{table}

For a given model and evaluation task, sampling parameters are fixed across evaluation runs.
For mathematical reasoning, EvalPlus, BFCL, and LongBench-v2, parameters not explicitly overridden in the task's evaluation configuration follow the model's defaults in \texttt{generation\_config.json}.

LongBench-v2 component results primarily use its Hard subset.
Composed-task evaluations use GSM-Infinite's 8K subset and DocMath test-mini for long-context math, LongBench-v2 Code for long-context coding, and SWE-bench Verified for agentic coding. 
Model comparisons are under matched benchmark subsets and scoring rules.

\subsubsection{Operational Definition of IND and OOD}
\label{app:operational-ind-ood}

For the cross-benchmark mathematical aggregates, we use an operational IND/OOD grouping based on the teachers' empirical results to post-training.
This choice addresses a practical challenge: training corpora assembled from multiple sources and independently constructed evaluation benchmarks need not provide corresponding train/test splits.
Evaluating across these benchmarks allows us to examine a broader range of post-training settings, while requiring an explicit criterion for grouping their relationships to training.

Performance-based, model-relative operationalizations have precedents.
\citet{kim2025rldistillation} define in-distribution questions through a model's probability of answering correctly, while \citet{guo2023predictfuture} construct ID/OOD labels from prediction correctness relative to reference models.
\citet{jesson2025generativeicl} similarly use performance to distinguish in-capability and out-of-capability tasks when the pre-training distribution is opaque.
These approaches motivate using observable model behavior to characterize task compatibility when distributional membership is difficult to establish.

Our specific criterion groups a benchmark as IND when each of the three self-RL teachers exceeds 97\% of the base model's score, identifying evaluations on which performance is improved or largely preserved after teacher training.
The remaining benchmarks are grouped as OOD.
Thresholds of 95\%, 96\%, and 98\% produce the same partition.

This grouping is a descriptive proxy for compatibility with the teachers' post-training regime.
Teacher responses also depend on optimization, interference, and ceiling effects, so the labels do not establish distributional membership or pre-training exposure.
Our controlled study in \Cref{fig:preliminary} provides complementary evidence using an explicitly specified training range, independently of this performance-based grouping.

\subsection{Scores and Aggregation}
\label{app:scoring}
AIME avg@8 averages correctness over eight sampled responses per problem; pass@8 counts whether any response is correct. Student/teacher score ratios compare performance, while OOD retention compares the student with the base when teacher is forgetting. 
For mean OOD forgetting $L_m=\bar{s}_{\mathrm{base}}-\bar{s}_m$, the reduction relative to reference $r$ is $100(L_r-L_m)/L_r\%$.
We assess OOD retention and improvement relative to the base, and separately compare the student with the teacher or best participating teacher to assess transfer and any additional benefit from consolidation.
%
%
Compositional summaries first average the component or composed benchmarks within each setting, then weight the three settings equally. Component references use the assigned teachers; composed-task references use the highest teacher score on each benchmark. 

\section{Additional Results}
\label{app:math-results}

\subsection{Single-Component OPD and Compositional Generalization}
\label{app:code-long-context-controls}
\begin{figure}[!htbp]
\centering
\includegraphics[width=0.97\linewidth]{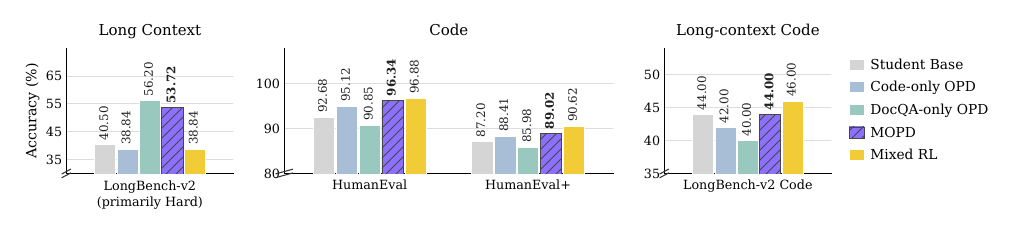}
\caption{\textbf{Single-component OPD improves its target benchmarks, while MOPD's component improvements do not yield a gain} over the base on LongBench-v2 Code. All students start from Qwen3-8B. Full scores, including both external teachers, appear in \Cref{tab:code-long-context-controls}.}
\label{fig:code-long-context-controls}
\end{figure}

\Cref{fig:code-long-context-controls,tab:code-long-context-controls} compare single-component OPD and MOPD from the same Qwen3-8B base. Code-only OPD improves HumanEval from 92.68 to 95.12 and HumanEval+ from 87.20 to 88.41; DocQA-only OPD improves LongBench-v2 (primarily Hard) from 40.50 to 56.20. Thus, both single-component OPD runs improve their respective target benchmarks. 

MOPD improves all three component benchmarks over the base, exceeds code-only OPD on both coding benchmarks, and retains $84.2\%$ of the DocQA-only student's positive LongBench-v2 (primarily Hard) gain. Nevertheless, its LongBench-v2 Code score remains at the base's 44.00. This exceeds the single-component students' 42.00 and 40.00 but remains below both teachers (50.00 and 58.00). These results show that gains on the individual components can coexist with a failure to improve the composed-task score. 

\begin{table}[!htbp]
\centering\small
\setlength{\tabcolsep}{5pt}
\caption{Code and long-context component and composed-task scores (\%). MOPD exceeds both single-component students on LongBench-v2 Code, but matches the base and remains below both teachers. Blank cells indicate teacher evaluations not reported in the original comparison. Teacher assignments are specified in \Cref{app:training-protocols}.}
\label{tab:code-long-context-controls}
\begin{tabular}{@{}lrrrr@{}}
\toprule
Model & HumanEval & HumanEval+ & \makecell{LongBench-v2\\(Hard)} & \makecell{LongBench-v2\\Code}\\
\midrule
Qwen3-8B base & 92.68 & 87.20 & 40.50 & 44.00\\
Code-only OPD  & 95.12 & 88.41 & 38.84 & 42.00\\
DocQA-only OPD  & 90.85 & 85.98 & 56.20 & 40.00\\
Code + DocQA MOPD  & 96.34 & 89.02 & 53.72 & 44.00\\
Code + DocQA mixed RL & 96.88 & 90.62 & 38.84 & 46.00\\
\midrule
Coding teacher & 97.56 & 90.85 & & 50.00\\
Long-context teacher & & & 61.98 & 58.00\\
\bottomrule
\end{tabular}
\end{table}
\FloatBarrier

\subsection{Distant Coding Evaluations}
\label{app:distant-coding}
\begin{figure}[!htbp]
\centering
\includegraphics[width=\linewidth]{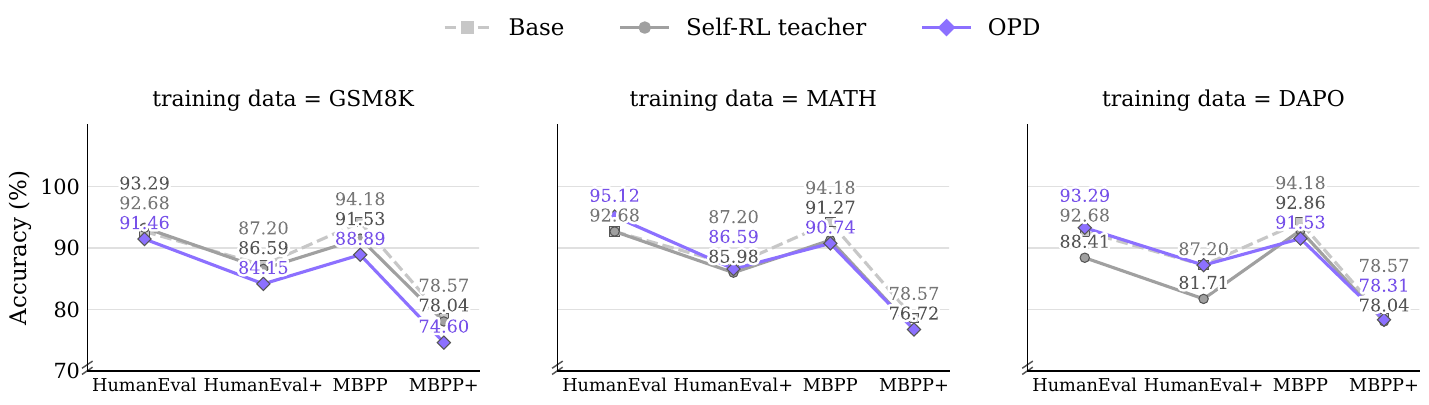}
\caption{Coding performance after mathematical reasoning specialization and distillation. Each panel compares the base, self-RL teacher, and OPD student trained on the named dataset. Scores are percentages under the EvalPlus protocol in \Cref{app:datasets}.}
\label{fig:distant-coding}
\end{figure}

Mathematical reasoning specialists and their OPD students are evaluated on coding benchmarks in \Cref{fig:distant-coding}. Overall, forgetting extends beyond math reasoning OOD tasks. On most coding benchmarks, OPD performs worse than the base.

\FloatBarrier

\FloatBarrier
\Needspace{8\baselineskip}
\section{Arithmetic Study: Setup and Additional Results}
\label{app:arithmetic}


\paragraph{Task.}
The base model is pretrained on arithmetic operators $\tau_i(x)=(a_i x+b_i)\bmod K$, for $i=1,\ldots,n$. 
Conceptually, we analogy each operator as an atomic capability.
Post-training introduces two held-out operators, $A$ and $B$. Component training covers $A$, $B$, $AA$, and $BB$, with a small set of bridge examples composing $A$ or $B$ with pretrained operators. The target compositions $AB$ and $BA$ are excluded from component training, bridge examples, and distillation, and are used to evaluate compositional generalization.

\paragraph{Training.}
SFT specialists learn $A/AA$ or $B/BB$ separately; warm-RL specialists use SFT warmup followed by RL with verifiable rewards. Mixed SFT and mixed warm RL jointly train on both components. MOPD students start from the shared pretrained base and distill either the SFT specialists or the warm-RL specialists, using reverse KL or its forward-KL counterpart. A separate direct-SFT control receives $AB/BA$ supervision to test task learnability. The mixed warm-RL trajectory in \Cref{fig:arithmetic-discussion} includes its 100-step SFT warmup; MOPD trajectories show student training, excluding teacher preparation.

\paragraph{Evaluation and findings.}
\Cref{fig:arithmetic-discussion} plots $\min(\mathrm{AB\ accuracy},\mathrm{BA\ accuracy})$ for seed 0; final accuracies are averaged over five seeds and results are consistent. 
On average, Mixed warm RL attains mean $AB/BA$ accuracies of $93\%/90\%$, compared with $78\%/74\%$ for MOPD from warm-RL teachers and $77\%/74\%$ for its forward-KL variant. 
For reference, direct SFT on the composed tasks reaches $100\%$ on both across all five seeds. 
The warm-RL specialists individually attain only $1$-$2\%$ mean accuracy on $AB/BA$, so MOPD already generalizes beyond their composed-task performance. 

\FloatBarrier
\Needspace{8\baselineskip}
\section{OPD-RL Geometry: Norm, Variance, and Task Interference}
\label{app:opd-rl-theory}
\begingroup
\renewcommand{\theequation}{\thesection.\arabic{equation}}
\setcounter{equation}{0}

%
%
%
We seek a local geometric explanation for two observations: OPD approaches RL's IND performance with less OOD forgetting, and MOPD exhibits less cross-task conflict than mixed RL. The analysis concerns the initial step at the base policy and provides a local approximation for multi-step training; it does not establish universal superiority of OPD.

%
%
%
%
\subsection{Local OPD and RL Update Directions}
\label{app:theory-local-directions}
For a fixed context $h$, let $p_0(a\mid h)=\pi_{\theta_0}(a\mid h)$ be the base student policy and $q(a\mid h)$ the fixed teacher. Define the score function and teacher log-ratio as
\begin{equation}
S_0(a,h)=\left.\nabla_\theta\log\pi_\theta(a\mid h)\right|_{\theta=\theta_0},
\qquad
L(a,h)=\log\frac{q(a\mid h)}{p_0(a\mid h)}.
\label{eq:theory-score-logratio}
\end{equation}
Unless stated otherwise, expectations, variances, and covariances below are over $a\sim p_0(\cdot\mid h)$, and we suppress the context argument where convenient. Let $A=R-\mathbb{E}_{p_0}[R]$ denote the reward advantage, with the reward scale chosen so that $\mathbb{E}_{p_0}[A]=0$ and $\operatorname{Var}_{p_0}(A)=1$, analogous to advantage normalization in GRPO. The sampled-token OPD update direction and ideal on-policy RL gradient are
\begin{equation}
g_{\mathrm{OPD}}(h)=\mathbb{E}_{p_0}[L S_0],
\qquad
g_{\mathrm{RL}}(h)=\mathbb{E}_{p_0}[A S_0].
\label{eq:theory-gradients}
\end{equation}
Here $g_{\mathrm{OPD}}$ follows the direction of decreasing reverse KL, while $g_{\mathrm{RL}}$ follows the direction of increasing reward.

\subsection{Decomposition for a General Reward-Better Teacher}
\label{app:theory-general-teacher}
Suppose the teacher improves expected reward over the base, $\mathbb{E}_{q}[R]>\mathbb{E}_{p_0}[R]$, without necessarily being RL-optimal. Writing $w=q/p_0$, this condition is equivalent to $\mathbb{E}_{p_0}[wA]>0$. It constrains the density ratio $w$, whereas sampled-token OPD uses $L=\log w$; higher teacher reward alone therefore does not imply alignment between the OPD and RL gradients.

Project the centered log-ratio onto the normalized advantage:
\begin{equation}
L-\mathbb{E}_{p_0}[L]=\eta A+\xi,
\qquad
\eta=\frac{\operatorname{Cov}_{p_0}(L,A)}{\operatorname{Var}_{p_0}(A)}
=\operatorname{Cov}_{p_0}(L,A),
\label{eq:theory-logratio-decomposition}
\end{equation}
where $\mathbb{E}_{p_0}[\xi]=0$ and $\operatorname{Cov}_{p_0}(\xi,A)=0$. The score identity $\mathbb{E}_{p_0}[S_0]=\left.\nabla_\theta\sum_a\pi_\theta(a\mid h)\right|_{\theta=\theta_0}=0$ eliminates the action-independent term $\mathbb{E}_{p_0}[L]$, giving
\begin{equation}
\begin{aligned}
g_{\mathrm{OPD}}(h)
&=\mathbb{E}_{p_0}[(L-\mathbb{E}_{p_0}[L])S_0]
=\eta\mathbb{E}_{p_0}[A S_0]+\mathbb{E}_{p_0}[\xi S_0]\\
&=\eta g_{\mathrm{RL}}(h)+r(h),
\qquad r(h)=\mathbb{E}_{p_0}[\xi S_0].
\end{aligned}
\label{eq:theory-gradient-decomposition}
\end{equation}
The first term is the reward-aligned component of OPD. The residual $r(h)$ captures teacher-specific changes not explained by the reward advantage, such as shifts in style, format, or domain distribution.

%
%
%
\subsection{Gradient Norm and Coefficient Variance}
\label{app:theory-norm-variance}
Define $F(h)=\mathbb{E}_{p_0}[\lVert S_0\rVert_2^2]$ and $\sigma_\xi^2(h)=\operatorname{Var}_{p_0}(\xi)$. Orthogonality of $A$ and $\xi$ gives $\operatorname{Var}_{p_0}(L)=\eta^2+\sigma_\xi^2(h)$. By Cauchy-Schwarz,
\begin{equation}
\begin{aligned}
\lVert g_{\mathrm{OPD}}(h)\rVert_2^2
&=\bigl\lVert\mathbb{E}_{p_0}[(L-\mathbb{E}_{p_0}[L])S_0]\bigr\rVert_2^2
\le \bigl(\eta^2+\sigma_\xi^2(h)\bigr)F(h),\\
\lVert g_{\mathrm{RL}}(h)\rVert_2^2
&=\bigl\lVert\mathbb{E}_{p_0}[A S_0]\bigr\rVert_2^2
\le \operatorname{Var}_{p_0}(A)F(h)=F(h).
\end{aligned}
\label{eq:theory-norm-bounds}
\end{equation}
Thus, when $\eta^2+\sigma_\xi^2(h)<1$, the OPD gradient-norm upper bound is smaller than the corresponding bound for normalized RL.

For stochastic updates, write $Z_{\mathrm{RL}}=\widehat A S_0$ and $Z_{\mathrm{OPD}}=L S_0$, where $\widehat A$ is the estimated advantage. If $\operatorname{Var}(\widehat A)\approx1$, the same condition gives OPD a smaller coefficient variance. With comparable score-norm weighting and baseline effects, this can also reduce gradient-estimator variance. This is a secondary mechanism: lower estimator variance can reduce random non-target drift during stochastic optimization.
\subsection{An RL-Optimal Teacher Preserves the RL Direction}
\label{app:theory-optimal-teacher}
Consider the ideal case in which the teacher solves KL-regularized policy improvement from the base, with $\eta>0$:
\begin{equation}
q^\star=\arg\max_\pi\left\{
\mathbb{E}_{a\sim\pi}[A(a,h)]
-\frac{1}{\eta}D_{\mathrm{KL}}\bigl(\pi(\cdot\mid h)\|p_0(\cdot\mid h)\bigr)
\right\}.
\label{eq:theory-kl-improvement}
\end{equation}
The solution is the exponentially tilted policy
\begin{equation}
q^\star(a\mid h)=\frac{p_0(a\mid h)e^{\eta A(a,h)}}{Z(h)},
\qquad
Z(h)=\mathbb{E}_{p_0}[e^{\eta A}],
\qquad
\log\frac{q^\star}{p_0}=\eta A-\log Z(h).
\label{eq:theory-optimal-policy}
\end{equation}
The score identity eliminates $-\log Z(h)$, yielding the exact local relation
\begin{equation}
g_{\mathrm{OPD}}^\star(h)
=\mathbb{E}_{p_0}[(\eta A-\log Z(h))S_0]
=\eta g_{\mathrm{RL}}(h).
\label{eq:theory-ideal-alignment}
\end{equation}
Here $\xi=r=0$, $\lVert g_{\mathrm{OPD}}^\star(h)\rVert_2=\eta\lVert g_{\mathrm{RL}}(h)\rVert_2$, and $\operatorname{Var}_{p_0}(\log(q^\star/p_0))=\eta^2$. An ideal RL teacher therefore preserves the RL direction and contracts its magnitude when $0<\eta<1$.

%
%
%
\paragraph{Interpreting the contraction factor.}
For normalized $A$ and sufficiently small $\eta$, $\log Z(h)=\log\mathbb{E}_{p_0}[e^{\eta A}]\approx\eta^2/2$. Consequently,
\begin{equation}
D_{\mathrm{KL}}(p_0\|q^\star)=\log Z(h)\approx\frac{\eta^2}{2},
\qquad
\eta\approx\sqrt{2D_{\mathrm{KL}}(p_0\|q^\star)}.
\label{eq:theory-contraction-kl}
\end{equation}
A token-level KL of $0.02$-$0.30$ nats/token corresponds approximately to $\eta\in[0.20,0.77]$. This range reflects the base models and post-training tasks in our experiments, rather than a universal property of post-training. Empirically, we observe OPD gradient norms of approximately $60\%$ of those of RL.

%
\subsection{Implications for IND Performance and OOD Retention}
\label{app:theory-ind-ood}
In the low-residual teacher regime, $g_{\mathrm{OPD}}=\eta g_{\mathrm{RL}}+r$ with $0<\eta<1$ and small $\lVert r\rVert_2$. OPD approximately preserves RL's reward-improving direction while contracting the update magnitude and reducing coefficient variance.

%
%
%
%
\paragraph{IND performance.}
A smaller update need not imply substantially worse IND performance: the local improvement depends on its projection onto the reward-improving direction. For an IND objective $J_{\mathrm{in}}$,
\begin{equation}
J_{\mathrm{in}}(\theta_0+\Delta\theta)-J_{\mathrm{in}}(\theta_0)
\approx\langle\nabla J_{\mathrm{in}}(\theta_0),\Delta\theta\rangle.
\label{eq:theory-ind-expansion}
\end{equation}
When $g_{\mathrm{OPD}}\approx\eta g_{\mathrm{RL}}$, OPD retains the reward-aligned direction. A sufficiently large positive projection onto that direction can therefore yield IND improvement close to RL's, despite a smaller total update norm.

\paragraph{OOD forgetting.}
OOD forgetting depends on the displacement from the base policy. A local second-order expansion of the OOD loss gives
\begin{equation}
\begin{aligned}
\Delta L_{\mathrm{ood}}
&\equiv L_{\mathrm{ood}}(\theta_0+\Delta\theta)-L_{\mathrm{ood}}(\theta_0)\\
&\approx\langle\nabla L_{\mathrm{ood}}(\theta_0),\Delta\theta\rangle
+\tfrac12\Delta\theta^\top H_{\mathrm{ood}}\Delta\theta,
\end{aligned}
\label{eq:theory-ood-expansion}
\end{equation}
where $H_{\mathrm{ood}}$ is the Hessian at $\theta_0$. If the base lies in a locally stable region for the OOD capability, the first-order term is small and the quadratic term dominates:
\begin{equation}
\Delta L_{\mathrm{ood}}
\lesssim\frac{\lambda_{\max}(H_{\mathrm{ood}})}{2}\lVert\Delta\theta\rVert_2^2.
\label{eq:theory-ood-norm-bound}
\end{equation}
The corresponding local policy-space displacement is
\begin{equation}
\mathbb{E}_{h\sim\mathcal{D}_{\mathrm{ood}}}
D_{\mathrm{KL}}\!\left(\pi_{\theta_0}(\cdot\mid h)\middle\|
\pi_{\theta_0+\Delta\theta}(\cdot\mid h)\right)
\approx\tfrac12\Delta\theta^\top F_{\mathrm{ood}}\Delta\theta,
\label{eq:theory-ood-policy-drift}
\end{equation}
where $F_{\mathrm{ood}}$ is the Fisher information matrix under the base policy and OOD contexts. For approximately aligned, contracted updates, $\Delta\theta_{\mathrm{OPD}}\approx\eta\Delta\theta_{\mathrm{RL}}$, we have $\lVert\Delta\theta_{\mathrm{OPD}}\rVert_2\approx\eta\lVert\Delta\theta_{\mathrm{RL}}\rVert_2$. The local quadratic approximation then suggests
\begin{equation}
\Delta L_{\mathrm{ood}}^{\mathrm{OPD}}
\lesssim\eta^2\Delta L_{\mathrm{ood}}^{\mathrm{RL}}.
\label{eq:theory-ood-contraction}
\end{equation}
Thus, for $\eta<1$, contraction reduces the quadratic contribution to OOD drift under these local assumptions. Lower stochastic coefficient variance, when $\operatorname{Var}(L)=\eta^2+\sigma_\xi^2<1$, provides a secondary mechanism for reducing random non-target drift and improving retention.
\subsection{Two-Task Interference in MOPD and Mixed RL}
\label{app:theory-task-interference}
Consider an equal mixture of tasks $A$ and $B$. With task-specific teachers satisfying $g_i^{\mathrm{OPD}}=\eta_i g_i^{\mathrm{RL}}+r_i$ for $i\in\{A,B\}$, the expected mixture gradients are
\begin{equation}
\begin{aligned}
g_{\mathrm{mix}}^{\mathrm{RL}}
&=\tfrac12g_A^{\mathrm{RL}}+\tfrac12g_B^{\mathrm{RL}},\\
g_{\mathrm{mix}}^{\mathrm{MOPD}}
&=\tfrac12(\eta_Ag_A^{\mathrm{RL}}+r_A)
+\tfrac12(\eta_Bg_B^{\mathrm{RL}}+r_B).
\end{aligned}
\label{eq:theory-mixture-gradients}
\end{equation}
Measure cross-task interference by the gradient inner product, $I_{\mathrm{RL}}(A,B)=\langle g_A^{\mathrm{RL}},g_B^{\mathrm{RL}}\rangle$. For MOPD,
\begin{equation}
\begin{aligned}
\langle g_A^{\mathrm{OPD}},g_B^{\mathrm{OPD}}\rangle
&=\eta_A\eta_B\langle g_A^{\mathrm{RL}},g_B^{\mathrm{RL}}\rangle
+\eta_A\langle g_A^{\mathrm{RL}},r_B\rangle\\
&\quad+\eta_B\langle r_A,g_B^{\mathrm{RL}}\rangle+\langle r_A,r_B\rangle.
\end{aligned}
\label{eq:theory-interference-expansion}
\end{equation}
The residual contribution admits the deterministic bound
\begin{equation}
\begin{aligned}
\bigl|\langle g_A^{\mathrm{OPD}},g_B^{\mathrm{OPD}}\rangle
-\eta_A\eta_B\langle g_A^{\mathrm{RL}},g_B^{\mathrm{RL}}\rangle\bigr|
&\le |\eta_A|\lVert g_A^{\mathrm{RL}}\rVert_2\lVert r_B\rVert_2\\
&\quad+|\eta_B|\lVert r_A\rVert_2\lVert g_B^{\mathrm{RL}}\rVert_2
+\lVert r_A\rVert_2\lVert r_B\rVert_2.
\end{aligned}
\label{eq:theory-interference-bound}
\end{equation}
When both residuals are small, $\langle g_A^{\mathrm{OPD}},g_B^{\mathrm{OPD}}\rangle\approx\eta_A\eta_B\langle g_A^{\mathrm{RL}},g_B^{\mathrm{RL}}\rangle$. For $0<\eta_A,\eta_B<1$, MOPD approximately preserves the RL task directions while reducing the magnitude of cross-task interference, providing a possible explanation for smaller multi-task conflict.

%
%
%
\endgroup

\end{document}